\documentclass{article}
\usepackage{iclr2027_conference,times}

\usepackage{graphicx}
\usepackage{wrapfig}
\usepackage{booktabs}
\usepackage{array}      
\usepackage{longtable}  
\usepackage{amsmath}
\usepackage{amssymb}
\usepackage[hidelinks]{hyperref}  
\usepackage{url}
\usepackage{xspace}

\usepackage{enumitem}
\setlist[itemize,1]{leftmargin=*}
\setlist[enumerate,1]{leftmargin=*}

\newcommand{\sysname}{CatchBench\xspace}

\newcommand{\artifactnote}{\ificlrfinal\footnote{Code, data, and the scored board:
\url{https://github.com/yzhao062/catchbench}.}\fi}

\usepackage{graphicx}
\usepackage{xcolor}
\usepackage{colortbl}  
\usepackage{tikz}
\usetikzlibrary{arrows.meta,calc,patterns.meta,positioning}

\definecolor{abGray}{HTML}{C9C9C9}
\definecolor{abMint}{HTML}{BFDFD2}
\definecolor{abCoral}{HTML}{ED8D5A}
\definecolor{abNearBlack}{HTML}{1A1A1A}
\definecolor{abSubtitle}{HTML}{666666}

\title{\sysname: When Can an Agent Failure Be Caught?}

\author{Yue Zhao$^{1}$ \quad Mengyuan Li$^{1}$ \quad Ruolin Li$^{1}$ \quad Prince Zizhuang Wang$^{2}$ \\
\textbf{Shuli Jiang$^{3}$ \quad Linsey Pang$^{4}$ \quad Xiongye Xiao$^{5}$ \quad Xiyang Hu$^{6}$} \\
$^{1}$University of Southern California \quad $^{2}$Carnegie Mellon University \\
$^{3}$AWS Agentic AI \quad $^{4}$PayPal AI \\
$^{5}$University of Tennessee, Knoxville \quad $^{6}$Arizona State University \\
\texttt{\{yue.z, mli49061, ruolinl\}@usc.edu} \quad \texttt{princewang@cmu.edu} \\
\texttt{shulij@alumni.cmu.edu} \quad \texttt{panglinsey@gmail.com} \\
\texttt{xxiao9@utk.edu} \quad \texttt{xiyanghu@asu.edu}}

\newif\ifcatchbenchpreprint
\newcommand{\catchbenchpreprintcopy}{\catchbenchpreprinttrue\iclrfinalcopy}
\catchbenchpreprintcopy

\begin{document}
\maketitle
\ifcatchbenchpreprint
  \fancyhead[L]{}
\else
  \ificlrfinal
    \fancyhead[L]{Published as a conference paper at ICLR 2027}
  \else
    \fancyhead[L]{Under review as a conference paper at ICLR 2027}
  \fi
\fi

\ifcatchbenchpreprint
  \begingroup
  \setlength{\parskip}{4pt}
  \setlength{\textfloatsep}{10pt plus 2pt minus 2pt}
  \setlength{\floatsep}{6pt plus 2pt minus 2pt}
  \setlength{\intextsep}{8pt plus 2pt minus 2pt}
\fi
\begin{abstract}
When can an agent failure be caught? A weak audit score alone cannot identify whether the record
or the method is limiting. \sysname therefore puts one auditor's question to three information states: the declared
configuration before a run (PRE), a growing prefix of its trace (LIVE), and the finished trace
(POST). Prior benchmarks fix one of these states or vary the telemetry; to our knowledge none
scores all three under one task-method interface. Each state admits different questions, so seven task contracts carry
their own labels and metrics rather than one leaderboard. Four are evidential; three are
Gold-derived mechanism diagnostics.

The release scores 72 entrants, from rule scanners and structural models to eleven LLM judges
across nine model families (GPT, Claude, Gemini, Gemma, Llama, Qwen, DeepSeek, Mistral, Nova), over
1187 declared configurations and 1162 recorded runs. Every recorded comparison is published as a
measured difference with its interval, uncorrected, and no board declares a winner it cannot show. The three sharpest results cut against our own data. One rule reads declaration
order alone and reaches a perfect F1 on one of six configuration sources, so a score there
measures how the corpus was built. Our admissibility bar then rejected one injected substrate and
withheld evidential status from the other. A published structural gain also turns on which size
reference it is measured against. A benchmark number is therefore not interpretable until the
process behind its labels is published and tested for the shortcut it may leave. We report all three, and
regenerate every ordering from released predictions with no model call.
\end{abstract}

\section{Introduction}
\label{sec:intro}

\begin{figure*}[t]
  \centering
  \includegraphics[width=\textwidth]{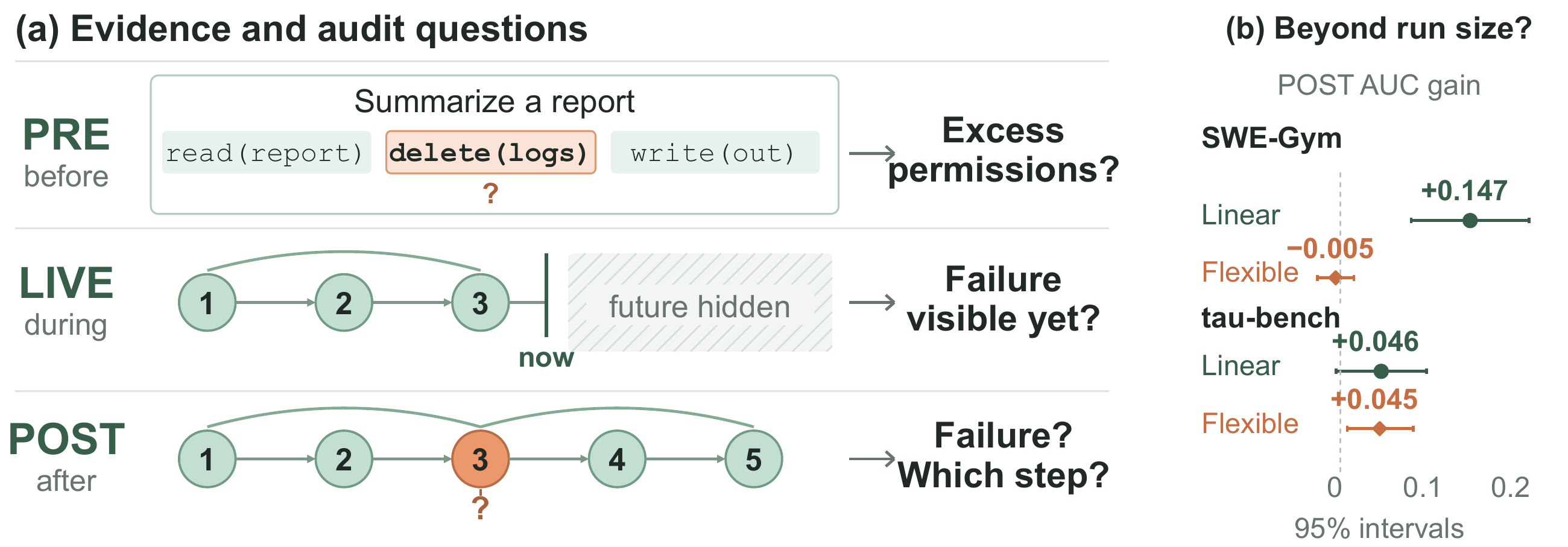}
  \caption{\textbf{The less of a run you can see, the less you can ask about it.}
  (a) What is on the record before, during, and after a run, and the question each version
  supports. Circles are steps and arcs are dependencies; hatching is the part that has not happened
  yet; coral marks the step a method flags, which is never handed to it. The PRE example uses a
  separate corpus.
  (b) POST AUC gains from adding dependency features to linear or flexible (matched spline)
  size controls. Points use seed-averaged out-of-fold predictions; bars are marginal 95\%
  intervals conditional on the saved fits (Table~\ref{tab:core-contrasts}). SWE-Gym's increment is
  unresolved with flexible control; tau-bench's point estimate persists.\looseness=-1}
  \label{fig:lifecycle}
\end{figure*}

An agent failure can only be caught with the evidence its record happens to carry. Three lines of
work have grown around three records: static audits of a declared harness, which never observe a run
\citep{li2026fortis,yan2026authbench,yang2026toolprivbench}; runtime monitors, which must act on a
prefix \citep{zhou2025guardian,wang2025gsafeguard}; and post-hoc attribution, which reads the trace
after the fact \citep{zhang2025whoandwhen,deshpande2025trail,cemri2025mast}. One task, \emph{agent
auditing}, is being measured at three budgets of evidence, its PRE, LIVE, and POST \emph{information
states}, illustrated in Figure~\ref{fig:lifecycle}. Auditing is distinct from tracing, which
represents a run \citep{zhao2026grade,ou2025agentdiagnose}, and from task evaluation, which scores
the final outcome \citep{yao2024taubench}.\artifactnote

What none of them varies is the budget itself. Benchmarks that score these threats fix one
information state, each with a corpus, a metric, and a label process of its own
\citep{yuan2024rjudge,deshpande2025trail}. AgentTelemetry exercises two without scoring either, and
compares telemetry schemas rather than auditors \citep{balusu2026agenttelemetry};
Table~\ref{tab:positioning} lays out the grid. A weak audit score is therefore unattributable: a thin
record and a weak method produce the same number.
\sysname puts that question to each state in turn: seven task contracts
over six scenarios, one task-method interface, on Who\&When, SWE-Gym, tau-bench
\citep{zhang2025whoandwhen,pan2024swegym,yao2024taubench}, and a PRE corpus of declared harnesses
built for this question. Every board carries its own labels, metric, and entrants, so across states
the arrangement fixes what a score means rather than isolating the evidence; the LIVE prefix sweep,
which holds corpus, labels, metric, and entrants fixed, is where evidence alone varies.\looseness=-1

\paragraph{What \sysname Contains.} Nine scored boards over the three states, built from three public
trace corpora (126 failed Who\&When runs, 376 SWE-Gym runs, 660 tau-bench runs) and 1187 declared harnesses from six
sources, each record tagged with the process that labelled it. Seventy-two entrants are scored, from
static rule scanners and structural, graph, and off-the-shelf detectors to eleven LLM judges across
nine model families.\looseness=-1

\paragraph{Contributions.} \textbf{(1)~The first arena, to our knowledge, that scores an auditor at
all three information states.} No prior benchmark in Table~\ref{tab:positioning} scores an auditor
at more than one of the three.
\textbf{(2)~The larger structural gain is less robust to the size control.} On seed-averaged
out-of-fold predictions, adding dependency features to SWE-Gym gives $+0.147$ ROC-AUC with linear
size controls but $-0.005$ with matched splines; on tau-bench the increments are $+0.046$ and
$+0.045$. Table~\ref{tab:core-contrasts} reports all four exploratory contrasts with their marginal
intervals. A linear size control alone therefore cannot establish a general dependency-feature
advantage. \textbf{(3)~A PRE corpus that records how
each label was made, which caught a leak in our own data.} In one of the six sources the capability a task needs is
always declared first, so a rule reading position alone, no name and no permission, finds all 510
excess capabilities there with no false alarm and no miss. That score measures how the corpus was
built, and a pooled number would have hidden it. \textbf{(4)~An admissibility bar that our own
injected data did not clear.} Labeled dependency-state failures are scarce, so \sysname-Gold plants known
faults inside real runs. Neither injected substrate cleared the bar, so the three Gold boards ship as
mechanism diagnostics.

\section{Related Work}
\label{sec:related}

Two lines of work bound this one. ADBench established a common tabular anomaly-detection arena with
fixed datasets, many methods, and comparisons across supervision regimes \citep{han2022adbench}, and
BOND carried that template to attributed graphs with outlier-type breakdowns and a runtime table
\citep{liu2022bond}. Neither audits agents; what they supply is the shape of the comparison.
\sysname inherits that fixed-data, many-method, per-type contract, and differs by organizing agent
audits around PRE, LIVE, and POST information states under one task-method interface.\looseness=-1

\begin{wrapfigure}{R}{0.42\textwidth}
  \vspace{-24pt}
  \centering
  \includegraphics[width=\linewidth]{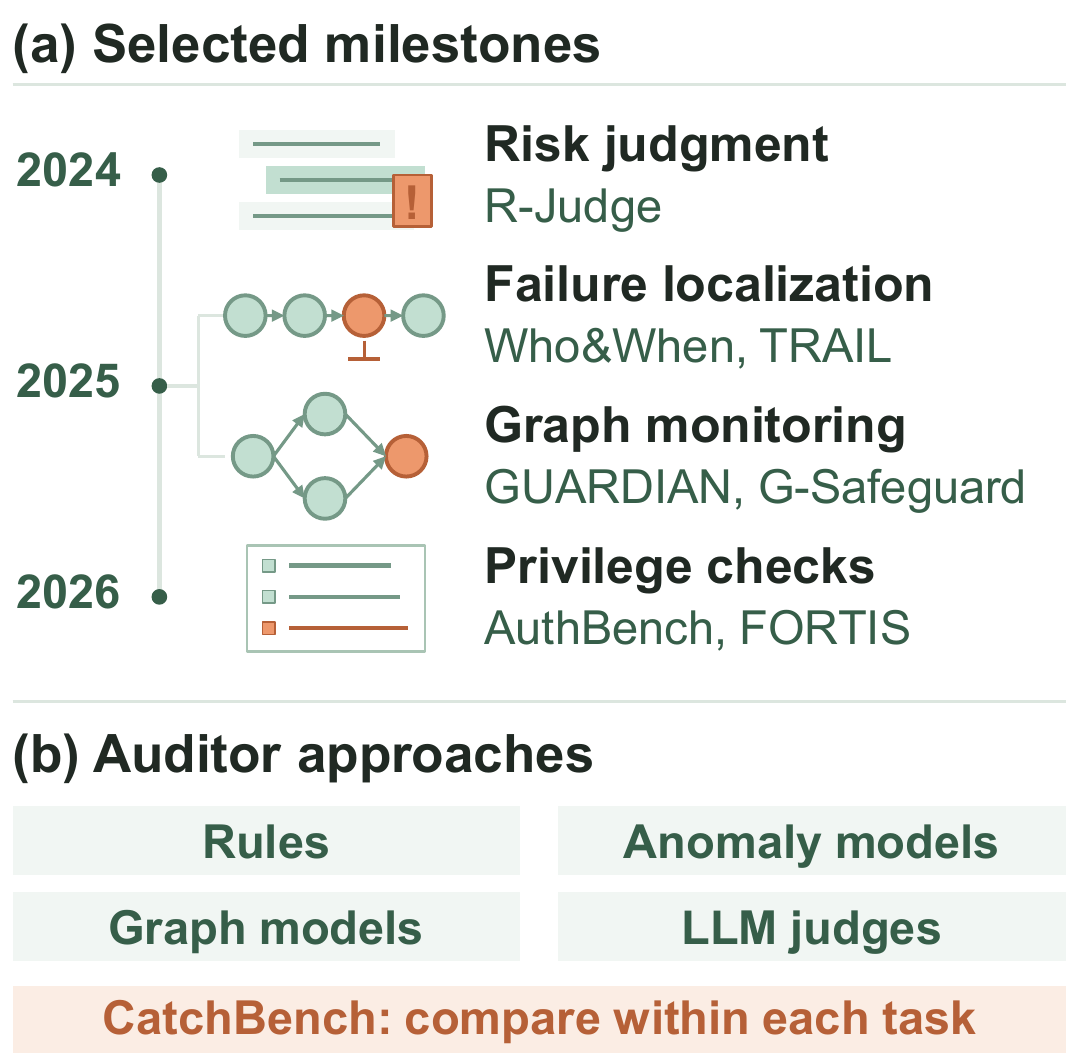}
  \caption{Selected milestones (first public year) and auditor families. Research branches coexist.}
  \label{fig:research-landscape}
  \vspace{-6pt}
\end{wrapfigure}

The second line scores the same threats on agent runs, under task definitions that differ from ours
in what the evaluated auditor may see. R-Judge asks models to identify safety risks in annotated
interaction records \citep{yuan2024rjudge}, and Agent Security Bench executes attacks and defenses
across tool-using agent scenarios \citep{zhang2025agentsecuritybench}; these make a broad
first-agent-auditing claim untenable. Table~\ref{tab:positioning} places \sysname among the
benchmarks that score an external auditor. The axis is the information a method is allowed to read.
Figure~\ref{fig:research-landscape} summarizes selected milestones and auditor families.\looseness=-1

\begin{table*}[t]
\centering
\footnotesize
\setlength{\tabcolsep}{1.8pt}
\renewcommand{\arraystretch}{1.06}
\caption{Which information state each benchmark lets its evaluated auditor read. $\bullet$ scored
there, $\circ$ exercised with no auditor scored, blank not covered. Label-evidence marks say how a
target was established, not how good it is; \emph{Auto.}\ covers construction, injection, and program
oracles. \emph{Scale} keeps each paper's own unit and does not compare across rows.}
\label{tab:positioning}
\vspace{2pt}
\begin{tabular}{@{}l !{\color{abGray}\vrule width 0.4pt} ccc !{\color{abGray}\vrule width 0.4pt} ccc !{\color{abGray}\vrule width 0.4pt} rl@{}}
\toprule
& \multicolumn{3}{c}{Auditor may read} & \multicolumn{3}{c}{Label evidence} & & \\
\cmidrule(lr){2-4}\cmidrule(lr){5-7}
Benchmark & PRE & LIVE & POST & Human & Model & Auto. & \multicolumn{2}{c}{Scale} \\
\midrule
\rowcolor{abGray!35}\multicolumn{9}{@{}l}{\textit{Pre-execution privilege}}\\
AuthBench \citep{yan2026authbench}
  & $\bullet$ &           &           & \checkmark &            & \checkmark &        120 & tasks \\
ToolPrivBench \citep{yang2026toolprivbench}
  & $\bullet$ & $\circ$   &           & \checkmark & \checkmark &            &        544 & scenarios \\
FORTIS \citep{li2026fortis}
  & $\bullet$ &           &           & \checkmark &            & \checkmark &    2{,}143 & queries \\
\rowcolor{abGray!35}\multicolumn{9}{@{}l}{\textit{Complete-trace diagnosis}}\\
R-Judge \citep{yuan2024rjudge}
  &           &           & $\bullet$ & \checkmark &            &            &        569 & records \\
TRAIL \citep{deshpande2025trail}
  &           &           & $\bullet$ & \checkmark &            &            &        148 & traces \\
MAST \citep{cemri2025mast}
  &           &           & $\bullet$ & \checkmark & \checkmark &            &    1{,}642 & traces \\
Who\&When \citep{zhang2025whoandwhen}
  &           &           & $\bullet$ & \checkmark &            &            &        184 & tasks \\
Who\&When Pro \citep{liu2026whowhenpro}
  &           & $\circ$   & $\bullet$ & \checkmark &            & \checkmark &   12{,}326 & traces \\
AgentErrorBench \citep{zhu2025agentdebug}
  &           &           & $\bullet$ & \checkmark &            &            &        200 & traces \\
CUAErrorBench \citep{zhang2026cuadebug}
  &           &           & $\bullet$ & \checkmark &            &            &        204 & trajectories \\
AgenTracer \citep{zhang2025agentracer}
  &           &           & $\bullet$ &            &            & \checkmark & $>$2{,}000 & pairs \\
\rowcolor{abGray!35}\multicolumn{9}{@{}l}{\textit{Runtime monitoring}}\\
AgentTelemetry \citep{balusu2026agenttelemetry}
  &           & $\circ$   & $\circ$   &            &            & \checkmark &    2{,}940 & configs \\
\midrule
\rowcolor{abCoral!25}\sysname (ours)
  & $\bullet$ & $\bullet$ & $\bullet$ & \checkmark & \checkmark & \checkmark &    1{,}187 & configs $+$ 1{,}162 runs \\
\bottomrule
\end{tabular}
\vspace{-0.15in}
\end{table*}

Appendix~\ref{app:related-extended} gives the per-work contract behind most of these rows and the
adjacent work the table leaves out, including how outcome and process evaluators define their
targets \citep{fan2026agentprocessbench,lu2025agentrewardbench,zheng2025processbench}, how
AgentTelemetry varies telemetry rather than auditors, and how AgentDebugX evaluates a debugger on
the Who\&When substrate. The distinction drawn here is a task contract, rather than a claim that
agent-auditing benchmarks do not exist.\looseness=-1

\section{The Benchmark}
\label{sec:benchmark}

Three constraints force this design, and each one is a fact about auditing agents rather than a
preference of ours. The record bounds the audit: a method that reads only a declared configuration
cannot be moved to a board that reads a finished trace. Labels for agent failures are scarce, so a
found corpus can be scored for how it was assembled rather than for how a method reasons. Every
record we release therefore carries its label process (Section~\ref{sec:data}), and synthesized
faults must clear an admissibility bar before they carry weight (Section~\ref{sec:data-gold}). The
boards are also small, so the uncertainty around a score is as much a part of a result as the
score. Every comparison this paper records therefore prints its difference and a marginal 95\%
interval (Section~\ref{sec:scoring-contract}). Marginal means the interval covers that comparison
alone and claims no simultaneous coverage across the others. The paper stops there.\looseness=-1

\subsection{The Three Information States}
\label{sec:info-states}

The evidence available at each state (Figure~\ref{fig:lifecycle}) fixes which audit is possible at all.
Before execution only the
plan and the harness exist, so the audit is static (PRE). During execution a growing prefix is
visible while the outcome is not, so it is predictive and runs under a false-alarm budget (LIVE).
After execution the complete trace is available, so it is forensic (POST). The scoring contract
controls access to outcome labels: detection entrants do not receive the outcome they must predict.
A PRE scanner and a POST detector never share a leaderboard column. On the current corpora,
the LIVE sweep at 100\% reads the finished trace and reproduces the shared entrants' POST detection scores.

\subsection{Audit Scenarios and Their Metrics}

Within an information state, each scenario is the specific question an auditor asks there, paired
with the label that answers it. PRE asks whether the declared plan or harness is over-privileged
before the run ever starts, judged against the minimal reference each source's label process
supplies (Section~\ref{sec:res-pre}). LIVE asks two questions of a growing prefix: whether failure
can be called early (streaming early warning), and whether a fault can be caught as it happens
without false-alarming (online detection). POST asks the three forensic questions of a finished
trace: which step failed (localization), whether it failed at all (detection), and what kind of
fault it was (cause attribution).

These six scenarios produce seven task contracts, because localization has two instantiations:
human-labeled natural failures on Who\&When and injected faults on \sysname-Gold. A contract fixes
the task, label, metric, and baseline set. Two of the seven run against two corpora each, so the
runner emits the nine scored blocks of Table~\ref{tab:boards}. Its textual output carries one further
block, the Gold v2 admissibility diagnostic of Section~\ref{sec:data-goldv2}, whose rows are
construction controls and oracles; it carries no scored entrant,
and every board, entrant, and contrast count in this paper excludes it.

\paragraph{Entrant Labels.} On the detection and early-warning boards \texttt{size (flat)} reads run
size and event counts; relative to that same block \texttt{auditable (size+deps)} adds
size-normalized dependency density and shape, \texttt{full} adds execution-topology and raw
dependency features, and \texttt{pyod-flatten (ECOD)} is an off-the-shelf unsupervised detector over
the flat features. The two agent-specific graph baselines, \texttt{guardian (recon-AE)} and
\texttt{g-safeguard (sup GNN)}, adapt published mechanisms to this representation rather than
reproducing them \citep{zhou2025guardian,wang2025gsafeguard}. Localization entrants differ:
\texttt{exec-rank (sup.)} is GRADE's supervised execution-feature ranker \citep{zhao2026grade},
\texttt{auditable} (blast) ranks steps by downstream dependency reach, and \texttt{pygod (graph AD)}
scores steps with an unsupervised graph autoencoder. Appendix~\ref{app:entrants} states what each
adaptation keeps and drops.

\begin{table}[!t]
\centering
\footnotesize
\setlength{\tabcolsep}{4pt}
\renewcommand{\arraystretch}{1.10}
\caption{The nine scored blocks. \emph{Floor} is the score of the board's trivial entrant: flag-all for PRE
and the empirical random reference elsewhere. \emph{Field} is the spread of the scored entrants, excluding trivial policies and
oracles, and orders nothing; the two streaming rows average their four prefixes, and the online
stale-state row is scored once at a fixed false-positive budget. $\dagger$ (mint): the Gold boards
are mechanism diagnostics (Section~\ref{sec:data-gold}); Appendix~\ref{app:stats} carries every
recorded comparison with its interval.}
\label{tab:boards}
\vspace{3pt}
\begin{tabular}{@{}ll rl l r c@{}}
\toprule
Board & Corpus & \multicolumn{2}{c}{Size} & Metric & Floor & Field \\
\midrule
\rowcolor{abGray!35}\multicolumn{7}{@{}l}{\textbf{PRE}: before the run, plan and harness only}\\
Over-privilege & 6 sources & 1{,}187 & configs & F1 & 0.601 & 0.020--0.695 \\
\midrule
\rowcolor{abGray!35}\multicolumn{7}{@{}l}{\textbf{LIVE}: during the run, a growing prefix}\\
Early warning & SWE-Gym & 376 & runs & ROC-AUC & 0.483 & 0.534--0.818 \\
Early warning & tau-bench & 660 & runs & ROC-AUC & 0.498 & 0.541--0.645 \\
\rowcolor{abMint!35}Online stale$^{\dagger}$ & Gold & 82 & injections & TPR@5\%FPR & 0.024 & 0.061--0.122 \\
\midrule
\rowcolor{abGray!35}\multicolumn{7}{@{}l}{\textbf{POST}: after the run, the complete trace}\\
Localization & Who\&When & 126 & runs & Top-1 & 0.119 & 0.048--0.452 \\
Detection & SWE-Gym & 376 & runs & ROC-AUC & 0.483 & 0.319--0.850 \\
Detection & tau-bench & 660 & runs & ROC-AUC & 0.498 & 0.490--0.665 \\
\rowcolor{abMint!35}Localization$^{\dagger}$ & Gold & 188 & runs & Top-1 & 0.032 & 0.000--0.309 \\
\rowcolor{abMint!35}Cause attribution$^{\dagger}$ & Gold & 166 & paired runs & ROC-AUC & 0.498 & 0.566--0.675 \\
\bottomrule
\end{tabular}
\vspace{-0.1in}
\end{table}

The metric follows from the question: Top-k and MRR for localization, ROC-AUC for detection and the
present two-class cause-attribution board, mean prefix ROC-AUC and time to detection for streaming,
and the true-positive rate at a fixed false-positive budget for the online question.

\subsection{The Scoring Contract}
\label{sec:scoring-contract}

Nine blocks with their own labels, metrics, and entrant fields are not yet one benchmark. What makes
them one is a shared task-method interface: adding a method means declaring which task ids it
supports and returning a metric dictionary.

Within a board, that interface licenses the comparisons. Within the 138 registered comparisons,
entrant-versus-entrant differences are paired, and Appendix~\ref{app:stats} names their interval
constructions. The two early-warning bar groups instead compare one entrant with the fixed constant
0.70, with 20 cells per group and one-arm intervals. Both are two-sided. The
tau-bench group was fixed before scoring; its SWE-Gym counterpart was assembled after those scores
were examined, so its readings are exploratory rather than confirmatory. No correction is applied
across the groups and none is claimed; the groups order a long appendix table for a reader rather
than defining a multiplicity this paper controls. \texttt{tools/statistical\_tests.py} regenerates
the 138 recorded comparisons using committed judge caches and PRE records together with scores
recomputed from upstream corpora.

Corpus size bounds how sharply any of this can be measured, and every interval in
Appendix~\ref{app:stats} is that bound made explicit for one comparison. Cross-validation
seed spread and run-level sampling answer different questions and can disagree, so each printed
interval states the axis it covers. \texttt{auditable} is the reference implementation released
with the benchmark, and it appears throughout as a scored entrant rather than as the referee.

\section{The Data}
\label{sec:data}

\sysname scores four data families. Who\&When supplies natural localization failures, while SWE-Gym
and tau-bench supply POST and LIVE runs. Six declared-harness sources supply PRE configurations.
SWE-Gym and tau-bench support the two Gold injection substrates. Natural labels anchor the failure
boards, while injected data isolate known fault mechanisms. Tables~\ref{tab:post-live-corpus-stats}
and~\ref{tab:pre-corpus-stats} give the population rules and label balances.

\subsection{Source Corpora and Populations}
\label{sec:data-corpora}

Who\&When localization uses 126 naturally occurring failures from its Algorithm-Generated split. The
board excludes the 58 Hand-Crafted runs, whose schema and median trace length define a distinct
population, and excludes Who\&When Pro, whose released cards document neither dependency edges nor a
common source-seed identifier, so structural entrants would model edges rather than read them and
repeated injections from one prefix could not be treated as independent
\citep{liu2026whowhenpro}. POST and LIVE use a balanced SWE-Gym population and the full pinned
tau-bench population; file-level Gold uses resolved SWE-Gym runs and named-value Gold v2 uses
tau-bench. Appendix~\ref{app:corpus-composition} gives every selection rule, the counts behind these
exclusions, and the label balances. Those three corpora were found. The next one had to be built, so
how each of its records came by its label is part of the data rather than a detail of it.

\subsection{The PRE Configuration Corpus}
\label{sec:data-pre}

PRE tests whether a static method can flag capabilities that a declared task or role does not need.
Its 1187 records come from crewai, injecagent, mcp, n8n, sweagent, and synthetic configurations under
four explicit label processes: cross-vendor LLM judging for crewai, n8n, and mcp; roster relabeling
for injecagent; declared-minus-used labels from paired sweagent traces; and controlled synthetic
over-grant injection \citep{liu2022bond}. Unused in one observed run is only a proxy for
unneeded and does not prove it. Pooling these processes silently would conflate different forms of
evidence, so every record carries its source and label-process tag. Appendix
Tables~\ref{tab:pre-corpus-stats}, \ref{tab:pre-licences}, and \ref{tab:pre-kappa} report construction,
licences, and agreement.

\subsection{\sysname-Gold: Injection and the Admissibility Bar}
\label{sec:data-gold}
\label{sec:data-goldv2}

\begin{figure}[t]
  \centering
  \includegraphics[width=\textwidth]{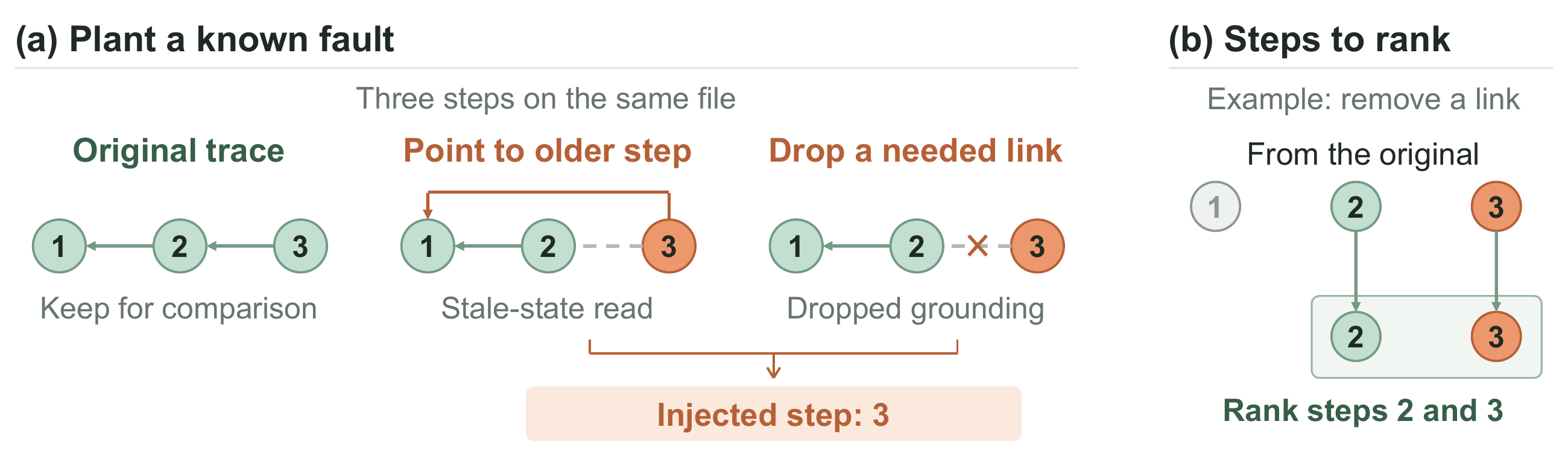}
  \vspace{-0.2in}
  \caption{\textbf{Gold fixes the label when it plants the fault.}
  (a) File-level Gold redirects or removes step 3's dependency; the edited step supplies the
  label independently of any detector. Arrows point to earlier dependencies; dashes mark the
  changed link. The original stays unedited for comparison. (b) Candidates come from the original, so
  step 3 remains eligible after deletion. This controls eligibility, not other construction
  shortcuts. Named-value v2 edits one argument value and rebuilds dependencies from the edited
  record. Both substrates remain mechanism diagnostics.}
  \label{fig:gold-mechanism}
  \vspace{-0.1in}
\end{figure}

Gold plants a known fault in a real run because labeled dependency-state failures do not exist at
scale, following ADBench and BOND \citep{han2022adbench, liu2022bond}, and the injection site
supplies a detector-independent localization and attribution target. Two faults are planted: a
stale-state read redirects a dependency from the latest event on a file to an earlier superseded
event, and dropped grounding removes one required dependency, leaving the step ungrounded.
Figure~\ref{fig:gold-mechanism} shows both edits and which steps a detector ranks in the
matched control.

An injection is admissible only if it clears five tests: (1) the fault maps to a documented mode,
(2) no construction-only baseline detects it trivially, (3) injected and clean runs stay
distributionally comparable, (4) labels are detector-independent, and (5) human validation confirms a
real fault on an airtight substrate. File-level Gold fails test 2 outright. Its edited edge breaks a
clean-graph predecessor invariant, and a construction-only baseline detects that marker perfectly.
Named-value Gold v2 exists to separate an objection to injection itself from an objection to that way
of doing it: it mutates an argument value and rebuilds dependency edges from the values present in a
real tau-bench trace, which removes the marker. The same five-item bar applies, and Gold v2 leaves
tests 1 to 3 undetermined. Both substrates pass test 4, and neither has yet met test 5.
Gold v2's stale-state arm fails corpus adequacy outright, with 16 injectable
sites in 6 runs of the 660-run corpus against 2077 dropped-grounding sites in 614 runs. On dropped
grounding no process control fired and a grounding oracle recovers the injected site at 1.000 Top-1;
what holds the bar at undetermined there is missing positive-control power rather than a failed
check. Both substrates therefore remain mechanism diagnostics and carry no evidential weight anywhere
in the results. Table~\ref{tab:defensibility-verdicts} gives the evidence behind each verdict, and
Appendix~\ref{app:defensibility-verdicts} the construction of both substrates.

\section{Results}
\label{sec:results}


Subsections follow PRE, LIVE, and POST, then detector stability. Gold-derived boards carry no
evidential weight and appear as mechanism diagnostics in Appendix~\ref{sec:res-gold} (cells in
Appendix~\ref{app:board-values}).

\subsection{PRE: Does a Static Harness Audit Beat Flagging Everything?}
\label{sec:res-pre}

This board scores rule-based scanners, a held-out LLM judge, and two trivial policies over 1187
configurations (Section~\ref{sec:data-pre}). The floor is \texttt{flag\_all} rather than zero
because flagging every declared capability has recall 1.000 by construction. Because the large
\texttt{mcp} servers dominate the pooled column, the per-source panels of Figure~\ref{fig:pre-source}
are the primary reading; Table~\ref{tab:pre-source} carries the exact cells.\looseness=-1

\paragraph{The Judge Clears the Scanner and the Scanner Clears the Floor.}
The combined scanner is the union of five token rules, so it keeps most of the floor's recall
(0.910 against 1.000) while lifting precision. Its 0.654 F1 is a gain of 0.053 over
\texttt{flag\_all} at 0.601. Trading further along the same axis, the held-out judge
reaches 0.695 F1 against the scanner's 0.648 on the 1182 configurations both judged, a difference
of 0.048 with marginal 95\% interval $[0.021, 0.074]$. It is held out because the label merge marks
excess only when two other models agree, so each of those models scores recall 1.000 by
construction. A builder who labels a corpus with one model and audits it with the same
model is grading that model's own outputs, and even cross-vendor agreement can reflect shared
model preferences (Appendix~\ref{app:pre-details}).\looseness=-1

\paragraph{Where the Label Follows Declaration Order, a Position Rule Is Perfect.}
\texttt{injecagent}'s harvester writes every roster as the user tool followed by the attacker
tools and preserves that order on release, so a reader of the declaration can read the label. There
the held-out judge reaches 0.990, and 0.972 on injected \texttt{synthetic}, but 0.362 to 0.744 on
the judge-labeled sources. A rule that reads nothing but position scores 1.000 F1 on
\texttt{injecagent} and far lower on \texttt{crewai} and \texttt{n8n}. The leak sits in the corpus rather than the rule (Appendix~\ref{app:pre-details}).

\paragraph{Five Sources Score Above Their Floor, but Two Measure Construction.}
Of the six sources, five place their best candidate above that source's \texttt{flag\_all} floor,
and on \texttt{sweagent} the best candidate is level with it. \texttt{injecagent} and
\texttt{synthetic} still measure construction, and \texttt{mcp}'s margin turns on one large server.
\texttt{crewai} and \texttt{n8n} carry a margin that survives the reported construction checks, on the two judge-labeled
sources with the lowest inter-judge kappa (Table~\ref{tab:pre-kappa}). Two source columns are therefore
evidence about methods, two are checks on construction, and two are readings that one file or one
label rule decides (Appendix~\ref{app:stats}).

\subsection{LIVE: How Early Is Failure Visible?}
\label{sec:res-live}

This board scores dependency-graph prefixes at 25\%, 50\%, 75\%, and 100\% under supervised,
batch-unsupervised (ECOD over prefix features), and per-run online regimes. Prefix fractions are
relative to eventual trace length, so a positive result at 25\% is no clock-time alarm. The 100\%
column reproduces the POST detection scores of Section~\ref{sec:res-post-det} under
cross-validation, as a control.

\begin{figure}[t]
\centering
\includegraphics[width=\textwidth]{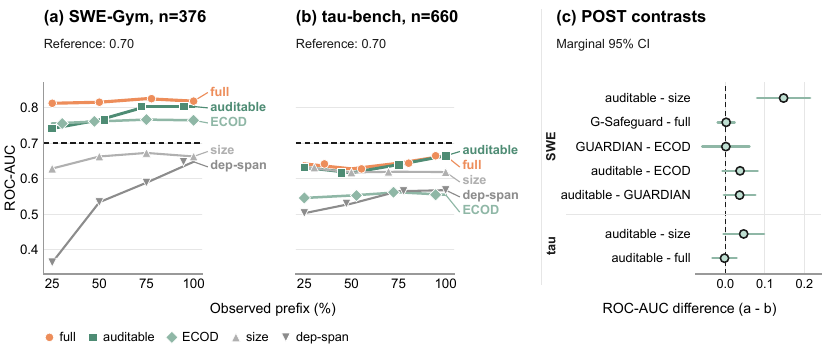}
\caption{\textbf{SWE-Gym point estimates cross 0.70; no tau-bench point estimate
reaches it.} (a,b) LIVE trajectories on shared axes; horizontal offsets separate near-tied markers, and
the dashed rule is the fixed 0.70 bar. (c) POST differences as $a-b$, with marginal 95\%
intervals. Exact cells are in
Tables~\ref{tab:live-stream}, \ref{tab:live-stream-tau} and~\ref{tab:det}.}
\label{fig:detection}
\end{figure}

\paragraph{Flexible Size Control Leaves the Early SWE-Gym Increment Unresolved.}
At 25\%, supervised \texttt{auditable} reaches 0.742 against 0.629 for linear size and counts, a
difference of 0.113 on board cells, and $+0.101$ on the seed-averaged out-of-fold estimand
(Appendix~\ref{app:stats}). Run size is a failure signal on SWE-Gym, where the failed runs are the
shorter ones, so a linear size term is a weak reference (Appendix~\ref{app:board-values}). On the
same out-of-fold estimand, the size-only spline reaches 0.794 at 25\% (Table~\ref{tab:live-size-control}).
Against that flexible reference the matched contrast is $+0.005$, marginal 95\% interval
$[-0.010, +0.020]$, which contains zero and leaves the sign of the early increment unresolved, and
no pre-endpoint matched contrast excludes zero on either corpus. This control was declared after
the linear results were inspected and remains exploratory.

\paragraph{Tau-Bench Remains Below the Warning Bar.}
On tau-bench, no nonrandom entrant's point estimate reaches the bar at any prefix, so no cell
records a time to detection (Table~\ref{tab:live-stream-tau}). Run size carries little on
tau-bench: both size arms read close to each other and below the bar at every prefix
(Table~\ref{tab:live-size-control}). Eighteen
of the twenty bar readings already separate below 0.70 at the present sample; the two that do not
are the full-trace cells of \texttt{auditable} and \texttt{full} (Table~\ref{tab:tau-design}).

\subsection{POST: Which Step Broke the Run?}
\label{sec:res-post-loc}

\begin{figure}[t]
\centering
\includegraphics[width=\columnwidth]{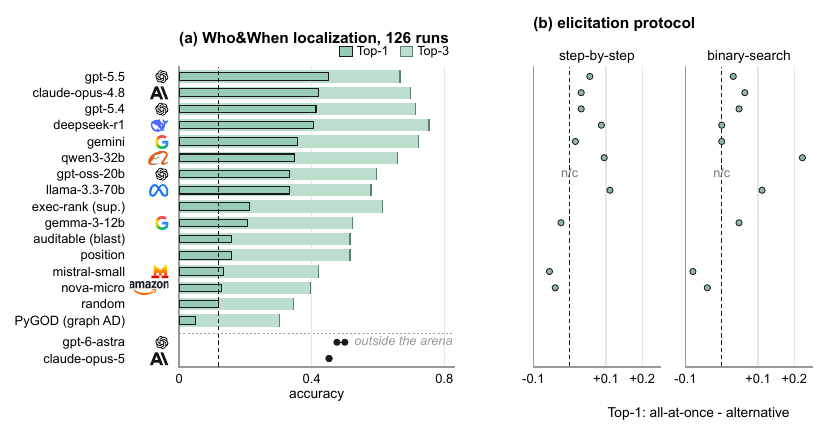}
\vspace{-8pt}
\caption{\textbf{The frozen board's top eight span 0.333 to 0.452 Top-1, and asking the judge
differently rarely helps.} (a) Each bar runs to Top-3 with Top-1 nested inside it, and the dashed
rule is random Top-1. Below the dashed rule, two later releases measured on one fixed channel
outside the arena, one dot per repeated run, in no declared contrast. (b) Top-1 differences,
all-at-once minus the named alternative, on the rows of (a), drawn as points with no interval. The
largest is Qwen3-32B under binary-search. \texttt{n/c}: no cached alternative.}
\label{fig:localization-panel}
\end{figure}

\paragraph{GPT-5.5 Leads the Structural Methods, and Later Releases Still Miss About Half.}
Eleven LLM judges name the decisive mistake step of each failed trace under the Who\&When
all-at-once protocol \citep{zhang2025whoandwhen}, one generation per run with each step capped at
1500 characters. The full-context dependency prior ranks steps exactly as position does; the
judges also read the capped content, so method and input differ (Appendix~\ref{app:board-values}).
GPT-5.5 scores 0.452 Top-1 against 0.211 for the highest structural point
estimate, and eight judges fall in the band drawn in Figure~\ref{fig:localization-panel}. Its
differences against the next three contain zero, so the benchmark cannot order its top. GPT-6 Astra
and Claude Opus 5 sit outside the arena with no declared contrast; on one fixed channel repeated
runs read 0.452 to 0.500 Top-1 (Figure~\ref{fig:localization-panel}, Table~\ref{tab:judge-addendum}). The lead belongs to the model
measured, so a reader cannot substitute another judge for the one scored, and the current
generation still misses the decisive step at Top-1 on about half of these runs. One structural entrant has a resolved contrast:
\texttt{exec-rank (sup.)}, GRADE's execution-feature ranker \citep{zhao2026grade}, clears position
on Top-3 by $+0.098$ $[0.035, 0.167]$, while its Top-1 difference, $+0.052$ $[-0.003, 0.110]$,
contains zero (Table~\ref{tab:all-contrasts}). Execution features therefore give a three-step
shortlist without a language model.

\subsection{POST: Does Dependency Structure Detect Failure Beyond Run Size and Counts?}
\label{sec:res-post-det}

\paragraph{The Structural Gain Depends on the Model Specification.}
We rerun GRADE's linear comparison \citep{zhao2026grade} under flexible size models. On SWE-Gym
the failed runs are the shorter ones (Appendix~\ref{app:board-values}), so a linear size term is a
weaker reference than the data support. The mean-fold matched-spline increment is $-0.010$, against
$+0.142$ linear and $+0.039$ from disclosed boosted trees. On seed-averaged out-of-fold predictions,
the estimand Table~\ref{tab:core-contrasts} and the introduction report, the linear and matched
increments read $+0.147$ and $-0.005$ $[-0.026, +0.015]$. That interval contains zero, which leaves
the increment beyond size unresolved without establishing that it is zero. The two readings differ
in the control rather than in the data; the linear contrast remains correct against its own
reference. Tau-bench's mean-fold increments, $+0.046$ linear, $+0.045$ matched spline, and
$+0.049$ boosted trees, keep their sign as point estimates. A
declared exploratory audit on OpenHands and ScienceWorld is unresolved or modest on its
prespecified criterion (Table~\ref{tab:post-generalization}).\looseness=-1

\paragraph{This Board Resolves Neither Architecture nor Labels.}
The seven registered contrasts of Figure~\ref{fig:detection}(c) separate neither architecture nor
labels. G-Safeguard holds the highest supervised point estimate, 0.828; the
single-seed GAAN entry at 0.850 is seed-unstable (Section~\ref{sec:res-transfer}). Across five joint split and initialization seeds G-Safeguard reaches 0.824 $\pm$ 0.007
against 0.819 $\pm$ 0.005 for the full feature model, a within-seed difference whose interval
contains zero. The board therefore cannot separate message passing over the typed graph from a
linear model on hand-built features. Without labels, ECOD over flat features reaches 0.765 and the
agent-specific GUARDIAN, whose adaptation drops its published adjacency-reconstruction and
bottleneck terms, reads 0.767. That difference, and the contrasts of both against
\texttt{auditable} (size+deps) at 0.804, $+0.036$ and $+0.037$, contain zero
(Appendix~\ref{app:stats}). An off-the-shelf tabular detector on flat features is the floor an
agent-specific unsupervised method has to clear. Here it does not clear it measurably.\looseness=-1

\subsection{Detector Stability}
\label{sec:res-transfer}

\paragraph{No Stable Board Lead Is Established, and the Graph Family Is Seed-Unstable.}
The tabular detectors are deterministic under the recorded settings, while the four PyGOD
detectors train neural models from a random initialization, so their score depends on it. Across twenty
initialization seeds, graph-detector spreads on SWE-Gym run twelve to twenty-two times the
supervised graph network's (Table~\ref{tab:transfer}), and quadrupling the seed count moves DOMINANT's mean by more than 0.10
\citep{ding2019dominant}. A stochastic detector's cell should therefore be reported as its seed
mean, since the single-seed GAAN maximum sits above its own twenty-seed mean. On tau-bench, where
size carries little, the graph family remains at chance.

Online detection of injected stale reads recovers little signal. Redirecting an edge preserves
every step's edge count, so the dependency-count control is signal-free by construction, and the
causal span z-score falls below it at the lower false-positive rate. Because file-level Gold does not meet the artifact
bar (Section~\ref{sec:data-gold}), Appendix~\ref{app:board-values} reports that diagnostic.

\section{Conclusion and Future Directions}
\label{sec:discussion}
\label{sec:conclusion}

\sysname puts one auditor's question to three information states, PRE, LIVE, and POST, under one
task-method interface. Nine scored boards and 72 entrants make up the arena, and every ordering
regenerates from released predictions with no model call. That decomposition, rather than any single
result, is what the benchmark adds. The LIVE prefix
sweep holds corpus, labels, metric, and entrants fixed, and is the one place where evidence alone
varies. Where the evidence is unresolved, the paper says so: the matched size control leaves the
SWE-Gym dependency increment unresolved at every prefix, and the localization board cannot order
its leading judges.

Two procedures generalize beyond it. First, run a declaration-order check. A rule that
keeps the first capability and flags the rest, reading no name or permission, scores 1.000 F1 on one
PRE source and is perfect on no other. That per-source pattern can measure corpus construction
rather than reasoning. Second, fit a flexible size control before crediting structural features. On
SWE-Gym the seed-averaged out-of-fold dependency increment reads $+0.147$ under a linear size
reference and $-0.005$ under a matched spline; the reference matters because run size itself
predicts failure there.

Measured headroom points to five extensions. \texttt{crewai} and \texttt{n8n} have the lowest
inter-judge kappa of the three judge-labeled PRE sources, yet their margin over the floor survives
the reported construction checks. A human-audited slice with reported inter-rater agreement would show whether that margin is
method signal or label noise. On tau-bench no point estimate reaches the 0.70 warning bar at any
prefix. Holding the observed effects fixed, measured scaling gives design targets of about 410 clusters
for \texttt{full} and 434 for \texttt{auditable} at 100\% (Table~\ref{tab:tau-design}). Dropped
grounding and online stale-state detection stay open on the Gold-derived substrate, which has not
met the human-validation bar. Whether AppWorld cross-app writes supply a substrate that resists a
one-line construction check is the next release's test. A localization corpus with recorded rather
than assumed dependency edges would let a dependency-based method be judged on its own signal. The
unsupervised graph family needs seed-stable detectors before its mean is operationally meaningful,
and a twenty-seed report per entrant is the evidence that would show it. \sysname is the arena
these extensions plug into, under the same task-method interface and admissibility bar, with
every ordering regenerated from committed predictions.



\clearpage
\ifcatchbenchpreprint\endgroup\fi
\newpage

\section*{Reproducibility Statement}
\label{sec:reproducibility}

\paragraph{Artifact.} The \sysname code, the committed PRE derived records, the cached LLM-judge
predictions, and the diagnostic scripts form the artifact, which is public
\ificlrfinal{} at \url{https://github.com/yzhao062/catchbench}\else{} as a supplement prepared for review\fi.
It does not redistribute the raw Who\&When, SWE-Gym, or tau-bench traces, which remain at their
cited upstream sources \citep{zhang2025whoandwhen,pan2024swegym,yao2024taubench}. The current PRE
release contains derived records for all 1187 configurations with the raw task and role field
omitted. Earlier public revisions included raw task and role prose for all 1187 configurations,
which remains accessible through repository history. The historical identifier exposure remains
unresolved.

The scored board calls no model service: it reads committed prediction caches, so a third party
reproduces every board value in this paper without an API key. The one exception is that the checker permits an absolute
difference of up to 0.005 for G-Safeguard and AnomalyDAE on each detection corpus (four cells), and
compares all other board cells exactly. All three trace corpora are pinned by
commit and the pins are checked before scoring. Appendix~\ref{app:artifact} states what is pinned,
what regenerating the cached predictions would additionally require, and what continuous integration
checks on every commit. Table~\ref{tab:revision-record} carries the revisions and what is left
unpinned.

\section*{Ethics and Data Statement}
\label{sec:ethics-data}

\sysname processes public agent traces, public software repositories, public MCP manifests,
public n8n template-gallery entries, and authored synthetic configurations. The PRE harvesters
initially collected each public task or role description together with its declared capability roster
and source record. Template authors sometimes included contact details in their own descriptions, so
the released records exclude that prose. \texttt{task\_or\_role\_spec} was removed from every PRE
row before release, and replaced by sorted \texttt{spec\_tokens} plus small
\texttt{spec\_token\_overrides} summaries sufficient for the shipped scanner. The conversion first
strips email addresses, URLs, home-directory paths, social handles, and phone numbers, then keeps only
scanner vocabulary and terms tied to the declared capability roster. All 1187 committed rows carry
\texttt{spec\_tokens} and none carries \texttt{task\_or\_role\_spec}.\looseness=-1

The released PII scan, run over the distributed records, reports zero
occurrences and zero distinct values for all seven patterns it checks: email addresses, LinkedIn
profile URLs, other profile URLs, Unix and Windows home paths, social handles, and phone numbers. That
is the output of seven regular expressions, not a guarantee that no indirect identifier survives. The
released records deliberately retain source repository, commit, path, capability names, labels, and
scanner features, because an auditing benchmark whose own sources cannot be checked is worth little. A
reader therefore cannot reconstruct the deleted prose from the released fields, but can follow the
retained references back to the public upstream material. Researchers should respect each source
licence and should not use this corpus to contact, profile, or rank individual template authors.

One redaction is recorded against the released prediction cache. On the nine PRE records declaring
GPL-3.0, spans of the cached judge reply that reproduce upstream prose are replaced by a marker. The
judge's own reasoning and its verdict stay in place, and no scored number moves, because the board
reads the published needed-list cache rather than the reply text.

\section*{AI Use Statement}
\label{sec:ai-use}

Large language models appear in two roles in this work. The distinction between these roles matters
when interpreting the results.

They are first of all objects of measurement. The eleven-model panel in
Section~\ref{sec:res-post-loc} is a baseline being scored. Each model is prompted once over a failed
trace under the Who\&When all-at-once protocol, and its answer is compared against human
decisive-step labels. On the PRE board (Section~\ref{sec:data-pre}), two models construct the
excess-capability key for 661 of the 1187 records, and a third is held out as a baseline.
Predictions from both roles are cached and committed, so the board rescores them without a model
call and a reader can inspect the exact outputs behind every number.

The division of labor is worth stating exactly, because ``no model was involved'' would be false here
and ``a model produced the results'' would be worse. Models supplied cached predictions and
annotations: the 31 committed POST prediction caches that the boards score, the two PRE judges
that build the excess-capability key, and the held-out PRE judge. Code with no model call computed
every metric and every test over those committed artifacts. No model chose which result to report,
ranked a method, or edited a number.

Separately from the experiments, the authors used AI coding and writing assistants while developing
the benchmark code, the analysis tooling, and the prose of this paper. Every claim, number, and
citation was checked by the authors against the committed records before submission, and the authors
take full responsibility for the content.

\bibliographystyle{iclr2027_conference}
\bibliography{references}

\begin{thebibliography}{38}
\providecommand{\natexlab}[1]{#1}
\providecommand{\url}[1]{\texttt{#1}}
\expandafter\ifx\csname urlstyle\endcsname\relax
  \providecommand{\doi}[1]{doi: #1}\else
  \providecommand{\doi}{doi: \begingroup \urlstyle{rm}\Url}\fi

\bibitem[Balusu(2026)]{balusu2026agenttelemetry}
Krishna~Chaitanya Balusu.
\newblock {AgentTelemetry}: A fault detection benchmark and toolkit for {LLM}
  agent observability.
\newblock In \emph{Proceedings of the 3rd ACM International Conference on
  AI-Powered Software}, pp.\  380--387. ACM, 2026.
\newblock \doi{10.1145/3805760.3814931}.

\bibitem[Barke et~al.(2026)Barke, Goyal, Khare, Singh, Nath, and
  Bansal]{barke2026agentrx}
Shraddha Barke, Arnav Goyal, Alind Khare, Avaljot Singh, Suman Nath, and Chetan
  Bansal.
\newblock {AgentRx}: Diagnosing {AI} agent failures from execution
  trajectories.
\newblock \emph{arXiv preprint arXiv:2602.02475}, 2026.
\newblock URL \url{https://arxiv.org/abs/2602.02475}.
\newblock Findings of the Association for Computational Linguistics: EMNLP
  2026.

\bibitem[Cemri et~al.(2025)Cemri, Pan, Yang, Agrawal, Chopra, Tiwari, Keutzer,
  Parameswaran, Klein, Ramchandran, Zaharia, Gonzalez, and
  Stoica]{cemri2025mast}
Mert Cemri, Melissa~Z. Pan, Shuyi Yang, Lakshya~A. Agrawal, Bhavya Chopra,
  Rishabh Tiwari, Kurt Keutzer, Aditya Parameswaran, Dan Klein, Kannan
  Ramchandran, Matei Zaharia, Joseph~E. Gonzalez, and Ion Stoica.
\newblock Why do multi-agent {LLM} systems fail?
\newblock In \emph{Advances in Neural Information Processing Systems}, 2025.
\newblock URL \url{https://openreview.net/forum?id=fAjbYBmonr}.
\newblock Datasets and Benchmarks Track.

\bibitem[Chen et~al.(2026)Chen, Wang, Mu, Wang, Liu, Feng, and
  Wang]{chen2026traceelephant}
Mengzhuo Chen, Junjie Wang, Fangwen Mu, Yawen Wang, Zhe Liu, Huanxiang Feng,
  and Qing Wang.
\newblock Seeing the whole elephant: A benchmark for failure attribution in
  {LLM}-based multi-agent systems.
\newblock In \emph{Proceedings of the 64th Annual Meeting of the Association
  for Computational Linguistics (Volume 1: Long Papers)}, pp.\  19888--19905,
  2026.
\newblock \doi{10.18653/v1/2026.acl-long.912}.
\newblock URL \url{https://aclanthology.org/2026.acl-long.912/}.

\bibitem[Deshpande et~al.(2025)Deshpande, Gangal, Mehta, Krishnan, Kannappan,
  and Qian]{deshpande2025trail}
Darshan Deshpande, Varun Gangal, Hersh Mehta, Jitin Krishnan, Anand Kannappan,
  and Rebecca Qian.
\newblock {TRAIL}: Trace reasoning and agentic issue localization.
\newblock \emph{arXiv preprint arXiv:2505.08638}, 2025.

\bibitem[Ding et~al.(2019)Ding, Li, Bhanushali, and Liu]{ding2019dominant}
Kaize Ding, Jundong Li, Rohit Bhanushali, and Huan Liu.
\newblock Deep anomaly detection on attributed networks.
\newblock In \emph{Proceedings of the SIAM International Conference on Data
  Mining (SDM)}, pp.\  594--602, 2019.

\bibitem[Fan et~al.(2026)Fan, Ye, Huo, Chen, Guo, Yang, Yang, Ye, Chen, Chen,
  Cong, and Lin]{fan2026agentprocessbench}
Shengda Fan, Xuyan Ye, Yupeng Huo, Zhi-Yuan Chen, Yiju Guo, Shenzhi Yang,
  Wenkai Yang, Shuqi Ye, Jingwen Chen, Haotian Chen, Xin Cong, and Yankai Lin.
\newblock {AgentProcessBench}: Diagnosing step-level process quality in
  tool-using agents.
\newblock In \emph{Proceedings of the 32nd ACM SIGKDD Conference on Knowledge
  Discovery and Data Mining V.2}, pp.\  8823--8834, 2026.
\newblock \doi{10.1145/3770855.3817494}.
\newblock URL \url{https://arxiv.org/abs/2603.14465}.

\bibitem[Han et~al.(2022)Han, Hu, Huang, Jiang, and Zhao]{han2022adbench}
Songqiao Han, Xiyang Hu, Hailiang Huang, Minqi Jiang, and Yue Zhao.
\newblock {ADBench}: Anomaly detection benchmark.
\newblock \emph{Advances in Neural Information Processing Systems},
  35:\penalty0 32142--32159, 2022.

\bibitem[He et~al.(2025)He, Wu, Zhai, and Sun]{he2025sentinelagent}
Xu~He, Di~Wu, Yan Zhai, and Kun Sun.
\newblock {SentinelAgent}: Graph-based anomaly detection in multi-agent
  systems.
\newblock \emph{arXiv preprint arXiv:2505.24201}, 2025.

\bibitem[Li et~al.(2026)Li, Yu, Wang, Yang, Rossi, Dernoncourt, Hu, Yu, Xiao,
  Zhang, and Zhao]{li2026fortis}
Shawn Li, Chenxiao Yu, Han Wang, Wei Yang, Ryan Rossi, Franck Dernoncourt,
  Xiyang Hu, Philip~S Yu, Chaowei Xiao, Huan Zhang, and Yue Zhao.
\newblock {FORTIS}: Benchmarking over-privilege in agent skills.
\newblock \emph{arXiv preprint arXiv:2605.09163}, 2026.

\bibitem[Li et~al.(2023)Li, Zhao, Hu, Botta, Ionescu, and Chen]{li2023ecod}
Zheng Li, Yue Zhao, Xiyang Hu, Nicola Botta, Cezar Ionescu, and George~H. Chen.
\newblock {ECOD}: Unsupervised outlier detection using empirical cumulative
  distribution functions.
\newblock \emph{IEEE Transactions on Knowledge and Data Engineering},
  35\penalty0 (12):\penalty0 12181--12193, 2023.

\bibitem[Liu et~al.(2026)Liu, Xi, Zhang, Zeng, Yue, Wang, Kang, Wu, and
  Wang]{liu2026whowhenpro}
Jiale Liu, Huajun Xi, Shaokun Zhang, Yifan Zeng, Tianwei Yue, Chi Wang, Jian
  Kang, Qingyun Wu, and Huazheng Wang.
\newblock {Who\&When Pro}: Can {LLM}s really attribute failures in {AI}
  agents?, 2026.
\newblock URL \url{https://arxiv.org/abs/2607.09996}.

\bibitem[Liu et~al.(2022)Liu, Dou, Zhao, Ding, Hu, Zhang, Ding, Chen, Peng,
  Shu, Sun, Li, Chen, Jia, and Yu]{liu2022bond}
Kay Liu, Yingtong Dou, Yue Zhao, Xueying Ding, Xiyang Hu, Ruitong Zhang, Kaize
  Ding, Canyu Chen, Hao Peng, Kai Shu, Lichao Sun, Jundong Li, George~H Chen,
  Zhihao Jia, and Philip~S Yu.
\newblock {BOND}: Benchmarking unsupervised outlier node detection on static
  attributed graphs.
\newblock In \emph{Advances in Neural Information Processing Systems},
  volume~35, pp.\  27021--27035, 2022.

\bibitem[L{\`u} et~al.(2025)L{\`u}, Kazemnejad, Meade, Patel, Shin, Zambrano,
  Sta{\'n}czak, Shaw, Pal, and Reddy]{lu2025agentrewardbench}
Xing~Han L{\`u}, Amirhossein Kazemnejad, Nicholas Meade, Arkil Patel, Dongchan
  Shin, Alejandra Zambrano, Karolina Sta{\'n}czak, Peter Shaw, Christopher~J.
  Pal, and Siva Reddy.
\newblock {AgentRewardBench}: Evaluating automatic evaluations of web agent
  trajectories.
\newblock In \emph{Conference on Language Modeling (COLM)}, 2025.
\newblock URL \url{https://arxiv.org/abs/2504.08942}.

\bibitem[Mateo-Torrej\'on \& S\'anchez-Maci\'an(2026)Mateo-Torrej\'on and
  S\'anchez-Maci\'an]{mateotorrejon2026gammaf}
Pablo Mateo-Torrej\'on and Alfonso S\'anchez-Maci\'an.
\newblock {GAMMAF}: A common framework for graph-based anomaly monitoring
  benchmarking in {LLM} multi-agent systems.
\newblock \emph{arXiv preprint arXiv:2604.24477}, 2026.

\bibitem[Nian et~al.(2026)Nian, Yuan, Zhang, Li, Li, Hu, Wei, Xiao, Xiao, and
  Zhao]{nian2026auditable}
Yi~Nian, Aojie Yuan, Haiyue Zhang, Jiate Li, Li~Li, Xiyang Hu, Hua Wei, Xiongye
  Xiao, Chaowei Xiao, and Yue Zhao.
\newblock Auditable agents.
\newblock \emph{arXiv preprint arXiv:2604.05485}, 2026.

\bibitem[Ou et~al.(2025)Ou, Guo, Gandhi, Neubig, and Yue]{ou2025agentdiagnose}
Tianyue Ou, Wanyao Guo, Apurva Gandhi, Graham Neubig, and Xiang Yue.
\newblock {AgentDiagnose}: An open toolkit for diagnosing {LLM} agent
  trajectories.
\newblock In \emph{Proceedings of the 2025 Conference on Empirical Methods in
  Natural Language Processing: System Demonstrations}, pp.\  207--215.
  Association for Computational Linguistics, 2025.
\newblock \doi{10.18653/v1/2025.emnlp-demos.15}.
\newblock URL \url{https://aclanthology.org/2025.emnlp-demos.15/}.

\bibitem[Palumbo et~al.(2026)Palumbo, Choudhary, Choi, Amir, Chalasani, and
  Jha]{palumbo2026formal}
Nils Palumbo, Sarthak Choudhary, Jihye Choi, Guy Amir, Prasad Chalasani, and
  Somesh Jha.
\newblock Formal policy enforcement for real-world agentic systems.
\newblock \emph{arXiv preprint arXiv:2602.16708}, 2026.

\bibitem[Pan et~al.(2025)Pan, Wang, Neubig, Jaitly, Ji, Suhr, and
  Zhang]{pan2024swegym}
Jiayi Pan, Xingyao Wang, Graham Neubig, Navdeep Jaitly, Heng Ji, Alane Suhr,
  and Yizhe Zhang.
\newblock Training software engineering agents and verifiers with {SWE-Gym}.
\newblock In \emph{Proceedings of the 42nd International Conference on Machine
  Learning}, volume 267 of \emph{Proceedings of Machine Learning Research},
  pp.\  47717--47737, 2025.

\bibitem[Wang et~al.(2026)Wang, Yuan, Zhang, Hu, Zhao, and
  Jiang]{wang2026weclawarena}
Prince~Zizhuang Wang, Aojie Yuan, Haiyue Zhang, Xiyang Hu, Yue Zhao, and Shuli
  Jiang.
\newblock {WeClawArena}: An auditable sandbox and benchmark for cross-user
  agents collaboration and security in human-centered agent networks.
\newblock \emph{arXiv preprint arXiv:2608.03499}, 2026.

\bibitem[Wang et~al.(2025)Wang, Zhang, Yu, Wan, Meng, Guo, Wang, and
  Wang]{wang2025gsafeguard}
Shilong Wang, Guibin Zhang, Miao Yu, Guancheng Wan, Fanci Meng, Chongye Guo,
  Kun Wang, and Yang Wang.
\newblock {G-Safeguard}: A topology-guided security lens and treatment on
  {LLM}-based multi-agent systems.
\newblock In \emph{Proceedings of the 63rd Annual Meeting of the Association
  for Computational Linguistics (Volume 1: Long Papers)}, pp.\  7261--7276.
  Association for Computational Linguistics, 2025.
\newblock \doi{10.18653/v1/2025.acl-long.359}.

\bibitem[Yan et~al.(2026)Yan, Weng, Chen, Peng, Qin, Guan, Liu, Yu, Yuan, Meng,
  Che, and Hu]{yan2026authbench}
Zheng Yan, Jingxiang Weng, Charles Chen, Dengyun Peng, Ethan Qin, Jiannan Guan,
  Jinhao Liu, Qiming Yu, Yixin Yuan, Fanqing Meng, Carl Che, and Mengkang Hu.
\newblock Do coding agents understand least-privilege authorization?
\newblock \emph{arXiv preprint arXiv:2605.14859}, 2026.

\bibitem[Yang et~al.(2024)Yang, Jimenez, Wettig, Lieret, Yao, Narasimhan, and
  Press]{yang2024sweagent}
John Yang, Carlos~E. Jimenez, Alexander Wettig, Kilian Lieret, Shunyu Yao,
  Karthik Narasimhan, and Ofir Press.
\newblock {SWE-agent}: Agent-computer interfaces enable automated software
  engineering.
\newblock In \emph{Advances in Neural Information Processing Systems}, 2024.

\bibitem[Yang et~al.(2026)Yang, Bu, Yi, Wang, Zhou, Dai, Hu, and
  Yang]{yang2026toolprivbench}
Kaiyue Yang, Yuyan Bu, Jingwei Yi, Yuchi Wang, Biyu Zhou, Juntao Dai, Songlin
  Hu, and Yaodong Yang.
\newblock When lower privileges suffice: Investigating over-privileged tool
  selection in {LLM} agents.
\newblock \emph{arXiv preprint arXiv:2606.20023}, 2026.

\bibitem[Yao et~al.(2025)Yao, Shinn, Razavi, and Narasimhan]{yao2024taubench}
Shunyu Yao, Noah Shinn, Pedram Razavi, and Karthik Narasimhan.
\newblock {$\tau$-bench}: A benchmark for tool-agent-user interaction in
  real-world domains.
\newblock In \emph{International Conference on Learning Representations},
  volume 2025, pp.\  9965--10017, 2025.

\bibitem[Yuan et~al.(2024)Yuan, He, Dong, Wang, Zhao, Xia, Xu, Zhou, Li, Zhang,
  Wang, and Liu]{yuan2024rjudge}
Tongxin Yuan, Zhiwei He, Lingzhong Dong, Yiming Wang, Ruijie Zhao, Tian Xia,
  Lizhen Xu, Binglin Zhou, Fangqi Li, Zhuosheng Zhang, Rui Wang, and Gongshen
  Liu.
\newblock {R}-judge: Benchmarking safety risk awareness for {LLM} agents.
\newblock In \emph{Findings of the Association for Computational Linguistics:
  EMNLP 2024}, pp.\  1467--1490. Association for Computational Linguistics,
  2024.
\newblock \doi{10.18653/v1/2024.findings-emnlp.79}.
\newblock URL \url{https://aclanthology.org/2024.findings-emnlp.79/}.

\bibitem[Zhan et~al.(2024)Zhan, Liang, Ying, and Kang]{zhan2024injecagent}
Qiusi Zhan, Zhixiang Liang, Zifan Ying, and Daniel Kang.
\newblock {InjecAgent}: Benchmarking indirect prompt injections in
  tool-integrated large language model agents.
\newblock In \emph{Findings of the Association for Computational Linguistics:
  ACL 2024}, 2024.

\bibitem[Zhang et~al.(2025{\natexlab{a}})Zhang, Wang, Chen, Zhou, Wang, and
  Yan]{zhang2025agentracer}
Guibin Zhang, Junhao Wang, Junjie Chen, Wangchunshu Zhou, Kun Wang, and
  Shuicheng Yan.
\newblock {AgenTracer}: Who is inducing failure in the {LLM} agentic systems?
\newblock \emph{arXiv preprint arXiv:2509.03312}, 2025{\natexlab{a}}.
\newblock URL \url{https://arxiv.org/abs/2509.03312}.

\bibitem[Zhang et~al.(2025{\natexlab{b}})Zhang, Huang, Mei, Yao, Wang, Zhan,
  Wang, and Zhang]{zhang2025agentsecuritybench}
Hanrong Zhang, Jingyuan Huang, Kai Mei, Yifei Yao, Zhenting Wang, Chenlu Zhan,
  Hongwei Wang, and Yongfeng Zhang.
\newblock Agent security bench ({ASB}): Formalizing and benchmarking attacks
  and defenses in {LLM}-based agents.
\newblock In \emph{International Conference on Learning Representations},
  2025{\natexlab{b}}.
\newblock URL \url{https://openreview.net/forum?id=V4y0CpX4hK}.

\bibitem[Zhang et~al.(2025{\natexlab{c}})Zhang, Yin, Zhang, Liu, Han, Zhang,
  Li, Wang, Wang, Chen, and Wu]{zhang2025whoandwhen}
Shaokun Zhang, Ming Yin, Jieyu Zhang, Jiale Liu, Zhiguang Han, Jingyang Zhang,
  Beibin Li, Chi Wang, Huazheng Wang, Yiran Chen, and Qingyun Wu.
\newblock Which agent causes task failures and when? on automated failure
  attribution of {LLM} multi-agent systems.
\newblock In \emph{Proceedings of the 42nd International Conference on Machine
  Learning}, volume 267 of \emph{Proceedings of Machine Learning Research},
  pp.\  76583--76599, 2025{\natexlab{c}}.

\bibitem[Zhang et~al.(2026)Zhang, Zhu, Liu, Chen, Ma, Liu, Zhang, Li, Tang, Ji,
  and You]{zhang2026cuadebug}
Weijia Zhang, Kunlun Zhu, Zeyi Liu, Yinting Chen, Tianyi Ma, Jiateng Liu,
  Jiaxun Zhang, Bingxuan Li, Xiangru Tang, Heng Ji, and Jiaxuan You.
\newblock {CUADebug}: Diagnosing and repairing computer-use agent failures.
\newblock \emph{arXiv preprint arXiv:2608.02643}, 2026.
\newblock URL \url{https://arxiv.org/abs/2608.02643}.

\bibitem[Zhao(2026)]{zhao2026grade}
Yue Zhao.
\newblock {GRADE}: Graph representation of {LLM} agent dependency and
  execution.
\newblock \emph{arXiv preprint arXiv:2606.22741}, 2026.

\bibitem[Zhao et~al.(2019)Zhao, Nasrullah, and Li]{zhao2019pyod}
Yue Zhao, Zain Nasrullah, and Zheng Li.
\newblock {PyOD}: A {Python} toolbox for scalable outlier detection.
\newblock \emph{Journal of Machine Learning Research}, 20\penalty0
  (96):\penalty0 1--7, 2019.

\bibitem[Zheng et~al.(2025)Zheng, Zhang, Zhang, Lin, Lu, Yu, Liu, Zhou, and
  Lin]{zheng2025processbench}
Chujie Zheng, Zhenru Zhang, Beichen Zhang, Runji Lin, Keming Lu, Bowen Yu,
  Dayiheng Liu, Jingren Zhou, and Junyang Lin.
\newblock {ProcessBench}: Identifying process errors in mathematical reasoning.
\newblock In \emph{Proceedings of the 63rd Annual Meeting of the Association
  for Computational Linguistics (Volume 1: Long Papers)}, pp.\  1009--1024,
  2025.
\newblock \doi{10.18653/v1/2025.acl-long.50}.
\newblock URL \url{https://arxiv.org/abs/2412.06559}.

\bibitem[Zhou et~al.(2025)Zhou, Wang, and Yang]{zhou2025guardian}
Jialong Zhou, Lichao Wang, and Xiao Yang.
\newblock {GUARDIAN}: Safeguarding {LLM} multi-agent collaborations with
  temporal graph modeling.
\newblock In \emph{Advances in Neural Information Processing Systems},
  volume~38, pp.\  7973--8001, 2025.
\newblock \doi{10.52202/085713-0272}.

\bibitem[Zhou et~al.(2026)Zhou, Liu, Li, Rossi, and Hu]{zhou2026counterfactual}
Xiaolin Zhou, Jinbo Liu, Li~Li, Ryan~A. Rossi, and Xiyang Hu.
\newblock Counterfactual trace auditing of {LLM} agent skills.
\newblock \emph{arXiv preprint arXiv:2605.11946}, 2026.

\bibitem[Zhu et~al.(2025)Zhu, Liu, Li, Tian, Yang, Zhang, Han, Xie, Cui, Zhang,
  Ma, Yu, Ramesh, Wu, Liu, Lu, Zou, and You]{zhu2025agentdebug}
Kunlun Zhu, Zijia Liu, Bingxuan Li, Muxin Tian, Yingxuan Yang, Jiaxun Zhang,
  Pengrui Han, Qipeng Xie, Fuyang Cui, Weijia Zhang, Xiaoteng Ma, Xiaodong Yu,
  Gowtham Ramesh, Jialian Wu, Zicheng Liu, Pan Lu, James Zou, and Jiaxuan You.
\newblock Where {LLM} agents fail and how they can learn from failures.
\newblock \emph{arXiv preprint arXiv:2509.25370}, 2025.
\newblock URL \url{https://arxiv.org/abs/2509.25370}.

\bibitem[Zhu et~al.(2026)Zhu, Ye, Han, Zhao, Li, Zhang, Tian, Tang, Lu, Zou,
  You, and Ji]{zhu2026agentdebugx}
Kunlun Zhu, Xuyan Ye, Zhiguang Han, Yuchen Zhao, Bingxuan Li, Weijia Zhang,
  Muxin Tian, Xiangru Tang, Pan Lu, James Zou, Jiaxuan You, and Heng Ji.
\newblock {AgentDebugX}: An open-source toolkit for failure observability,
  attribution, and recovery in {LLM} agents.
\newblock \emph{arXiv preprint arXiv:2607.18754}, 2026.
\newblock URL \url{https://arxiv.org/abs/2607.18754}.

\end{thebibliography}

\newpage

\appendix
\raggedbottom
\makeatletter
\setlength{\@fptop}{0pt}
\setlength{\@fpsep}{12pt plus 2pt minus 2pt}
\setlength{\@fpbot}{0pt plus 1fil}
\setlength{\@dblfptop}{0pt}
\setlength{\@dblfpsep}{12pt plus 2pt minus 2pt}
\setlength{\@dblfpbot}{0pt plus 1fil}
\makeatother

\section{Extended Related Work}
\label{app:related-extended}

Section~\ref{sec:related} places \sysname among the benchmarks that score an external auditor and
states the distinction in one paragraph. This appendix gives the per-work mechanics behind that
placement. It also covers adjacent systems that Table~\ref{tab:positioning} leaves out, including the
tracing, static-analysis, and graph-security work that scores no external auditor.

\subsection{Benchmarks for Anomaly and Graph Anomaly Detection}

\sysname takes its construction protocol from this line, not only its comparison format. Because
labeled anomalies are scarce in real graphs, BOND injects structural and contextual anomalies into
real data and reports the construction \citep{liu2022bond}. \sysname-Gold follows that precedent for agent
runs: Section~\ref{sec:data-gold} plants a known fault in a real trace and releases its
site as a detector-independent target.

\subsection{Agent Safety, Security, and Observability Benchmarks}

The two nearest safety benchmarks differ from \sysname in contract rather than in ambition. R-Judge
is a POST risk-judgment task over human-annotated interaction records \citep{yuan2024rjudge}. Agent
Security Bench instead executes attacks and defenses across tool-using agent scenarios and scores
both security and utility \citep{zhang2025agentsecuritybench}. It does not score PRE over-privilege,
LIVE failure prediction, or POST benign-failure localization on a common dataset. WeClawArena
is the closest security-framed neighbor on the collaboration axis. It runs an auditable sandbox
for multi-party agent collaboration over personal workspaces, expands 124 base tasks into 620
scenario variants, one benign control and four attack-vector variants per base task, and audits
attack success from bounded runtime evidence \citep{wang2026weclawarena}. Its comparison is fixed
across attack vectors, while \sysname compares methods at three information states, each on its own
labels and metric.

AgentTelemetry is a closer observability-centered neighbor. It injects fourteen fault types and
compares five telemetry conditions to test which faults a span vocabulary can expose
\citep{balusu2026agenttelemetry}. Its unit of comparison is the telemetry schema and its injected
fault-detection coverage. \sysname instead fixes what a score means within a state, then compares
methods at three information states.\looseness=-1

\subsection{Agent Failure Attribution}

Who\&When provides the human-verified agent and decisive-step labels used as the anchor for this
paper's localization score \citep{zhang2025whoandwhen}. Its all-at-once, step-by-step, and
binary-search prompts also establish direct LLM-judge controls. Who\&When Pro is the strongest
direct POST-side neighbor. It scales decisive-error attribution to 12{,}326 controlled-injection
trajectories from 26 source benchmarks across text, image, and video, with responsible-agent,
decisive-step, and failure-mode labels \citep{liu2026whowhenpro}. This evaluation begins from a
failed trajectory and evaluates attribution, so it defines no PRE harness audit or LIVE prediction
under a false-alarm budget. TRAIL supplies 148 human-annotated agent traces carrying 841 errors
across GAIA and SWE-Bench Lite, scored by category F1, location accuracy, and joint accuracy
\citep{deshpande2025trail}. Every annotation there sits inside one completed trace, so PRE
configuration records and streaming prefixes fall outside its scope.\looseness=-1

AgentDebug pairs a modular error taxonomy with AgentErrorBench, 200 expert-annotated failure
trajectories from ALFWorld, GAIA, and WebShop \citep{zhu2025agentdebug}. Its scored diagnosis reads
complete failed trajectories, which supports the POST-only row in Table~\ref{tab:positioning}.
Recovery reruns from an identified step, but the paper reports no growing-prefix diagnosis score.
AgentDebugX is a debugging toolkit and method, rather than a new benchmark. It evaluates DeepDebug
on all 184 Who\&When traces and reports strict joint agent-and-step attribution
\citep{zhu2026agentdebugx}. A separate row would therefore duplicate the Who\&When substrate under
a method name. Its strict metric requires both the agent and exact step. Our Top-1 board scores step
localization alone on the 126-run Algorithm-Generated split, so the values do not compare directly.
CUADebug extends this line to computer-use agents through CUAErrorBench, 204 human-annotated failed
OSWorld trajectories \citep{zhang2026cuadebug}. Its evaluated debugger inspects completed traces,
including paired screenshots and action records, so CUAErrorBench is also POST-only.
Other recent methods deepen this POST task. AgenTracer constructs counterfactual and injected
training traces and trains a dedicated localizer, while AgentRx couples 170 human-annotated failed
traces to constraint-based step diagnosis \citep{zhang2025agentracer,barke2026agentrx}. MAST assigns
failure taxonomies over complete traces, and TraceElephant measures attribution with fuller
execution visibility \citep{cemri2025mast,chen2026traceelephant}. A frontier LLM judge holds the
highest Top-1 point estimate on this paper's own localization board, inside a band of eight judges
at $n = 126$, placing \sysname inside this line.\looseness=-1

Table~\ref{tab:positioning} already carries eight complete-trace rows, so three further
benchmarks that score auditors of finished work are described here rather than added with the same
cell pattern. AgentRewardBench evaluates twelve LLM judges against expert review of 1302 completed
web trajectories from five benchmarks \citep{lu2025agentrewardbench}. What a judge reads varies,
from observation summaries to action histories with a final screenshot or accessibility tree, and
the scored output is a success judgment, together with side effects and repetition.
AgentProcessBench presents the task and the complete tool-use trajectory, and labels 8509 assistant
steps across 1000 trajectories as correct, neutral, or incorrect at 89.1 percent inter-annotator
agreement \citep{fan2026agentprocessbench}. It reports both step-label accuracy and the first
incorrect step of each trajectory. ProcessBench presents a mathematical problem and a complete
solution, and its label is the earliest erroneous step, or no error \citep{zheng2025processbench}.
These targets overlap with individual POST tasks here. Our detection boards also score a
success-or-failure judgment, and a first-error index is close to what our localization board asks.
The contracts differ in the label scored: our localization board scores agreement with Who\&When's
human decisive-step annotation, and a run may hold other poor steps besides the one marked
decisive.

\subsection{Graph Anomaly Detection for Agents}

Agent-specific graph anomaly detection in the cited line takes a security framing. G-Safeguard
detects injected adversarial agents from interaction topology, GUARDIAN reconstructs temporal
collaboration graphs, and SentinelAgent combines dynamic execution graphs with LLM oversight
\citep{wang2025gsafeguard,zhou2025guardian,he2025sentinelagent}. GAMMAF is the standardized
benchmark entrant and the closest security-framed sibling \citep{mateotorrejon2026gammaf}. It
injects adversarial agents into its own debate simulator, uses attack-injection ground truth, and
reports ASR, uASR, ADR, AIR, and AUROC. Its baseline set carries no PyGOD, ADBench, or BOND lineage.
GAMMAF's definition of malfunctioning admits inherent model failure, and uASR includes
non-adversarial wrong answers. That metric still measures security evasion among wrong final
answers; it does not label benign stale, drifted, or out-of-distribution dependency state.

The named graph methods target prompt injection, tool or metadata poisoning, and malicious or
colluding agents. None of those cited evaluations centers its task on the benign dependency-state
failures scored here. \sysname evaluates simplified, task-adapted versions of G-Safeguard and
GUARDIAN on its dependency graph, and reports both on the detection board, following the ADBench and
BOND practice of running a method on the benchmark's own representation.

\subsection{Tracing, Evaluation, and Static Agent Analysis}

GRADE represents dependencies and execution structure for agent traces \citep{zhao2026grade}, while
task benchmarks such as $\tau$-bench score final outcomes \citep{yao2024taubench}. Counterfactual
Trace Auditing pairs a with-skill trace against a without-skill run of the same task and annotates
where the two diverge; across 49 software-engineering tasks it records 522 behavioral effects while
pass rate moves 0.3 points \citep{zhou2026counterfactual}, which measures how much an outcome
number leaves unread. AgentDiagnose adds
trace-level scores for five behavioral competencies and an interactive diagnostic view
\citep{ou2025agentdiagnose}. These systems supply representations, outcomes, or diagnostic views,
but they do not by themselves define the same scored audit across information states. Formal policy
systems evaluate authorization predicates over execution context and can statically analyze the
policy \citep{palumbo2026formal}. On the PRE side, \sysname instead scores whether declared
capabilities exceed what each fixed task record requires. Auditable Agents separates detection
before deployment, enforcement during execution, and recovery afterward by their information
and intervention constraints \citep{nian2026auditable}. The overall distinction is the
conjunction: PRE, LIVE, and POST share a task-method interface. Every scored method then receives
only the information permitted at its state.\looseness=-1

\section{Recorded Comparisons and Their Intervals}
\label{app:stats}

\texttt{tools/statistical\_tests.py} regenerates the comparisons recorded below using
committed judge caches and PRE records together with scores recomputed from upstream corpora.
The learned detection, LIVE, and localization baselines are refitted, including PyGOD in the Gold
comparison. Deterministic controls and seeded Gold injections are recomputed. No API key for a model
service is required, but corpus loading can call Hugging Face; SWE-Gym shard discovery queries the
Hub even when shard files are cached. The script writes test summaries to
\texttt{tools/statistical\_tests\_results.json}; the held-out run-score arrays and pairing identifiers
are omitted. The 138 comparisons this appendix records are grouped by scope, metric, interval
method, and sampling axis, and Table~\ref{tab:all-contrasts} carries those groups as headers over
the rows they cover, which is how a reader navigates a heterogeneous table. A point-estimate
ordering that appears in a table but is not among the recorded comparisons is descriptive and is
not claimed as a result.

We report the observed scores and the differences between them under each board's evaluation
procedure. This appendix retains all 138 recorded comparisons with marginal intervals under the
stated sampling models, and the released \texttt{tools/statistical\_tests\_results.json} carries
their $p$-values. Those quantities provide no family-wise error control, and the text does not use
significance thresholds to establish entrant orderings or equality. We discuss the
structural-feature increments under every tested specification together, as an exploratory
analysis, because the set of specifications was fixed after the effects were inspected. The matched
flexible-size controls restrict the interpretation of the linear-model increments. A comparison of a
LIVE cell against the fixed 0.70 bar is a marginal statement about that one cell, not a simultaneous
guarantee across methods or prefixes.

A table in this appendix carries per-arm intervals when the committed records support one for every
cell it prints. Tables~\ref{tab:protocol}, \ref{tab:pre-source}, \ref{tab:pre-main},
\ref{tab:live-stream}, \ref{tab:live-stream-tau} and~\ref{tab:transfer} meet that condition and
carry them. Five do not, and print point estimates alone: Tables~\ref{tab:loc}, \ref{tab:det},
\ref{tab:gold-full}, \ref{tab:gold-matched} and~\ref{tab:live-stale}. Each caption says what is
missing. In every case the quantity is well defined and the released record does not
carry enough to compute it; what differs is how much would have to change. Most of these cells need
only a rerun of the scorer, because the release omits the score or rank vector the interval would be
built from. A few need the pipeline to retain more than it currently does: Gold averages tie-aware
credit before recording, so the per-run credits are gone rather than merely unpublished, and a
judge's reciprocal rank needs a truth-to-run join that is never formed. We supply no interval on
part of a table, except where a row is a declared non-measurement and its caption says so. A table
carrying intervals on some rows and blanks on others reads as a complete one, and the blanks fall
on the cells a reader is least equipped to notice.

Binary Top-1 and Top-3 outcomes use the exact conditional McNemar test on discordant runs rather than
its chi-square approximation, because at these discordant counts the two straddle $0.05$ in opposite
directions. Reciprocal rank uses Wilcoxon's paired signed-rank test, with the exact binomial sign test
reported alongside as a diagnostic. Ordinary paired ROC-AUC contrasts use paired DeLong, cross-checked
by a stratified paired bootstrap. Gold attribution uses a pair-cluster bootstrap and a within-pair
label-swap randomization, because its construction pairs each injected run with its own clean control
and the runs are therefore not independent. Tests against an analytic Gold floor use the exact
Poisson-binomial randomization distribution rather than a sign test, whose null median is negative for
any pool of three or more candidates while its null mean is zero.

The tau-bench rows carry a dependence of their own, and it changes how their intervals are built.
Its 660 runs are 165 task instances, each attempted by four agent models, so a run-level interval
would treat four attempts at one task as four independent observations and would be narrower than
the design supports. Every tau-bench interval whose sampling axis is the run therefore comes from a
task-clustered percentile bootstrap that resamples task instances within domain, carrying all four
of a selected instance's model rows together, rather than resampling runs. Intervals on other axes
are unaffected, including the seed intervals of Table~\ref{tab:transfer}. SWE-Gym needs no such
repair and keeps the run-level construction: its 376 runs carry 376 distinct task identifiers, so no
instance is attempted twice there.

Two limits apply to every interval reported here, including the per-entrant intervals in
Tables~\ref{tab:live-stream} and~\ref{tab:live-stream-tau}. Each cached judge prediction is a single generation
pass, and the caches record no temperature, seed, or repeat index, so the localization and protocol
intervals reflect run sampling alone and are optimistic about what regeneration would show. DeLong and
the bootstrap both condition on the fitted models, so they omit the variance contributed by refitting
inside each resample; the supervised intervals should be read as test-sample intervals rather than as
full end-to-end uncertainty.

Table~\ref{tab:core-contrasts} collects the four structural contrasts the body discusses, the linear
and matched-spline increments on each corpus, so they can be read together on one estimand with their
intervals. It is separate from the specification comparison in Section~\ref{sec:res-post-det}, which
reports mean fold-score increments, and the two are not interchangeable: the same SWE-Gym spline
effect is $-0.010$ on that estimand and $-0.005$ on this one. Reading a row from one table beside a
row from the other compares two different quantities.

Table~\ref{tab:all-contrasts} prints all 138 recorded comparisons: the two entrants, their
scores, the difference, and the interval on that difference. Two of the four rows above are among
them; the two matched-spline contrasts are computed for that table alone and are not in the registry,
which stays at 138. The $p$-values stay in the released
record rather than in the table, and the text draws no conclusion from them. The body discusses a
subset of these rows and says which; this table is where the rest live, so a reader can check a
point-estimate ordering the body reports without claiming.

\begin{table}[t]
\centering
\small
\setlength{\tabcolsep}{5pt}
\caption{Four exploratory structural contrasts on the released fixed-prediction scoring functional: ROC-AUC of the mean of the five out-of-fold score vectors for arm A, minus the same for arm B.
This differs from the mean fold-score increments in the existing specification comparison; it also differs from averaging five pooled per-seed AUC differences.
Linear compares \texttt{auditable (size+deps)} with \texttt{size (flat)}; matched spline compares size-spline plus linear-dependency terms with size-only spline.
SWE-Gym retains paired run-level DeLong intervals: the recovered identity table verifies 376 runs with 376 distinct task ids.
For tau-bench, 660 runs are 165 tasks attempted by four agent models. Its task-clustered bootstrap resamples 50 airline and 115 retail tasks with replacement within domain, carrying all four model rows, labels, and both arms' scores together. Intervals are the 2.5 and 97.5 percentiles of 10,000 draws; single-class draws are discarded.
Usable draws: linear 10000; matched spline 10000. NumPy PCG64 uses base seed 20260907 with the recorded SHA-256 contrast labels. The last interval digit carries Monte Carlo noise of about 0.001.
Intervals condition on the released fitted predictions and fixed model roster; they omit training, tuning, and fold-redraw uncertainty and do not establish performance on unseen tasks, models, or domains.
Generated by \texttt{tools/emit\_core\_contrasts.py}.}
\label{tab:core-contrasts}
\begin{tabular}{@{}llrr@{}}
\toprule
Corpus & Specification & AUC difference & 95\% interval \\
\midrule
SWE-Gym & Linear & $+0.147$ & $[+0.080, +0.214]$ \\
tau-bench & Linear & $+0.046$ & $[-0.005, +0.098]$ \\
SWE-Gym & Matched spline & $-0.005$ & $[-0.026, +0.015]$ \\
tau-bench & Matched spline & $+0.045$ & $[+0.008, +0.083]$ \\
\bottomrule
\end{tabular}
\end{table}

{\scriptsize
\setlength{\tabcolsep}{3pt}
\begin{longtable}{@{}p{8.87cm}rrr>{\centering\arraybackslash}p{2.15cm}@{}}
\caption{Every recorded comparison, by group. $A$ and $B$ are the two entrants named in the contrast, in that order, and the interval on their difference is built under the construction and sampling axis the group header states. These rows are measurements reported with their uncertainty: no multiplicity correction is applied, no family-wise error control is claimed, and the text draws no conclusion from whether an interval excludes zero. The unadjusted and adjusted $p$-values stay in the released record, \texttt{tools/statistical\_tests\_results.json}, and are not printed here. This table is the printed home of every claim in Appendix~\ref{app:stats}; the body reports the subset it argues from. Generated by \texttt{tools/emit\_stats\_table.py -{}-contrasts}, whose \texttt{-{}-check} mode runs before a submission and fails when the paper falls behind the shipped results.}\label{tab:all-contrasts}\\
\toprule
Contrast & $A$ & $B$ & $A-B$ & 95\% CI \\
\midrule
\endfirsthead
\multicolumn{5}{@{}p{13.6cm}@{}}{\emph{Continued from the previous page.}}\\
\toprule
Contrast & $A$ & $B$ & $A-B$ & 95\% CI \\
\midrule
\endhead
\bottomrule
\endlastfoot
\addlinespace
\multicolumn{5}{@{}p{13.6cm}@{}}{\texttt{localization\_band\_top1} $\cdot$ All 28 pairs in the eight-judge all-at-once Top-1 band. $\cdot$ top1 $\cdot$ paired percentile bootstrap, run-level sampling}\\
gpt-5.5 vs claude-opus-4.8 & 0.452 & 0.421 & $+0.032$ & $[-0.040, 0.103]$ \\
gpt-5.5 vs gpt-5.4 & 0.452 & 0.413 & $+0.040$ & $[-0.032, 0.111]$ \\
gpt-5.5 vs deepseek-r1 & 0.452 & 0.405 & $+0.048$ & $[-0.056, 0.151]$ \\
gpt-5.5 vs gemini & 0.452 & 0.357 & $+0.095$ & $[0.016, 0.175]$ \\
gpt-5.5 vs qwen3-32b & 0.452 & 0.349 & $+0.103$ & $[-0.008, 0.214]$ \\
gpt-5.5 vs gpt-oss-20b & 0.452 & 0.333 & $+0.119$ & $[0.016, 0.222]$ \\
gpt-5.5 vs llama-3.3-70b & 0.452 & 0.333 & $+0.119$ & $[0.008, 0.230]$ \\
claude-opus-4.8 vs gpt-5.4 & 0.421 & 0.413 & $+0.008$ & $[-0.048, 0.071]$ \\
claude-opus-4.8 vs deepseek-r1 & 0.421 & 0.405 & $+0.016$ & $[-0.079, 0.111]$ \\
claude-opus-4.8 vs gemini & 0.421 & 0.357 & $+0.063$ & $[-0.024, 0.151]$ \\
claude-opus-4.8 vs qwen3-32b & 0.421 & 0.349 & $+0.071$ & $[-0.032, 0.175]$ \\
claude-opus-4.8 vs gpt-oss-20b & 0.421 & 0.333 & $+0.087$ & $[-0.008, 0.183]$ \\
claude-opus-4.8 vs llama-3.3-70b & 0.421 & 0.333 & $+0.087$ & $[-0.024, 0.190]$ \\
gpt-5.4 vs deepseek-r1 & 0.413 & 0.405 & $+0.008$ & $[-0.087, 0.103]$ \\
gpt-5.4 vs gemini & 0.413 & 0.357 & $+0.056$ & $[-0.024, 0.127]$ \\
gpt-5.4 vs qwen3-32b & 0.413 & 0.349 & $+0.063$ & $[-0.032, 0.167]$ \\
gpt-5.4 vs gpt-oss-20b & 0.413 & 0.333 & $+0.079$ & $[-0.016, 0.175]$ \\
gpt-5.4 vs llama-3.3-70b & 0.413 & 0.333 & $+0.079$ & $[-0.024, 0.183]$ \\
deepseek-r1 vs gemini & 0.405 & 0.357 & $+0.048$ & $[-0.048, 0.143]$ \\
deepseek-r1 vs qwen3-32b & 0.405 & 0.349 & $+0.056$ & $[-0.040, 0.151]$ \\
deepseek-r1 vs gpt-oss-20b & 0.405 & 0.333 & $+0.071$ & $[-0.016, 0.159]$ \\
deepseek-r1 vs llama-3.3-70b & 0.405 & 0.333 & $+0.071$ & $[-0.032, 0.167]$ \\
gemini vs qwen3-32b & 0.357 & 0.349 & $+0.008$ & $[-0.087, 0.111]$ \\
gemini vs gpt-oss-20b & 0.357 & 0.333 & $+0.024$ & $[-0.071, 0.119]$ \\
gemini vs llama-3.3-70b & 0.357 & 0.333 & $+0.024$ & $[-0.079, 0.127]$ \\
qwen3-32b vs gpt-oss-20b & 0.349 & 0.333 & $+0.016$ & $[-0.079, 0.111]$ \\
qwen3-32b vs llama-3.3-70b & 0.349 & 0.333 & $+0.016$ & $[-0.063, 0.095]$ \\
gpt-oss-20b vs llama-3.3-70b & 0.333 & 0.333 & $+0.000$ & $[-0.095, 0.095]$ \\
\addlinespace
\multicolumn{5}{@{}p{13.6cm}@{}}{\texttt{localization\_gpt55\_no\_llm\_top1} $\cdot$ GPT-5.5 against the five registered no-LLM entrants. $\cdot$ top1 $\cdot$ paired percentile bootstrap, run-level sampling}\\
gpt-5.5 vs random & 0.452 & 0.119 & $+0.333$ & $[0.241, 0.425]$ \\
gpt-5.5 vs auditable (blast) & 0.452 & 0.159 & $+0.294$ & $[0.175, 0.413]$ \\
gpt-5.5 vs position & 0.452 & 0.159 & $+0.294$ & $[0.175, 0.405]$ \\
gpt-5.5 vs pygod (graph AD) & 0.452 & 0.048 & $+0.405$ & $[0.310, 0.500]$ \\
gpt-5.5 vs exec-rank (sup.) & 0.452 & 0.211 & $+0.241$ & $[0.133, 0.351]$ \\
\addlinespace
\multicolumn{5}{@{}p{13.6cm}@{}}{\texttt{localization\_exec\_position} $\cdot$ Exec-rank against position on Top-1, Top-3, and MRR. $\cdot$ metric on each row $\cdot$ paired percentile bootstrap, run-level sampling}\\
exec-rank (sup.) vs position (top1) & 0.211 & 0.159 & $+0.052$ & $[-0.003, 0.110]$ \\
exec-rank (sup.) vs position (top3) & 0.614 & 0.516 & $+0.098$ & $[0.035, 0.167]$ \\
exec-rank (sup.) vs position (mrr) & 0.454 & 0.407 & $+0.048$ & $[0.015, 0.080]$ \\
\addlinespace
\multicolumn{5}{@{}p{13.6cm}@{}}{\texttt{localization\_small\_position\_top1} $\cdot$ Position against Mistral-Small and Nova-Micro. $\cdot$ top1 $\cdot$ paired percentile bootstrap, run-level sampling}\\
position vs mistral-small & 0.159 & 0.135 & $+0.024$ & $[-0.032, 0.079]$ \\
position vs nova-micro & 0.159 & 0.127 & $+0.032$ & $[-0.056, 0.119]$ \\
\addlinespace
\multicolumn{5}{@{}p{13.6cm}@{}}{\texttt{localization\_protocol\_top1} $\cdot$ All-at-once against two alternatives for ten cached judges. $\cdot$ top1 $\cdot$ paired percentile bootstrap, run-level sampling}\\
gpt-5.5: all-at-once vs step-by-step & 0.452 & 0.397 & $+0.056$ & $[-0.040, 0.151]$ \\
gpt-5.5: all-at-once vs binary-search & 0.452 & 0.421 & $+0.032$ & $[-0.024, 0.087]$ \\
claude-opus-4.8: all-at-once vs step-by-step & 0.421 & 0.389 & $+0.032$ & $[-0.056, 0.119]$ \\
claude-opus-4.8: all-at-once vs binary-search & 0.421 & 0.357 & $+0.063$ & $[-0.008, 0.135]$ \\
gpt-5.4: all-at-once vs step-by-step & 0.413 & 0.381 & $+0.032$ & $[-0.063, 0.135]$ \\
gpt-5.4: all-at-once vs binary-search & 0.413 & 0.365 & $+0.048$ & $[-0.032, 0.127]$ \\
deepseek-r1: all-at-once vs step-by-step & 0.405 & 0.317 & $+0.087$ & $[-0.024, 0.190]$ \\
deepseek-r1: all-at-once vs binary-search & 0.405 & 0.405 & $+0.000$ & $[-0.095, 0.095]$ \\
gemini: all-at-once vs step-by-step & 0.357 & 0.341 & $+0.016$ & $[-0.095, 0.127]$ \\
gemini: all-at-once vs binary-search & 0.357 & 0.357 & $+0.000$ & $[-0.103, 0.103]$ \\
qwen3-32b: all-at-once vs step-by-step & 0.349 & 0.254 & $+0.095$ & $[0.000, 0.190]$ \\
qwen3-32b: all-at-once vs binary-search & 0.349 & 0.127 & $+0.222$ & $[0.135, 0.310]$ \\
llama-3.3-70b: all-at-once vs step-by-step & 0.333 & 0.222 & $+0.111$ & $[0.024, 0.198]$ \\
llama-3.3-70b: all-at-once vs binary-search & 0.333 & 0.222 & $+0.111$ & $[0.016, 0.206]$ \\
gemma-3-12b: all-at-once vs step-by-step & 0.206 & 0.230 & $-0.024$ & $[-0.119, 0.071]$ \\
gemma-3-12b: all-at-once vs binary-search & 0.206 & 0.159 & $+0.048$ & $[-0.032, 0.127]$ \\
mistral-small: all-at-once vs step-by-step & 0.135 & 0.190 & $-0.056$ & $[-0.127, 0.016]$ \\
mistral-small: all-at-once vs binary-search & 0.135 & 0.214 & $-0.079$ & $[-0.159, 0.000]$ \\
nova-micro: all-at-once vs step-by-step & 0.127 & 0.167 & $-0.040$ & $[-0.127, 0.048]$ \\
nova-micro: all-at-once vs binary-search & 0.127 & 0.167 & $-0.040$ & $[-0.119, 0.048]$ \\
\addlinespace
\multicolumn{5}{@{}p{13.6cm}@{}}{\texttt{post\_detection\_auc} $\cdot$ Seven prespecified POST detection AUC contrasts. $\cdot$ roc\_auc $\cdot$ interval construction on each row}\\
swegym: auditable (size+deps) vs size (flat) (interval: paired DeLong, run-level sampling) & 0.803 & 0.655 & $+0.147$ & $[0.080, 0.214]$ \\
tau: auditable (size+deps) vs size (flat) (interval: task-clustered stratified percentile bootstrap, task-cluster resampling) & 0.662 & 0.616 & $+0.046$ & $[-0.005, 0.098]$ \\
swegym: g-safeguard (sup GNN) vs full (interval: paired DeLong, run-level sampling) & 0.816 & 0.814 & $+0.002$ & $[-0.020, 0.023]$ \\
swegym: guardian (recon-AE) vs pyod-flatten (ECOD) (interval: paired DeLong, run-level sampling) & 0.767 & 0.765 & $+0.001$ & $[-0.059, 0.061]$ \\
swegym: auditable (size+deps) vs pyod-flatten (ECOD) (interval: paired DeLong, run-level sampling) & 0.803 & 0.765 & $+0.037$ & $[-0.007, 0.082]$ \\
swegym: auditable (size+deps) vs guardian (recon-AE) (interval: paired DeLong, run-level sampling) & 0.803 & 0.767 & $+0.036$ & $[-0.004, 0.076]$ \\
tau: auditable (size+deps) vs full (interval: task-clustered stratified percentile bootstrap, task-cluster resampling) & 0.662 & 0.665 & $-0.002$ & $[-0.033, 0.028]$ \\
\addlinespace
\multicolumn{5}{@{}p{13.6cm}@{}}{\texttt{live\_swegym\_25\_auc} $\cdot$ Four prespecified contrasts at the first SWE-Gym prefix. $\cdot$ roc\_auc $\cdot$ paired DeLong, run-level sampling}\\
SWE-Gym 25\%: auditable vs size & 0.738 & 0.637 & $+0.101$ & $[0.041, 0.161]$ \\
SWE-Gym 25\%: auditable vs ECOD & 0.738 & 0.756 & $-0.018$ & $[-0.072, 0.036]$ \\
SWE-Gym 25\%: full vs auditable (size+deps) & 0.808 & 0.738 & $+0.070$ & $[0.029, 0.112]$ \\
SWE-Gym 25\%: full vs pyod (ECOD) & 0.808 & 0.756 & $+0.053$ & $[0.019, 0.086]$ \\
\addlinespace
\multicolumn{5}{@{}p{13.6cm}@{}}{\texttt{live\_swegym\_auditable\_ecod\_later\_auc} $\cdot$ Auditable against ECOD at the three later SWE-Gym prefixes. $\cdot$ roc\_auc $\cdot$ paired DeLong, run-level sampling}\\
SWE-Gym 50\%: auditable vs ECOD & 0.763 & 0.762 & $+0.001$ & $[-0.056, 0.058]$ \\
SWE-Gym 75\%: auditable vs ECOD & 0.800 & 0.767 & $+0.033$ & $[-0.011, 0.077]$ \\
SWE-Gym 100\%: auditable vs ECOD & 0.803 & 0.765 & $+0.037$ & $[-0.007, 0.082]$ \\
\addlinespace
\multicolumn{5}{@{}p{13.6cm}@{}}{\texttt{live\_swegym\_threshold\_auc} $\cdot$ Exploratory family added after its scores were examined: twenty nonrandom SWE-Gym method-prefix cells against 0.70, two-sided. $\cdot$ roc\_auc $\cdot$ single-curve DeLong, run-level sampling}\\
SWE-Gym 25\%: size (flat) vs 0.70 & 0.637 & 0.700 & $-0.063$ & $[-0.123, -0.004]$ \\
SWE-Gym 25\%: auditable (size+deps) vs 0.70 & 0.738 & 0.700 & $+0.038$ & $[-0.013, 0.089]$ \\
SWE-Gym 25\%: full vs 0.70 & 0.808 & 0.700 & $+0.108$ & $[0.062, 0.154]$ \\
SWE-Gym 25\%: pyod (ECOD) vs 0.70 & 0.756 & 0.700 & $+0.056$ & $[0.005, 0.107]$ \\
SWE-Gym 25\%: dep-span (online) vs 0.70 & 0.364 & 0.700 & $-0.336$ & $[-0.391, -0.281]$ \\
SWE-Gym 50\%: size (flat) vs 0.70 & 0.659 & 0.700 & $-0.041$ & $[-0.101, 0.018]$ \\
SWE-Gym 50\%: auditable (size+deps) vs 0.70 & 0.763 & 0.700 & $+0.063$ & $[0.012, 0.115]$ \\
SWE-Gym 50\%: full vs 0.70 & 0.810 & 0.700 & $+0.110$ & $[0.064, 0.156]$ \\
SWE-Gym 50\%: pyod (ECOD) vs 0.70 & 0.762 & 0.700 & $+0.062$ & $[0.012, 0.112]$ \\
SWE-Gym 50\%: dep-span (online) vs 0.70 & 0.534 & 0.700 & $-0.166$ & $[-0.225, -0.107]$ \\
SWE-Gym 75\%: size (flat) vs 0.70 & 0.671 & 0.700 & $-0.029$ & $[-0.087, 0.029]$ \\
SWE-Gym 75\%: auditable (size+deps) vs 0.70 & 0.800 & 0.700 & $+0.100$ & $[0.053, 0.148]$ \\
SWE-Gym 75\%: full vs 0.70 & 0.821 & 0.700 & $+0.121$ & $[0.077, 0.165]$ \\
SWE-Gym 75\%: pyod (ECOD) vs 0.70 & 0.767 & 0.700 & $+0.067$ & $[0.018, 0.116]$ \\
SWE-Gym 75\%: dep-span (online) vs 0.70 & 0.589 & 0.700 & $-0.111$ & $[-0.169, -0.052]$ \\
SWE-Gym 100\%: size (flat) vs 0.70 & 0.655 & 0.700 & $-0.045$ & $[-0.104, 0.014]$ \\
SWE-Gym 100\%: auditable (size+deps) vs 0.70 & 0.803 & 0.700 & $+0.103$ & $[0.055, 0.150]$ \\
SWE-Gym 100\%: full vs 0.70 & 0.814 & 0.700 & $+0.114$ & $[0.069, 0.159]$ \\
SWE-Gym 100\%: pyod (ECOD) vs 0.70 & 0.765 & 0.700 & $+0.065$ & $[0.016, 0.115]$ \\
SWE-Gym 100\%: dep-span (online) vs 0.70 & 0.648 & 0.700 & $-0.052$ & $[-0.108, 0.005]$ \\
\addlinespace
\multicolumn{5}{@{}p{13.6cm}@{}}{\texttt{live\_tau\_auditable\_ecod\_auc} $\cdot$ Auditable against ECOD at all four tau-bench prefixes. $\cdot$ roc\_auc $\cdot$ task-clustered stratified percentile bootstrap, task-cluster resampling}\\
tau-bench 25\%: auditable vs ECOD & 0.625 & 0.546 & $+0.079$ & $[0.001, 0.156]$ \\
tau-bench 50\%: auditable vs ECOD & 0.617 & 0.553 & $+0.064$ & $[-0.018, 0.146]$ \\
tau-bench 75\%: auditable vs ECOD & 0.639 & 0.562 & $+0.077$ & $[-0.005, 0.162]$ \\
tau-bench 100\%: auditable vs ECOD & 0.662 & 0.555 & $+0.107$ & $[0.029, 0.184]$ \\
\addlinespace
\multicolumn{5}{@{}p{13.6cm}@{}}{\texttt{live\_tau\_threshold\_auc} $\cdot$ Twenty nonrandom tau method-prefix cells against 0.70, two-sided. $\cdot$ roc\_auc $\cdot$ task-clustered stratified percentile bootstrap, task-cluster resampling}\\
tau-bench 25\%: size (flat) vs 0.70 & 0.626 & 0.700 & $-0.074$ & $[-0.134, -0.017]$ \\
tau-bench 25\%: auditable (size+deps) vs 0.70 & 0.625 & 0.700 & $-0.075$ & $[-0.134, -0.016]$ \\
tau-bench 25\%: full vs 0.70 & 0.635 & 0.700 & $-0.065$ & $[-0.126, -0.007]$ \\
tau-bench 25\%: pyod (ECOD) vs 0.70 & 0.546 & 0.700 & $-0.154$ & $[-0.209, -0.097]$ \\
tau-bench 25\%: dep-span (online) vs 0.70 & 0.503 & 0.700 & $-0.197$ & $[-0.200, -0.193]$ \\
tau-bench 50\%: size (flat) vs 0.70 & 0.617 & 0.700 & $-0.083$ & $[-0.145, -0.021]$ \\
tau-bench 50\%: auditable (size+deps) vs 0.70 & 0.617 & 0.700 & $-0.083$ & $[-0.145, -0.023]$ \\
tau-bench 50\%: full vs 0.70 & 0.628 & 0.700 & $-0.072$ & $[-0.132, -0.015]$ \\
tau-bench 50\%: pyod (ECOD) vs 0.70 & 0.553 & 0.700 & $-0.147$ & $[-0.206, -0.087]$ \\
tau-bench 50\%: dep-span (online) vs 0.70 & 0.530 & 0.700 & $-0.170$ & $[-0.193, -0.146]$ \\
tau-bench 75\%: size (flat) vs 0.70 & 0.617 & 0.700 & $-0.083$ & $[-0.147, -0.022]$ \\
tau-bench 75\%: auditable (size+deps) vs 0.70 & 0.639 & 0.700 & $-0.061$ & $[-0.121, -0.004]$ \\
tau-bench 75\%: full vs 0.70 & 0.643 & 0.700 & $-0.057$ & $[-0.117, -0.003]$ \\
tau-bench 75\%: pyod (ECOD) vs 0.70 & 0.562 & 0.700 & $-0.138$ & $[-0.197, -0.080]$ \\
tau-bench 75\%: dep-span (online) vs 0.70 & 0.565 & 0.700 & $-0.135$ & $[-0.188, -0.086]$ \\
tau-bench 100\%: size (flat) vs 0.70 & 0.616 & 0.700 & $-0.084$ & $[-0.146, -0.023]$ \\
tau-bench 100\%: auditable (size+deps) vs 0.70 & 0.662 & 0.700 & $-0.038$ & $[-0.099, 0.020]$ \\
tau-bench 100\%: full vs 0.70 & 0.665 & 0.700 & $-0.035$ & $[-0.093, 0.019]$ \\
tau-bench 100\%: pyod (ECOD) vs 0.70 & 0.555 & 0.700 & $-0.145$ & $[-0.204, -0.086]$ \\
tau-bench 100\%: dep-span (online) vs 0.70 & 0.568 & 0.700 & $-0.132$ & $[-0.199, -0.067]$ \\
\addlinespace
\multicolumn{5}{@{}p{13.6cm}@{}}{\texttt{gold\_localization\_top1} $\cdot$ Three primary Gold localization Top-1 claims. $\cdot$ top1 $\cdot$ paired run bootstrap, run-level sampling}\\
Gold full-pool max-span on stale-state vs analytic floor & 0.703 & 0.029 & $+0.674$ & $[0.581, 0.762]$ \\
Gold full-pool has-dep on dropped grounding vs analytic floor & 0.005 & 0.035 & $-0.030$ & $[-0.034, -0.027]$ \\
Gold matched-pool PyGOD vs max-span overall & 0.404 & 0.394 & $+0.011$ & $[-0.053, 0.074]$ \\
\addlinespace
\multicolumn{5}{@{}p{13.6cm}@{}}{\texttt{gold\_attribution\_auc} $\cdot$ Two Gold attribution features against chance. $\cdot$ roc\_auc $\cdot$ pair-cluster percentile bootstrap, run-level sampling}\\
Gold attribution max-span (higher=stale) vs chance & 0.675 & 0.500 & $+0.175$ & $[0.146, 0.207]$ \\
Gold attribution edge-count (higher=stale) vs chance & 0.566 & 0.500 & $+0.066$ & $[0.058, 0.081]$ \\
\addlinespace
\multicolumn{5}{@{}p{13.6cm}@{}}{\texttt{pre\_rules\_flag\_all\_f1} $\cdot$ Seven pooled rule-based PRE methods against flag-all. $\cdot$ micro\_f1 $\cdot$ paired configuration-cluster percentile bootstrap of a ratio of sums, configuration-cluster sampling}\\
PRE pooled: flag\_risky\_perms vs flag\_all & 0.480 & 0.601 & $-0.121$ & $[-0.176, -0.076]$ \\
PRE pooled: owasp\_excess\_permissions vs flag\_all & 0.505 & 0.601 & $-0.096$ & $[-0.158, -0.047]$ \\
PRE pooled: owasp\_excess\_functionality vs flag\_all & 0.642 & 0.601 & $+0.041$ & $[-0.005, 0.079]$ \\
PRE pooled: owasp\_privilege\_escalation vs flag\_all & 0.020 & 0.601 & $-0.581$ & $[-0.624, -0.529]$ \\
PRE pooled: unrequested\_high\_impact vs flag\_all & 0.240 & 0.601 & $-0.361$ & $[-0.413, -0.308]$ \\
PRE pooled: sensitive\_access vs flag\_all & 0.030 & 0.601 & $-0.571$ & $[-0.612, -0.525]$ \\
PRE pooled: owasp\_asi\_combined vs flag\_all & 0.654 & 0.601 & $+0.053$ & $[0.030, 0.077]$ \\
\addlinespace
\multicolumn{5}{@{}p{13.6cm}@{}}{\texttt{pre\_combined\_judge\_f1} $\cdot$ Pooled combined PRE scanner against the held-out judge. $\cdot$ micro\_f1 $\cdot$ paired configuration-cluster percentile bootstrap of a ratio of sums, configuration-cluster sampling}\\
PRE pooled: held-out judge vs combined scanner & 0.695 & 0.648 & $+0.048$ & $[0.021, 0.074]$ \\
\addlinespace
\multicolumn{5}{@{}p{13.6cm}@{}}{\texttt{pre\_source\_best\_flag\_all\_f1} $\cdot$ Selection-aware best non-oracle PRE candidate against flag-all in six sources. $\cdot$ micro\_f1 $\cdot$ configuration-cluster percentile bootstrap with candidate reselection, configuration-cluster sampling}\\
PRE crewai: best candidate (llm\_judge\_needed(llama-3.3-70b)) vs flag\_all & 0.518 & 0.388 & $+0.130$ & $[0.066, 0.194]$ \\
PRE n8n: best candidate (owasp\_excess\_functionality) vs flag\_all & 0.528 & 0.154 & $+0.374$ & $[0.237, 0.505]$ \\
PRE mcp: best candidate (llm\_judge\_needed(llama-3.3-70b)) vs flag\_all & 0.744 & 0.640 & $+0.104$ & $[0.058, 0.158]$ \\
PRE injecagent: best candidate (llm\_judge\_needed(llama-3.3-70b)) vs flag\_all & 0.990 & 0.750 & $+0.240$ & $[0.232, 0.249]$ \\
PRE sweagent: best candidate (owasp\_excess\_functionality) vs flag\_all & 0.574 & 0.574 & $-0.000$ & $[-0.007, 0.007]$ \\
PRE synthetic: best candidate (llm\_judge\_needed(llama-3.3-70b)) vs flag\_all & 0.972 & 0.763 & $+0.209$ & $[0.163, 0.252]$ \\
\addlinespace
\multicolumn{5}{@{}p{13.6cm}@{}}{\texttt{pre\_narrow\_precision} $\cdot$ Three narrow PRE rules' precision against the capability base rate. $\cdot$ precision $\cdot$ paired configuration-cluster percentile bootstrap of a ratio of sums, configuration-cluster sampling}\\
PRE pooled: owasp\_privilege\_escalation precision vs capability base rate & 0.811 & 0.430 & $+0.381$ & $[0.249, 0.513]$ \\
PRE pooled: unrequested\_high\_impact precision vs capability base rate & 0.633 & 0.430 & $+0.204$ & $[0.150, 0.264]$ \\
PRE pooled: sensitive\_access precision vs capability base rate & 0.763 & 0.430 & $+0.333$ & $[0.047, 0.515]$ \\
\end{longtable}
}


\section{Entrant Adaptations}
\label{app:entrants}

Two graph baselines carry a published agent-specific mechanism onto this representation. Neither
reproduces its source system, and the differences are stated here so a reader does not read either
row as a reimplementation.\looseness=-1

\texttt{guardian (recon-AE)} keeps GUARDIAN's unsupervised reconstruction objective and simplifies
its adjacency-reconstruction and information-bottleneck terms to attribute reconstruction
\citep{zhou2025guardian}. \texttt{g-safeguard (sup GNN)} carries G-Safeguard's supervised
graph-message-passing idea over to structural features and run-level failure scoring, and omits its
adversarial-agent localization and its topological remediation step \citep{wang2025gsafeguard}. Both
therefore score the mechanism on this benchmark's evidence rather than the published system, and a
weak row is evidence about the adapted mechanism rather than about the original tool.

\section{Full Board Values}
\label{app:board-values}

The body reports the principal comparisons as figures or compact prose, so exact cells and supporting
diagnostics live here. The relocated board tables carry the values the body previously printed,
unchanged; Appendix~\ref{app:stats} prints the recorded comparisons themselves.

\begin{table}[htbp]
\centering
\caption{Fault localization on Who\&When (126 failed runs, human mistake-step labels). Top-1 / Top-3 /
MRR. The LLM-judge panel uses the all-at-once protocol \citep{zhang2025whoandwhen}, the default of
three protocols compared in Table~\ref{tab:protocol}; predictions are
cached and committed, so scoring is deterministic and API-free. The GPT-5.5 reference is documented
as generated through the Codex subscription CLI (\texttt{codex exec}); its historical cache does
not record the channel. The panel runner supports the NAIRR gateway for the other frontier judges
and AWS Bedrock for open-weights and small proprietary judges. The leading judges' cells sit close
together at $n = 126$; Appendix~\ref{app:stats} carries each difference recorded on this board with
its interval, and it records no comparison between the position prior and the random floor. Rows are
ordered by Top-1 point estimate, so the generic graph detector sits below the random floor as a
displayed cell rather than as a recorded comparison. $\dagger$ marks the one
stochastic entrant: across twenty detector-initialization seeds it reaches 0.057 $\pm$ 0.019 Top-1,
so its displayed single-seed cell understates it and both readings stay under the floor.
Point estimates only. Some of these quantities need a score or rank regeneration, and a judge's
reciprocal rank additionally needs a truth-to-run join the released records omit, so no cell here is
given an interval.}
\label{tab:loc}
\begin{tabular}{lccc}
\toprule
Method & Top-1 & Top-3 & MRR \\
\midrule
\multicolumn{4}{l}{\emph{LLM-judge panel (all-at-once)}} \\
GPT-5.5 & 0.452 & 0.667 & 0.618 \\
Claude-Opus-4.8 & 0.421 & 0.698 & 0.605 \\
GPT-5.4 & 0.413 & 0.714 & 0.601 \\
DeepSeek-R1 & 0.405 & 0.754 & 0.606 \\
Gemini & 0.357 & 0.722 & 0.572 \\
Qwen3-32B & 0.349 & 0.659 & 0.541 \\
GPT-oss-20B & 0.333 & 0.595 & 0.521 \\
Llama-3.3-70B & 0.333 & 0.579 & 0.515 \\
Gemma-3-12B & 0.206 & 0.524 & 0.427 \\
Mistral-Small & 0.135 & 0.421 & 0.363 \\
Nova-Micro & 0.127 & 0.397 & 0.342 \\
\midrule
\multicolumn{4}{l}{\emph{Structural / baseline (no LLM)}} \\
\texttt{exec-rank (sup.)} & 0.211 & 0.614 & 0.454 \\
\texttt{auditable} (blast) & 0.159 & 0.516 & 0.407 \\
position & 0.159 & 0.516 & 0.407 \\
random (empirical) & 0.119 & 0.346 & 0.324 \\
PyGOD (graph AD)$^{\dagger}$ & 0.048 & 0.302 & 0.258 \\
\bottomrule
\end{tabular}
\end{table}

\begin{table}[t]
\centering
\small
\setlength{\tabcolsep}{5pt}
\caption{LLM-judge elicitation protocols on Who\&When (126 runs, Top-1), adapted from Who\&When \citep{zhang2025whoandwhen}.
Each cell gives its hit rate above a marginal 95\% interval over the 126 runs, conditional on the recorded predictions.
Procedure BIN uses 10,000 binomial draws with success probability $k/126$, then the 2.5 and 97.5 percentiles of the sampled hit rates (linear quantiles).
The generator uses NumPy PCG64, base seed 20260907, and the released SHA-256 label rule for each protocol, model, and metric.
These marginal intervals support no comparison between two cells: the runs are shared, and this construction cannot recover their pairing.
Appendix~\ref{app:stats} carries each model's two protocol differences with their paired intervals.
All protocols normalize whitespace. All-at-once shows every step in one call with a 1500-character content cap per step; its point estimates repeat the Top-1 column of Table~\ref{tab:loc}, which prints no interval of its own.
Step-by-step reveals growing prefixes (1200 characters per visible step); binary-search retains every step index while halving the suspect interval (900 characters per step).
Structural baselines appear in Table~\ref{tab:loc}. GPT-oss-20B is omitted because Bedrock capacity was unavailable during regeneration of the two alternatives.
Generated offline from the committed marginal scalars by \texttt{tools/emit\_protocol\_table.py}; \texttt{-{}-check} detects paper drift.}
\label{tab:protocol}
\begin{tabular}{@{}lccc@{}}
\toprule
Model & All-at-once & Step-by-step & Binary-search \\
\midrule
GPT-5.5 & \shortstack{0.452\\{\footnotesize $[0.365, 0.540]$}} & \shortstack{0.397\\{\footnotesize $[0.310, 0.484]$}} & \shortstack{0.421\\{\footnotesize $[0.333, 0.508]$}} \\
\addlinespace[3pt]
Claude-Opus-4.8 & \shortstack{0.421\\{\footnotesize $[0.333, 0.508]$}} & \shortstack{0.389\\{\footnotesize $[0.302, 0.476]$}} & \shortstack{0.357\\{\footnotesize $[0.278, 0.444]$}} \\
\addlinespace[3pt]
GPT-5.4 & \shortstack{0.413\\{\footnotesize $[0.325, 0.500]$}} & \shortstack{0.381\\{\footnotesize $[0.294, 0.468]$}} & \shortstack{0.365\\{\footnotesize $[0.286, 0.452]$}} \\
\addlinespace[3pt]
DeepSeek-R1 & \shortstack{0.405\\{\footnotesize $[0.317, 0.492]$}} & \shortstack{0.317\\{\footnotesize $[0.238, 0.397]$}} & \shortstack{0.405\\{\footnotesize $[0.317, 0.492]$}} \\
\addlinespace[3pt]
Gemini & \shortstack{0.357\\{\footnotesize $[0.278, 0.444]$}} & \shortstack{0.341\\{\footnotesize $[0.262, 0.421]$}} & \shortstack{0.357\\{\footnotesize $[0.278, 0.444]$}} \\
\addlinespace[3pt]
Qwen3-32B & \shortstack{0.349\\{\footnotesize $[0.270, 0.437]$}} & \shortstack{0.254\\{\footnotesize $[0.183, 0.333]$}} & \shortstack{0.127\\{\footnotesize $[0.071, 0.190]$}} \\
\addlinespace[3pt]
Llama-3.3-70B & \shortstack{0.333\\{\footnotesize $[0.246, 0.413]$}} & \shortstack{0.222\\{\footnotesize $[0.151, 0.294]$}} & \shortstack{0.222\\{\footnotesize $[0.151, 0.302]$}} \\
\addlinespace[3pt]
Gemma-3-12B & \shortstack{0.206\\{\footnotesize $[0.135, 0.278]$}} & \shortstack{0.230\\{\footnotesize $[0.159, 0.310]$}} & \shortstack{0.159\\{\footnotesize $[0.095, 0.222]$}} \\
\addlinespace[3pt]
Mistral-Small & \shortstack{0.135\\{\footnotesize $[0.079, 0.198]$}} & \shortstack{0.190\\{\footnotesize $[0.127, 0.262]$}} & \shortstack{0.214\\{\footnotesize $[0.143, 0.286]$}} \\
\addlinespace[3pt]
Nova-Micro & \shortstack{0.127\\{\footnotesize $[0.071, 0.190]$}} & \shortstack{0.167\\{\footnotesize $[0.103, 0.230]$}} & \shortstack{0.167\\{\footnotesize $[0.103, 0.230]$}} \\
\bottomrule
\end{tabular}
\end{table}
\begin{table*}[t]
\centering
\scriptsize
\setlength{\tabcolsep}{2pt}
\renewcommand{\arraystretch}{1.15}
\begin{tabular}{@{}>{\raggedright\arraybackslash}p{2.35cm}*{7}{>{\centering\arraybackslash}p{\dimexpr(\linewidth-2.35cm-28pt)/7\relax}}@{}}
\toprule
Method & crewai & n8n & mcp & injecagent & sweagent & synthetic & overall \\
\midrule
Flag all & \shortstack{0.388\\{\fontsize{6}{7}\selectfont $[0.334,0.439]$}} & \shortstack{0.154\\{\fontsize{6}{7}\selectfont $[0.111,0.199]$}} & \shortstack{0.654\\{\fontsize{6}{7}\selectfont $[0.544,0.731]$}} & \shortstack{0.750\\{\fontsize{6}{7}\selectfont $[0.743,0.756]$}} & \shortstack{0.574\\{\fontsize{6}{7}\selectfont $[0.549,0.598]$}} & \shortstack{0.763\\{\fontsize{6}{7}\selectfont $[0.734,0.792]$}} & \shortstack{0.601\\{\fontsize{6}{7}\selectfont $[0.551,0.644]$}} \\
\addlinespace[2pt]
Flag none & \shortstack{0.000\\{\fontsize{6}{7}\selectfont $[0.000,0.000]$}} & \shortstack{0.000\\{\fontsize{6}{7}\selectfont $[0.000,0.000]$}} & \shortstack{0.000\\{\fontsize{6}{7}\selectfont $[0.000,0.000]$}} & \shortstack{0.000\\{\fontsize{6}{7}\selectfont $[0.000,0.000]$}} & \shortstack{0.000\\{\fontsize{6}{7}\selectfont $[0.000,0.000]$}} & \shortstack{0.000\\{\fontsize{6}{7}\selectfont $[0.000,0.000]$}} & \shortstack{0.000\\{\fontsize{6}{7}\selectfont $[0.000,0.000]$}} \\
\addlinespace[2pt]
Risky permissions & \shortstack{0.326\\{\fontsize{6}{7}\selectfont $[0.243,0.407]$}} & \shortstack{0.095\\{\fontsize{6}{7}\selectfont $[0.053,0.138]$}} & \shortstack{0.575\\{\fontsize{6}{7}\selectfont $[0.445,0.657]$}} & \shortstack{0.827\\{\fontsize{6}{7}\selectfont $[0.809,0.847]$}} & \shortstack{0.025\\{\fontsize{6}{7}\selectfont $[0.012,0.039]$}} & \shortstack{0.803\\{\fontsize{6}{7}\selectfont $[0.759,0.846]$}} & \shortstack{0.480\\{\fontsize{6}{7}\selectfont $[0.401,0.548]$}} \\
\addlinespace[2pt]
Excess permissions & \shortstack{0.327\\{\fontsize{6}{7}\selectfont $[0.235,0.416]$}} & \shortstack{0.052\\{\fontsize{6}{7}\selectfont $[0.000,0.119]$}} & \shortstack{0.566\\{\fontsize{6}{7}\selectfont $[0.421,0.654]$}} & \shortstack{0.801\\{\fontsize{6}{7}\selectfont $[0.777,0.825]$}} & \shortstack{0.007\\{\fontsize{6}{7}\selectfont $[0.000,0.015]$}} & \shortstack{0.825\\{\fontsize{6}{7}\selectfont $[0.762,0.880]$}} & \shortstack{0.505\\{\fontsize{6}{7}\selectfont $[0.417,0.575]$}} \\
\addlinespace[2pt]
Excess functionality & \shortstack{0.451\\{\fontsize{6}{7}\selectfont $[0.393,0.505]$}} & \shortstack{0.514\\{\fontsize{6}{7}\selectfont $[0.355,0.642]$}} & \shortstack{0.632\\{\fontsize{6}{7}\selectfont $[0.468,0.738]$}} & \shortstack{0.957\\{\fontsize{6}{7}\selectfont $[0.945,0.969]$}} & \shortstack{0.574\\{\fontsize{6}{7}\selectfont $[0.548,0.599]$}} & \shortstack{0.539\\{\fontsize{6}{7}\selectfont $[0.417,0.643]$}} & \shortstack{0.642\\{\fontsize{6}{7}\selectfont $[0.579,0.694]$}} \\
\addlinespace[2pt]
Privilege escalation & \shortstack{0.000\\{\fontsize{6}{7}\selectfont $[0.000,0.000]$}} & \shortstack{0.000\\{\fontsize{6}{7}\selectfont $[0.000,0.000]$}} & \shortstack{0.018\\{\fontsize{6}{7}\selectfont $[0.006,0.028]$}} & \shortstack{0.065\\{\fontsize{6}{7}\selectfont $[0.038,0.096]$}} & \shortstack{0.000\\{\fontsize{6}{7}\selectfont $[0.000,0.000]$}} & \shortstack{0.000\\{\fontsize{6}{7}\selectfont $[0.000,0.000]$}} & \shortstack{0.020\\{\fontsize{6}{7}\selectfont $[0.013,0.028]$}} \\
\addlinespace[2pt]
Unrequested impact & \shortstack{0.066\\{\fontsize{6}{7}\selectfont $[0.014,0.123]$}} & \shortstack{0.041\\{\fontsize{6}{7}\selectfont $[0.000,0.133]$}} & \shortstack{0.211\\{\fontsize{6}{7}\selectfont $[0.123,0.266]$}} & \shortstack{0.605\\{\fontsize{6}{7}\selectfont $[0.575,0.633]$}} & \shortstack{0.000\\{\fontsize{6}{7}\selectfont $[0.000,0.000]$}} & \shortstack{0.248\\{\fontsize{6}{7}\selectfont $[0.157,0.333]$}} & \shortstack{0.240\\{\fontsize{6}{7}\selectfont $[0.208,0.269]$}} \\
\addlinespace[2pt]
Sensitive access & \shortstack{0.014\\{\fontsize{6}{7}\selectfont $[0.000,0.045]$}} & \shortstack{0.000\\{\fontsize{6}{7}\selectfont $[0.000,0.000]$}} & \shortstack{0.058\\{\fontsize{6}{7}\selectfont $[0.023,0.086]$}} & \shortstack{0.000\\{\fontsize{6}{7}\selectfont $[0.000,0.000]$}} & \shortstack{0.000\\{\fontsize{6}{7}\selectfont $[0.000,0.000]$}} & \shortstack{0.000\\{\fontsize{6}{7}\selectfont $[0.000,0.000]$}} & \shortstack{0.030\\{\fontsize{6}{7}\selectfont $[0.011,0.049]$}} \\
\addlinespace[2pt]
ASI combined & \shortstack{0.448\\{\fontsize{6}{7}\selectfont $[0.391,0.502]$}} & \shortstack{0.411\\{\fontsize{6}{7}\selectfont $[0.271,0.540]$}} & \shortstack{0.644\\{\fontsize{6}{7}\selectfont $[0.534,0.720]$}} & \shortstack{0.961\\{\fontsize{6}{7}\selectfont $[0.950,0.972]$}} & \shortstack{0.570\\{\fontsize{6}{7}\selectfont $[0.545,0.596]$}} & \shortstack{0.842\\{\fontsize{6}{7}\selectfont $[0.798,0.884]$}} & \shortstack{0.654\\{\fontsize{6}{7}\selectfont $[0.605,0.691]$}} \\
\addlinespace[2pt]
Oracle & \shortstack{1.000$^{\mathrm{I}}$\\{\tiny identity}} & \shortstack{1.000$^{\mathrm{I}}$\\{\tiny identity}} & \shortstack{1.000$^{\mathrm{I}}$\\{\tiny identity}} & \shortstack{1.000$^{\mathrm{I}}$\\{\tiny identity}} & \shortstack{1.000$^{\mathrm{I}}$\\{\tiny identity}} & \shortstack{1.000$^{\mathrm{I}}$\\{\tiny identity}} & \shortstack{1.000$^{\mathrm{I}}$\\{\tiny identity}} \\
\addlinespace[2pt]
Held-out LLM judge & \shortstack{0.518\\{\fontsize{6}{7}\selectfont $[0.442,0.586]$}} & \shortstack{0.362\\{\fontsize{6}{7}\selectfont $[0.241,0.484]$}} & \shortstack{0.744\\{\fontsize{6}{7}\selectfont $[0.599,0.845]$}} & \shortstack{0.990\\{\fontsize{6}{7}\selectfont $[0.984,0.996]$}} & \shortstack{0.467\\{\fontsize{6}{7}\selectfont $[0.428,0.506]$}} & \shortstack{0.972\\{\fontsize{6}{7}\selectfont $[0.945,0.995]$}} & \shortstack{0.695\\{\fontsize{6}{7}\selectfont $[0.633,0.745]$}} \\
\addlinespace[2pt]
\bottomrule
\end{tabular}
\caption{PRE capability-level micro-F1 by configuration source. Each bracket is a marginal 95\% percentile interval on that one cell, source-stratified over configuration clusters, conditional on the released labels and fixed predictions. Each draw carries a configuration's capabilities together and pools TP, FP, and FN before computing the metric. A source column resamples only that source; overall preserves the six source counts. These marginal intervals support no comparison between two cells because they cannot recover their pairing; overlap establishes neither equality nor simultaneous coverage. New intervals use 10,000 draws, linear quantiles, PCG64, and base seed 20260907. Zero denominators give zero. Equal endpoints, including flag-none's $[0,0]$, are empirical degenerate intervals. $^{\mathrm{I}}$ marks a construction identity with no performance interval: declared minus minimal equals the released excess labels. The held-out judge is Llama-3.3-70B, scored on 1182 parsed configurations: 298 crewai, 215 n8n, 143 mcp, 340 injecagent, 130 sweagent, and 56 synthetic. Other rows use all 1187 (219 n8n and 144 mcp). The overall column mixes four label processes: joint LLM labels for crewai, n8n, and mcp; roster labels for injecagent; declared-minus-used labels for sweagent; and injected labels for synthetic. The judge abstains on five configurations, including an MCP server with 622 capabilities and 337 excess labels. Generated by \texttt{tools/emit\_pre\_tables.py}; full-precision quantities and input hashes accompany the source as comments.}
\label{tab:pre-source}
\end{table*}

\begin{table}[t]
\centering
\scriptsize
\setlength{\tabcolsep}{4pt}
\caption{LIVE streaming early warning on SWE-Gym. Each nonrandom cell prints
the board ROC-AUC (B), the interval's point estimate (P), and an unadjusted
95\% single-curve DeLong interval for P (run-level sampling axis). For supervised
size, auditable, and full, B is mean fold AUC and P is the AUC of seed-averaged
out-of-fold scores; their absolute gap is at most 0.008 here.
For ECOD and dep-span, B and P agree at the displayed precision. See
Appendix~\ref{app:board-values} for the same convention gap on the POST board.
Each interval is one entrant's own, over the 376 labeled runs of this
corpus. It is not a comparison: the entrants share those runs, so reading two intervals
against each other discards the pairing a paired test keeps, and overlap is therefore not
evidence that two entrants perform alike. Comparing P's interval with the fixed 0.70
bar gives a marginal, unadjusted reading for that cell. We claim no simultaneous
coverage across methods or prefixes.
The 376 runs of this corpus carry 376 distinct task
identifiers, so resampling runs and resampling tasks are the same draw here.
PyOD (ECOD) is batch-unsupervised; dep-span is
strict per-run online.
Random is a label-independent reference with board point estimates only; no interval
is recorded. Time to detection (t2d) is the earliest prefix whose board ROC-AUC reaches
0.70 (``none'' if no prefix does). Prefix fractions are relative to eventual trace length.}
\label{tab:live-stream}
\begin{tabular}{lccccc}
\toprule
Method & 25\% & 50\% & 75\% & 100\% & t2d \\
\midrule
random & 0.483 & 0.483 & 0.483 & 0.483 & none \\
size (flat)
  & \shortstack{B: 0.629\\P: 0.637\\$[0.577, 0.696]$}
  & \shortstack{B: 0.663\\P: 0.659\\$[0.599, 0.718]$}
  & \shortstack{B: 0.673\\P: 0.671\\$[0.613, 0.729]$}
  & \shortstack{B: 0.663\\P: 0.655\\$[0.596, 0.714]$}
  & none \\
\texttt{auditable} (size+deps)
  & \shortstack{B: 0.742\\P: 0.738\\$[0.687, 0.789]$}
  & \shortstack{B: 0.766\\P: 0.763\\$[0.712, 0.815]$}
  & \shortstack{B: 0.804\\P: 0.800\\$[0.753, 0.848]$}
  & \shortstack{B: 0.804\\P: 0.803\\$[0.755, 0.850]$}
  & 25\% \\
full
  & \shortstack{B: 0.813\\P: 0.808\\$[0.762, 0.854]$}
  & \shortstack{B: 0.816\\P: 0.810\\$[0.764, 0.856]$}
  & \shortstack{B: 0.826\\P: 0.821\\$[0.777, 0.865]$}
  & \shortstack{B: 0.819\\P: 0.814\\$[0.769, 0.859]$}
  & 25\% \\
\midrule
PyOD (ECOD, unsup.)
  & \shortstack{B: 0.756\\P: 0.756\\$[0.705, 0.807]$}
  & \shortstack{B: 0.762\\P: 0.762\\$[0.712, 0.812]$}
  & \shortstack{B: 0.767\\P: 0.767\\$[0.718, 0.816]$}
  & \shortstack{B: 0.765\\P: 0.765\\$[0.716, 0.815]$}
  & 25\% \\
dep-span (online)
  & \shortstack{B: 0.364\\P: 0.364\\$[0.309, 0.419]$}
  & \shortstack{B: 0.534\\P: 0.534\\$[0.475, 0.593]$}
  & \shortstack{B: 0.589\\P: 0.589\\$[0.531, 0.648]$}
  & \shortstack{B: 0.648\\P: 0.648\\$[0.592, 0.705]$}
  & none \\
\bottomrule
\end{tabular}
\end{table}

\begin{table}[t]
\centering
\scriptsize
\setlength{\tabcolsep}{4pt}
\caption{LIVE streaming early warning on tau-bench. Each nonrandom cell prints
the board ROC-AUC (B), the interval's point estimate (P), and an unadjusted
95\% task-clustered stratified percentile bootstrap interval for P
(task-cluster resampling axis). For supervised
size, auditable, and full, B is mean fold AUC and P is the AUC of seed-averaged
out-of-fold scores; their absolute gap is at most 0.008 here.
For ECOD and dep-span, B and P agree at the displayed precision. See
Appendix~\ref{app:board-values} for the same convention gap on the POST board.
Each interval is one entrant's own, over the 660 labeled runs of this
corpus. It is not a comparison: the entrants share those runs, so reading two intervals
against each other discards the pairing a paired test keeps, and overlap is therefore not
evidence that two entrants perform alike. Comparing P's interval with the fixed 0.70
bar gives a marginal, unadjusted reading for that cell. We claim no simultaneous
coverage across methods or prefixes.
The 660 runs of this corpus are 165 task instances, each
attempted by four agent models. Counting those attempts as independent observations
gives an interval narrower than the data support, so each draw resamples task instances
with replacement within domain (50 airline, 115 retail), carrying all four attempts at a
drawn task together. The interval is the 2.5 and 97.5 percentiles of 10000 such draws
(seed 20260907), so it is asymmetric about P and its last printed digit carries
Monte Carlo noise of about 0.002.
PyOD (ECOD) is batch-unsupervised; dep-span is
strict per-run online.
Random is a label-independent reference with board point estimates only; no interval
is recorded. Time to detection (t2d) is the earliest prefix whose board ROC-AUC reaches
0.70 (``none'' if no prefix does). Prefix fractions are relative to eventual trace length.}
\label{tab:live-stream-tau}
\begin{tabular}{lccccc}
\toprule
Method & 25\% & 50\% & 75\% & 100\% & t2d \\
\midrule
random & 0.498 & 0.498 & 0.498 & 0.498 & none \\
size (flat)
  & \shortstack{B: 0.632\\P: 0.626\\$[0.566, 0.683]$}
  & \shortstack{B: 0.618\\P: 0.617\\$[0.555, 0.679]$}
  & \shortstack{B: 0.620\\P: 0.617\\$[0.553, 0.678]$}
  & \shortstack{B: 0.619\\P: 0.616\\$[0.554, 0.677]$}
  & none \\
\texttt{auditable} (size+deps)
  & \shortstack{B: 0.632\\P: 0.625\\$[0.566, 0.684]$}
  & \shortstack{B: 0.617\\P: 0.617\\$[0.555, 0.677]$}
  & \shortstack{B: 0.640\\P: 0.639\\$[0.579, 0.696]$}
  & \shortstack{B: 0.665\\P: 0.662\\$[0.601, 0.720]$}
  & none \\
full
  & \shortstack{B: 0.642\\P: 0.635\\$[0.574, 0.693]$}
  & \shortstack{B: 0.628\\P: 0.628\\$[0.568, 0.685]$}
  & \shortstack{B: 0.644\\P: 0.643\\$[0.583, 0.697]$}
  & \shortstack{B: 0.665\\P: 0.665\\$[0.607, 0.719]$}
  & none \\
\midrule
PyOD (ECOD, unsup.)
  & \shortstack{B: 0.546\\P: 0.546\\$[0.491, 0.603]$}
  & \shortstack{B: 0.553\\P: 0.553\\$[0.494, 0.613]$}
  & \shortstack{B: 0.562\\P: 0.562\\$[0.503, 0.620]$}
  & \shortstack{B: 0.555\\P: 0.555\\$[0.496, 0.614]$}
  & none \\
dep-span (online)
  & \shortstack{B: 0.503\\P: 0.503\\$[0.500, 0.507]$}
  & \shortstack{B: 0.530\\P: 0.530\\$[0.507, 0.554]$}
  & \shortstack{B: 0.565\\P: 0.565\\$[0.512, 0.614]$}
  & \shortstack{B: 0.568\\P: 0.568\\$[0.501, 0.633]$}
  & none \\
\bottomrule
\end{tabular}
\end{table}

\begin{table}[t]
\centering
\caption{\sysname-Gold, full candidate pool. Top-1 by fault kind. Equal scores are resolved in
expectation over the tied ranks, so a baseline that assigns one constant score lands on the random
floor rather than on whatever the pool order happens to give it. Every row but PyGOD is deterministic
given the injection seed. PyGOD also fits an autoencoder, so its cell fixes injection seed 0 and
initialization seed 0; the text reports the two seed axes separately.
Point estimates only. Most of these estimates need a regeneration run, and Gold's tie-aware credits
make per-run uncertainty unreconstructable from the averages displayed here. The analytic floors are
references and take no interval.}
\label{tab:gold-full}
\begin{tabular}{lccc}
\toprule
Method & overall & stale-state & dropped-grounding \\
\midrule
random (seed-averaged) & 0.032 & 0.029 & 0.035 \\
position (leak check) & 0.000 & 0.000 & 0.000 \\
degree (leak check) & 0.045 & 0.073 & 0.023 \\
has-dep (control) & 0.078 & 0.173 & 0.005 \\
max-span (control) & 0.309 & 0.703 & 0.005 \\
\texttt{auditable} (dep-anomaly) & 0.309 & 0.703 & 0.005 \\
PyGOD (graph AD) & 0.165 & 0.256 & 0.094 \\
\bottomrule
\end{tabular}
\end{table}

\begin{table}[t]
\centering
\caption{\sysname-Gold, eligibility-matched pool. Top-1 by fault kind.
Point estimates only. The method estimates need the per-run matched-pool credits, which the released
record does not retain. The random entries are analytic references and take no interval.}
\label{tab:gold-matched}
\begin{tabular}{lccc}
\toprule
Method & overall & stale-state & dropped-grounding \\
\midrule
random (matched floor) & 0.308 & 0.350 & 0.277 \\
position & 0.330 & 0.341 & 0.321 \\
degree & 0.225 & 0.394 & 0.095 \\
has-dep & 0.195 & 0.350 & 0.075 \\
max-span & 0.394 & 0.805 & 0.075 \\
\texttt{auditable} (dep-anomaly) & 0.391 & 0.799 & 0.075 \\
PyGOD (graph AD) & 0.404 & 0.622 & 0.236 \\
\bottomrule
\end{tabular}
\end{table}

\clearpage
Table~\ref{tab:transfer} carries the stability diagnostics of
Section~\ref{sec:res-transfer}: each unsupervised detector's spread across initialization
seeds, the run-size control on the one entrant that approaches the supervised reference,
and the Welch comparison between that entrant and the reference on the seed axis. Every value is derived from the two
committed seed records rather than from a console session, so a reader can recompute all of
it without a corpus, a GRADE checkout, or torch.

\begin{table}[t]
\centering
\scriptsize
\setlength{\tabcolsep}{4pt}
\caption{Transfer diagnostics for Section~\ref{sec:res-transfer}, derived from the two committed seed records. The upper block is each unsupervised detector's ROC-AUC over 20 initialization seeds, with the interval on its mean. The supervised reference varies on a different axis and over five seeds, so its row is separated. The lower block is the run-size control on GAAN, and the Welch comparison of GAAN with the supervised reference. SD is the seed record's own \texttt{std} field, which is the population form over the seeds drawn, and is what the body prints after a plus-or-minus; the interval is the ordinary t interval and therefore uses the sample form, so its half-width is slightly wider than SD alone implies. Generated by \texttt{tools/emit\_transfer\_table.py}, which reads only \texttt{tools/pygod\_seed\_stability\_results\{,\_20seeds\}.json}; its \texttt{-{}-check} mode runs before a submission and fails when this table falls behind them.}
\label{tab:transfer}
\begin{tabular}{@{}llrrr@{}}
\toprule
Corpus & Entrant & Mean & SD & 95\% CI on the mean \\
\midrule
SWE-Gym & DOMINANT & 0.631 & 0.164 & $[0.553, 0.710]$ \\
 & AnomalyDAE & 0.488 & 0.149 & $[0.417, 0.560]$ \\
 & CONAD & 0.596 & 0.099 & $[0.549, 0.644]$ \\
 & GAAN & 0.774 & 0.091 & $[0.730, 0.817]$ \\
\addlinespace
tau-bench & DOMINANT & 0.513 & 0.042 & $[0.493, 0.533]$ \\
 & AnomalyDAE & 0.523 & 0.031 & $[0.508, 0.538]$ \\
 & CONAD & 0.513 & 0.039 & $[0.494, 0.532]$ \\
 & GAAN & 0.514 & 0.023 & $[0.503, 0.525]$ \\
\addlinespace
SWE-Gym & g-safeguard (sup GNN), 5 seeds & 0.824 & 0.007 & $[0.814, 0.834]$ \\
\midrule
\multicolumn{5}{@{}l}{\emph{Run-size control on GAAN, SWE-Gym}} \\
Quantity & \multicolumn{4}{l}{Value} \\
Spearman of run score with node count & \multicolumn{4}{l}{$-0.736$ over 20 seeds (the five-seed record reads $-0.745$)} \\
Runs in strata holding both outcomes & \multicolumn{4}{l}{224 of 376, in 42 node-count strata, giving 262 positive-negative pairs} \\
Within-size ROC-AUC on that pair set & \multicolumn{4}{l}{0.652 $\pm$ 0.169 over 20 seeds, 95\% CI $[0.571, 0.734]$} \\
\addlinespace
\multicolumn{5}{@{}l}{\emph{GAAN against the supervised graph network, Welch on the seed samples}} \\
Difference in means & \multicolumn{4}{l}{$-0.050$, 95\% CI $[-0.095, -0.006]$, $\nu = 20.1$, $p = 0.027$} \\
\bottomrule
\end{tabular}
\end{table}

\clearpage
\subsection{Additional PRE Board Values}
\begin{table*}[t]
\centering
\scriptsize
\setlength{\tabcolsep}{2pt}
\renewcommand{\arraystretch}{1.15}
\begin{tabular}{@{}*{3}{>{\raggedright\arraybackslash}p{\dimexpr(\linewidth-8pt)/3\relax}}@{}}
\toprule
Public category & Scanner rule & Standard reference \\
\midrule
Excessive permissions / least privilege & \texttt{owasp\_\allowbreak{}excess\_\allowbreak{}permissions} & OWASP LLM06:2025; CWE-272, CWE-250 \\
Excessive functionality & \texttt{owasp\_\allowbreak{}excess\_\allowbreak{}functionality} & OWASP LLM06:2025 \\
Privilege compromise / escalation & \texttt{owasp\_\allowbreak{}privilege\_\allowbreak{}escalation} & CWE-269; OWASP ASI \\
Excessive autonomy (approximation) & \texttt{unrequested\_\allowbreak{}high\_\allowbreak{}impact} & OWASP LLM06:2025 \\
Sensitive-access exposure surface & \texttt{sensitive\_\allowbreak{}access} & OWASP LLM02:2025 \\
\bottomrule
\end{tabular}
\par\medskip
\begin{tabular}{@{}>{\raggedright\arraybackslash}p{4.2cm}*{3}{>{\centering\arraybackslash}p{\dimexpr(\linewidth-4.2cm-12pt)/3\relax}}@{}}
\toprule
Method & Precision & Recall & F1 \\
\midrule
Flag all & \shortstack{0.430\\{\fontsize{6}{7}\selectfont $[0.380,0.474]$}} & \shortstack{1.000\\{\fontsize{6}{7}\selectfont $[1.000,1.000]$}} & \shortstack{0.601\\{\fontsize{6}{7}\selectfont $[0.551,0.644]$}} \\
\addlinespace[2pt]
Flag none & \shortstack{0.000\\{\fontsize{6}{7}\selectfont $[0.000,0.000]$}} & \shortstack{0.000\\{\fontsize{6}{7}\selectfont $[0.000,0.000]$}} & \shortstack{0.000\\{\fontsize{6}{7}\selectfont $[0.000,0.000]$}} \\
\addlinespace[2pt]
Risky permissions & \shortstack{0.418\\{\fontsize{6}{7}\selectfont $[0.343,0.479]$}} & \shortstack{0.564\\{\fontsize{6}{7}\selectfont $[0.463,0.667]$}} & \shortstack{0.480\\{\fontsize{6}{7}\selectfont $[0.401,0.548]$}} \\
\addlinespace[2pt]
Excess permissions & \shortstack{0.504\\{\fontsize{6}{7}\selectfont $[0.412,0.577]$}} & \shortstack{0.506\\{\fontsize{6}{7}\selectfont $[0.393,0.622]$}} & \shortstack{0.505\\{\fontsize{6}{7}\selectfont $[0.417,0.575]$}} \\
\addlinespace[2pt]
Excess functionality & \shortstack{0.538\\{\fontsize{6}{7}\selectfont $[0.464,0.598]$}} & \shortstack{0.796\\{\fontsize{6}{7}\selectfont $[0.720,0.858]$}} & \shortstack{0.642\\{\fontsize{6}{7}\selectfont $[0.579,0.694]$}} \\
\addlinespace[2pt]
Privilege escalation & \shortstack{0.811\\{\fontsize{6}{7}\selectfont $[0.659,0.947]$}} & \shortstack{0.010\\{\fontsize{6}{7}\selectfont $[0.007,0.014]$}} & \shortstack{0.020\\{\fontsize{6}{7}\selectfont $[0.013,0.028]$}} \\
\addlinespace[2pt]
Unrequested impact & \shortstack{0.633\\{\fontsize{6}{7}\selectfont $[0.592,0.682]$}} & \shortstack{0.148\\{\fontsize{6}{7}\selectfont $[0.124,0.170]$}} & \shortstack{0.240\\{\fontsize{6}{7}\selectfont $[0.208,0.269]$}} \\
\addlinespace[2pt]
Sensitive access & \shortstack{0.763\\{\fontsize{6}{7}\selectfont $[0.444,0.961]$}} & \shortstack{0.016\\{\fontsize{6}{7}\selectfont $[0.005,0.025]$}} & \shortstack{0.030\\{\fontsize{6}{7}\selectfont $[0.011,0.049]$}} \\
\addlinespace[2pt]
ASI combined & \shortstack{0.511\\{\fontsize{6}{7}\selectfont $[0.455,0.558]$}} & \shortstack{0.910\\{\fontsize{6}{7}\selectfont $[0.866,0.944]$}} & \shortstack{0.654\\{\fontsize{6}{7}\selectfont $[0.605,0.691]$}} \\
\addlinespace[2pt]
Held-out LLM judge & \shortstack{0.594\\{\fontsize{6}{7}\selectfont $[0.514,0.661]$}} & \shortstack{0.839\\{\fontsize{6}{7}\selectfont $[0.809,0.865]$}} & \shortstack{0.695\\{\fontsize{6}{7}\selectfont $[0.633,0.745]$}} \\
\addlinespace[2pt]
Oracle & \shortstack{1.000$^{\mathrm{I}}$\\{\tiny identity}} & \shortstack{1.000$^{\mathrm{I}}$\\{\tiny identity}} & \shortstack{1.000$^{\mathrm{I}}$\\{\tiny identity}} \\
\addlinespace[2pt]
\bottomrule
\end{tabular}
\caption{PRE static coverage and capability-level micro precision, recall, and F1. Each bracket is a marginal 95\% percentile interval on that one cell, source-stratified over configuration clusters, conditional on the released labels and fixed predictions. Each draw carries a configuration's capabilities together and pools TP, FP, and FN before computing the metric. A source column resamples only that source; overall preserves the six source counts. These marginal intervals support no comparison between two cells because they cannot recover their pairing; overlap establishes neither equality nor simultaneous coverage. New intervals use 10,000 draws, linear quantiles, PCG64, and base seed 20260907. Zero denominators give zero. Equal endpoints, including flag-none's $[0,0]$, are empirical degenerate intervals. $^{\mathrm{I}}$ marks a construction identity with no performance interval: declared minus minimal equals the released excess labels. The held-out judge is Llama-3.3-70B, scored on 1182 parsed configurations: 298 crewai, 215 n8n, 143 mcp, 340 injecagent, 130 sweagent, and 56 synthetic. Other rows use all 1187 (219 n8n and 144 mcp). The overall column mixes four label processes: joint LLM labels for crewai, n8n, and mcp; roster labels for injecagent; declared-minus-used labels for sweagent; and injected labels for synthetic. The privilege-escalation, unrequested-impact, and sensitive-access precision intervals retain their committed endpoints, 10,000 draws, and base seed 20260817. Standards identifiers name public categories. Scanner inputs are derived task or role tokens and declared capabilities. Generated by \texttt{tools/emit\_pre\_tables.py}; full-precision quantities and input hashes accompany the source as comments.}
\label{tab:pre-main}
\end{table*}

\begin{figure}[t]
\centering
\includegraphics[width=\textwidth]{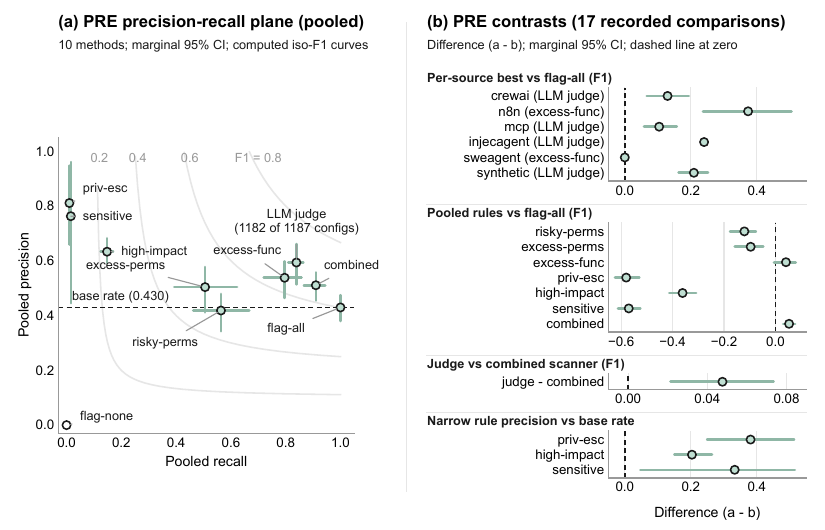}
\caption{PRE static precision-recall trade-off and recorded contrasts. (a)~Square precision-recall plane for ten scored PRE methods pooled over sources. Markers show point estimates with thin crosshair bars for marginal and uncorrected 95\% source-stratified configuration-cluster percentile intervals (Table~\ref{tab:pre-main}). The origin holds \texttt{flag\_none} as a hollow marker denoting a zero-point baseline; the oracle is omitted as a construction identity. A dashed horizontal line marks the pooled capability base rate (0.430), and grey reference curves show computed iso-F1 contours. These marginal intervals support no comparison between two cells because they cannot recover their pairing; overlap establishes neither equality nor simultaneous coverage. The held-out LLM judge is scored on 1182 parsed configurations; other methods use 1187 configurations. Pooled cells reflect capability micro-averaging where large configurations dominate the denominator. (b)~Forest of the seventeen recorded PRE contrasts across four families: selection-aware best non-oracle candidates against \texttt{flag\_all} across six sources, seven pooled rules against \texttt{flag\_all}, the held-out judge against the combined scanner on common configurations, and three narrow-rule precisions against the pooled capability base rate. Points show the estimated difference $a - b$ with marginal and uncorrected 95\% configuration-cluster percentile intervals against zero (dashed vertical line). Each block has its own x range. The \texttt{injecagent} and \texttt{synthetic} contrasts reflect dataset construction. Exact cells: Table~\ref{tab:pre-main} for cells and intervals, and Table~\ref{tab:all-contrasts} (PRE groups) for contrast rows.}
\label{fig:pre_precision_recall}
\end{figure}

Figure~\ref{fig:pre_precision_recall} visualizes the pooled precision-recall trade-off alongside the seventeen recorded PRE contrasts.

\subsection{Additional LIVE Prefix Results}
\paragraph{How the Prefixes Are Built and Which Runs Qualify.}
A finished run either failed or did not, and Section~\ref{sec:res-post-det} scores that. The
LIVE streaming board asks the harder question of whether failure is visible early, from a growing
prefix of the trace. We build the
dependency graph over the first $k$ steps at prefix fractions of 25\%, 50\%, 75\%, and 100\%, then score
each prefix three ways. Supervised scoring uses the same seed-averaged five-fold cross-validation as
POST detection. Batch-unsupervised scoring runs an off-the-shelf ECOD \citep{li2023ecod} over the run
population's prefix
flat features with no labels. Strict per-run online scoring reads a single raw structural scalar from
a run's own prefix, with no labels and no other runs. Under cross-validation the 100\% prefix is the whole
run, so that column is the POST detection board of Section~\ref{sec:res-post-det}. LIVE keeps
runs of at least
four steps, so that an early prefix is distinct from a late one, while POST keeps runs of at least
two. Every current tau-bench run passing the POST filter also clears four steps, so the two
populations coincide today rather than by construction.\looseness=-1

\begin{figure}[t]
\centering
\includegraphics[width=\textwidth]{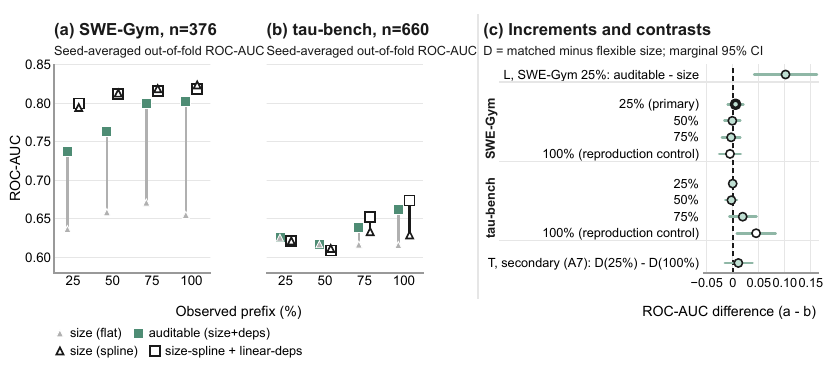}
\caption{Flexible size control and dependency increments across LIVE streaming prefixes.
(a,b)~Seed-averaged out-of-fold ROC-AUC (the table's arm cells, the P estimand; the board's mean-fold B cells in Tables~\ref{tab:live-stream} and~\ref{tab:live-stream-tau} differ from these by up to 0.008 and are not drawn) for SWE-Gym and tau-bench.
At each prefix, vertical dumbbells connect the linear pair (size (flat) to \texttt{auditable} (size+deps), filled) and the matched pair (size (spline) to size-spline + linear-deps, hollow), so each dumbbell length is an increment.
(c)~The dependency increment $D$ (matched minus flexible size) with marginal and uncorrected 95\% intervals and a dashed zero line. Every bar runs to its recorded endpoints, so an interval narrower than its marker is hidden behind it.
SWE-Gym intervals are paired DeLong and tau-bench intervals task-clustered percentile bootstrap, both conditional on the saved fits (A6); $T$ is a class-stratified paired run bootstrap.
The SWE-Gym 25\% cell is the nominated primary cell (heavier edge), and the 100\% cells are hollow reproduction controls.
Separated rows display the SWE-Gym 25\% linear increment and the secondary temporal contrast $T$ (A7).
The $0.03$ editorial magnitude that rule A8 names for the primary cell is not drawn; Table~\ref{tab:live-size-control} carries that reading.
Exact cells: Table~\ref{tab:live-size-control} for the prefix cells, intervals, and the A5, A8, and A9 roles, and Table~\ref{tab:core-contrasts}, whose 100\% effects the two hollow rows reproduce (its tau-bench interval is a separate bootstrap draw).}
\label{fig:live_size_control}
\end{figure}

\paragraph{Flexible Size Control Leaves the Early SWE-Gym Increment Unresolved.}
At the first 25\% of the run, \texttt{auditable} displays ROC-AUC 0.742 against 0.629 for
the flat baseline. The registered contrast on seed-averaged out-of-fold predictions is
$+0.101$ with a 95\% interval of $[0.041, 0.161]$ (Table~\ref{tab:all-contrasts}), and that
contrast is measured against a linear size reference. Against the matched control this benchmark
applies at the endpoint, a fixed cubic spline for size with linear dependency terms, the same
difference is $+0.005$ with $[-0.010, +0.020]$
(Table~\ref{tab:live-size-control}). The matched-control interval leaves the increment's sign
unresolved while excluding the declared $+0.03$ editorial magnitude under this conditional
analysis. Figure~\ref{fig:live_size_control} shows the linear and matched pairs alongside the resulting increments across prefixes.
ECOD displays 0.756.
Twenty nonrandom method-prefix cells are also read against the fixed 0.70 bar, two-sided, as the
tau-bench cells below are. Those readings were added after these scores were examined. Each is a
marginal statement about the one cell it concerns rather than a simultaneous statement across
methods or prefixes, and Table~\ref{tab:all-contrasts} prints each cell's distance from the bar with
the interval on that distance. The full-feature reference, \texttt{auditable} (size+deps), and ECOD
all display above 0.70 at every prefix; \texttt{size (flat)} displays below it at every prefix; and
the raw per-run dependency-span scalar climbs from 0.364 to 0.648 without reaching it, which is what
a length-confounded online statistic should do. Each bar cell is
scored on a different estimand from the board's. For a supervised cell the interval is formed on one
curve built from the per-run out-of-fold scores averaged over five split seeds, while the board
reports the mean of the twenty-five fold-level AUCs; for \texttt{auditable} at 25\% these are 0.738
and 0.742. The gap reaches 0.008 on the supervised cells and stays under 0.001 on the unsupervised
ones, which have no folds to average. Every bar reading refers to the first estimand.

\begin{figure}[t]
\centering
\includegraphics[width=\textwidth]{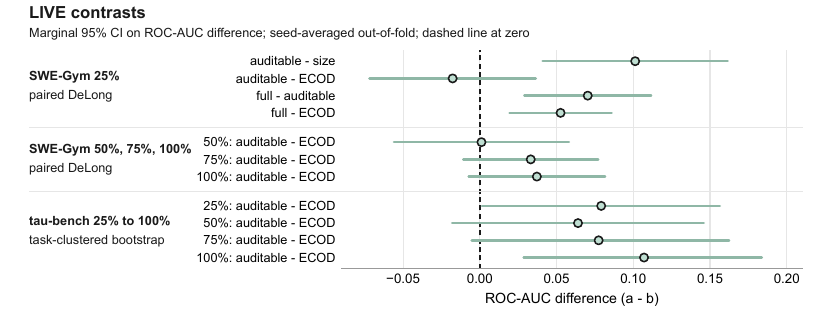}
\caption{LIVE streaming contrasts on SWE-Gym and tau-bench. Each row shows the ROC-AUC difference between two entrants ($a$ minus $b$) with its marginal and uncorrected 95\% interval. Points are dots and intervals are thin bars; the dashed vertical line marks zero difference. Rows are grouped into three blocks: the four prespecified SWE-Gym contrasts at the 25\% prefix, the three later SWE-Gym auditable against ECOD contrasts (50\%, 75\%, and 100\% prefixes), and the four tau-bench auditable against ECOD contrasts (25\% to 100\% prefixes). SWE-Gym intervals are paired DeLong on run-level sampling; tau-bench intervals are task-clustered stratified percentile bootstrap draws resampling task instances within domain. Each row's arms are the P cells (seed-averaged out-of-fold) of Tables~\ref{tab:live-stream} and~\ref{tab:live-stream-tau}, not their B cells; time to detection is not drawn. Exact cells: Table~\ref{tab:all-contrasts}.}
\label{fig:live_contrasts}
\end{figure}

\paragraph{Tau-Bench Remains Below the Warning Bar.}
On tau-bench
every method is weak and late. \texttt{auditable} (size+deps) scores 0.632, 0.617, 0.640, and 0.665 at
the four prefixes. No point estimate reaches 0.70,
so its time to detection is undefined: no prefix crosses the 0.70 bar drawn in
Figure~\ref{fig:detection}. Read that as a statement about the printed values. All twenty nonrandom
cells are read against the fixed bar, two-sided, and Table~\ref{tab:all-contrasts} prints each
cell's distance from it with the interval on that distance; each such reading is marginal and
concerns one cell. The two cells sitting closest to the bar are the full-feature reference at 0.6647
and \texttt{auditable} (size+deps) at 0.6624, both at the full trace. This is the same descriptive
domain split the detection board shows. Both corpora are read against the same bar in the same
way, and they differ in where each sits relative to it: every tau-bench cell displays below 0.70,
while the SWE-Gym cells straddle it.
Figure~\ref{fig:live_contrasts} displays the eleven recorded LIVE entrant-versus-entrant contrasts (the forty bar readings stay in Table~\ref{tab:all-contrasts}).

The four arms are the declaration's: size (flat) is the linear size reference, \texttt{auditable} (size+deps) is the linear structural reference, size (spline) is the fixed flexible size reference and size-spline + linear-deps is the matched structural control.
Arm cells are the ROC-AUC of five-seed-averaged out-of-fold failure probabilities, the quantity $D$ is defined on: ROC-AUC of the matched structural arm's five-seed-averaged out-of-fold failure probabilities minus that of the fixed flexible size arm (A5).
Table~\ref{tab:live-stream} and Table~\ref{tab:live-stream-tau} print the board's quantity instead, the mean of the fold AUCs, so their cells for the two linear arms are not these; A5: the mean of fold AUCs is a separate reproduction quantity and never receives the interval computed for the averaged-probability quantity, and both quantities are carried at full precision in this block's comment lines.
Every column is rounded on its own, so the difference of two printed arm scores can differ from the printed $D$ in the last place.
A6: folds and seeds are not independent new runs. This interval is conditional on the saved fits and does not include resample-and-refit training uncertainty.
SWE-Gym intervals are paired DeLong, conditional on the saved fits; tau-bench intervals are the task-clustered stratified percentile bootstrap over 165 task clusters (50 airline and 115 retail) at 10,000 draws.
Effects and endpoints are printed to four decimals because several are smaller than a thousandth, and the record puts the Monte Carlo noise on one clustered endpoint at 0.002 at that tau-bench draw count, so a tau-bench endpoint is printed to more digits than one draw of the bootstrap carries.
Each row also carries a second interval in this block's comment lines, SWE-Gym a crosscheck only and tau-bench a reproduction diagnostic under the independence assumption; neither is a reported uncertainty.
A9: the 100 percent effects reproduce a cell whose value is already committed. They are reproduction controls, not independent new evidence.
The Role row carries that label beside the cells it applies to.
The cell A5 nominated before any of the eight was computed is SWE-Gym at 25\%: $D$ = $+0.0052$, $[-0.0099, +0.0204]$.
Its branch, by the rule A8 froze before the run, is unresolved.
A8: substantial-positive when the point is at least +0.03 AND the marginal interval excludes zero; small-positive when the interval excludes zero on the positive side at a smaller point; erased-or-reversed when it excludes zero on the negative side; unresolved when it contains zero.
An interval containing zero is unresolved whatever the point is, and is reported as unresolved rather than as a small effect.
The upper endpoint sits below the declared 0.03, which rules out that planning magnitude under this conditional analysis and does not establish zero information.
The 0.03 threshold is a declared editorial judgment about what would be a worthwhile result for this paper, stated as such in A8; not a validated operational threshold.
$T$ is the secondary temporal contrast A7 declared before the batch: 10,000 draws of a class-stratified paired run percentile bootstrap carrying all four score vectors in every draw, no additional classifier fit.
It is secondary. A7: it never replaces a failed primary result, and if D(swegym, 0.25) is unresolved a positive T does not rescue an early-signal claim.
A4: row-level cross-validation on tau-bench is the existing evaluation convention. The clustered interval describes sampling uncertainty and does not convert the evaluation into an unseen-task evaluation.
A3: the prefix fractions index a retrospective sweep over final trace length. A positive result establishes neither a deployable clock-time alarm nor knowledge of the eventual horizon during execution.
A9: no new Holm family and no search over prefixes. These are marginal readings with one nominated primary quantity, and the 138-record registry is unchanged.
A2 and the chronology section: this extension was proposed after the linear LIVE prefix scores and the POST specification-sensitivity results were already in the manuscript. It stays labelled exploratory whichever way it comes out.
A11: one declared fit batch and one reporting pass. A disappointing effect is not an implementation defect and is not grounds for a rerun.
The batch recorded no failures and no warnings.
The A10.4 preflight did not clear on its own terms and admitted this batch under the declaration's dated entry, with two disagreements in the \texttt{variance\_axes.board\_point\_estimate.value} field of \texttt{statistical\_tests\_results.json} carried forward as findings about that record; the batch reads feature matrices and fits arms and consumes no field of it.
Generated by \texttt{tools/emit\_live\_size\_audit\_table.py}.

\begin{table}[t]
\centering
\scriptsize
\setlength{\tabcolsep}{4pt}
\caption{At each LIVE prefix, how much of the dependency increment survives once the size reference is allowed to bend: the ROC-AUC of the matched structural arm minus that of the fixed flexible size arm.}
\label{tab:live-size-control}
\begin{tabular}{@{}lrrrr@{}}
\toprule
Arm or quantity & 25\% & 50\% & 75\% & 100\% \\
\midrule
\multicolumn{5}{@{}l}{\textbf{SWE-Gym}, 376 runs, 188 failed} \\
size (flat) & 0.637 & 0.659 & 0.671 & 0.655 \\
\texttt{auditable} (size+deps) & 0.738 & 0.763 & 0.800 & 0.803 \\
size (spline) & 0.794 & 0.813 & 0.819 & 0.824 \\
size-spline + linear-deps & 0.800 & 0.812 & 0.816 & 0.818 \\
$D$ (matched $-$ flexible size) & $+0.0052$ & $-0.0009$ & $-0.0032$ & $-0.0054$ \\
95\% interval for $D$ & $[-0.0099, +0.0204]$ & $[-0.0163, +0.0145]$ & $[-0.0209, +0.0144]$ & $[-0.0261, +0.0153]$ \\
Role & primary & declared cell & declared cell & reproduction control \\
\midrule
\multicolumn{5}{@{}l}{\textbf{tau-bench}, 660 runs, 363 failed} \\
size (flat) & 0.626 & 0.617 & 0.617 & 0.616 \\
\texttt{auditable} (size+deps) & 0.625 & 0.617 & 0.639 & 0.662 \\
size (spline) & 0.621 & 0.612 & 0.633 & 0.629 \\
size-spline + linear-deps & 0.621 & 0.609 & 0.652 & 0.674 \\
$D$ (matched $-$ flexible size) & $-0.0002$ & $-0.0027$ & $+0.0190$ & $+0.0447$ \\
95\% interval for $D$ & $[-0.0007, +0.0003]$ & $[-0.0144, +0.0080]$ & $[-0.0066, +0.0455]$ & $[+0.0077, +0.0823]$ \\
Role & declared cell & declared cell & declared cell & reproduction control \\
\midrule
\multicolumn{5}{@{}l}{Secondary (A7): $T$ = SWE-Gym $D$ at 25\% minus $D$ at 100\% = $+0.0107$, $[-0.0153, +0.0379]$} \\
\bottomrule
\end{tabular}
\end{table}

\begin{table}[t]
\centering
\scriptsize
\setlength{\tabcolsep}{4pt}
\caption{For each committed live.tau.bar cell, the number of tau-bench task clusters at which a 95\% interval would exclude zero at the observed point estimate.
The present sample is 165 task clusters and 660 runs, 50 airline and 115 retail.
Eighteen of the twenty cells already separate from the 0.70 bar at that sample and two do not: 100.auditable (size+deps) at 434 clusters (1,736 runs, 2.6 times the present sample, 412 clusters under the closed form); 100.full at 410 clusters (1,640 runs, 2.5 times the present sample, 408 clusters under the closed form).
Point is the committed ROC-AUC minus the 0.70 bar, SD the committed clustered bootstrap standard deviation, and the interval and verdict the committed ones; Table~\ref{tab:live-stream-tau} prints the same cells on the same scale.
Measured is the number this table reads. Not every measured exponent is consistent with the reference, so the printed figure comes from the measured scaling and the closed form stands beside it as the idealized comparison.
Closed form assumes what the declaration set out to check, that the clustered bootstrap SD falls as clusters ** -0.5.
A row marked $^{\dagger}$ had its own scaling ladder run (Table~\ref{tab:tau-scaling}) and uses that cell's own exponent; every other row uses a bracket across the four measured exponents, reported at the end that asks for more clusters; the four do not agree, so no single measured exponent is licensed for a cell whose own ladder was not run.
A count marked $^{\ast}$ falls below 40 clusters, the smallest rung the ladder measured, so it extrapolates outside the range where the scaling was checked.
The cluster counts are design targets for a future study on this benchmark. They are not a claim that a cell would clear the bar at that sample: the solve holds the observed point estimate fixed, and a larger sample can move it either way.
Every row extrapolates the SD alone. Nothing here re-estimates a point, a label, or a threshold.
The solve uses a symmetric normal interval at the committed bootstrap SD. The printed intervals are percentile bootstrap intervals and are asymmetric, so a solved count is an approximation of the sample at which the percentile interval would clear zero, not a simulation of it.
Achieved power: never computed. Power at the sample already run is a monotone restatement of the observed p-value and adds no information about what a future study would resolve.
Minimum detectable difference: never computed. Converting an achieved interval into the smallest effect it would have excluded fixes the sample and solves for the effect, which is the direction Part B3 forbids; this file fixes the effect at the observed point and solves for the sample.
Part B was proposed after the twenty committed cells and their verdicts were read. It stays exploratory in the manuscript whichever way it comes out.
Generated by \texttt{tools/emit\_tau\_design\_table.py}.}
\label{tab:tau-design}
\begin{tabular}{@{}llrrclrr@{}}
\toprule
& & & & & & \multicolumn{2}{c}{Clusters needed} \\
\cmidrule(l){7-8}
Prefix & Method & Point & SD & 95\% interval & Verdict & Measured & Closed form \\
\midrule
25\% & size (flat)$^{\dagger}$ & $-0.074$ & 0.030 & $[-0.134, -0.017]$ & separates & 103 & 104 \\
25\% & \texttt{auditable} (size+deps) & $-0.075$ & 0.030 & $[-0.134, -0.016]$ & separates & 103 & 103 \\
25\% & full & $-0.065$ & 0.030 & $[-0.126, -0.007]$ & separates & 137 & 137 \\
25\% & PyOD (ECOD, unsup.)$^{\dagger}$ & $-0.154$ & 0.029 & $[-0.209, -0.097]$ & separates & 19$^{\ast}$ & 23$^{\ast}$ \\
25\% & dep-span (online) & $-0.197$ & 0.002 & $[-0.200, -0.193]$ & separates & 1$^{\ast}$ & 1$^{\ast}$ \\
\midrule
50\% & size (flat) & $-0.083$ & 0.031 & $[-0.145, -0.021]$ & separates & 92 & 92 \\
50\% & \texttt{auditable} (size+deps) & $-0.083$ & 0.031 & $[-0.145, -0.023]$ & separates & 91 & 91 \\
50\% & full & $-0.072$ & 0.030 & $[-0.132, -0.015]$ & separates & 106 & 106 \\
50\% & PyOD (ECOD, unsup.) & $-0.147$ & 0.030 & $[-0.206, -0.087]$ & separates & 27$^{\ast}$ & 27$^{\ast}$ \\
50\% & dep-span (online) & $-0.170$ & 0.012 & $[-0.193, -0.146]$ & separates & 4$^{\ast}$ & 4$^{\ast}$ \\
\midrule
75\% & size (flat) & $-0.083$ & 0.032 & $[-0.147, -0.022]$ & separates & 95 & 95 \\
75\% & \texttt{auditable} (size+deps) & $-0.061$ & 0.030 & $[-0.121, -0.004]$ & separates & 151 & 151 \\
75\% & full & $-0.057$ & 0.029 & $[-0.117, -0.003]$ & separates & 163 & 163 \\
75\% & PyOD (ECOD, unsup.) & $-0.138$ & 0.030 & $[-0.197, -0.080]$ & separates & 31$^{\ast}$ & 31$^{\ast}$ \\
75\% & dep-span (online) & $-0.135$ & 0.026 & $[-0.188, -0.086]$ & separates & 24$^{\ast}$ & 24$^{\ast}$ \\
\midrule
100\% & size (flat) & $-0.084$ & 0.032 & $[-0.146, -0.023]$ & separates & 91 & 91 \\
100\% & \texttt{auditable} (size+deps)$^{\dagger}$ & $-0.038$ & 0.030 & $[-0.099, 0.020]$ & does not separate & 434 & 412 \\
100\% & full$^{\dagger}$ & $-0.035$ & 0.028 & $[-0.093, 0.019]$ & does not separate & 410 & 408 \\
100\% & PyOD (ECOD, unsup.) & $-0.145$ & 0.030 & $[-0.204, -0.086]$ & separates & 27$^{\ast}$ & 28$^{\ast}$ \\
100\% & dep-span (online) & $-0.132$ & 0.034 & $[-0.199, -0.067]$ & separates & 42 & 42 \\
\bottomrule
\end{tabular}
\end{table}

\begin{table}[t]
\centering
\footnotesize
\setlength{\tabcolsep}{5pt}
\caption{The measured scaling of the clustered bootstrap standard deviation, on the four cells the declaration named before any subsample was drawn.
The assumption under test is that the clustered bootstrap SD falls as clusters ** -0.5, against a reference exponent of -0.5.
Ladder: 40, 60, 80, 100, 120, 140 and 165 clusters; the airline and retail proportions of the full sample are held at every rung; twenty subsamples per rung and 2,000 bootstrap draws inside each subsample; base seed 20260907, RNG label \texttt{live\_tau\_scaling.\{cell\}.\{n\_clusters\}.\{replicate\}}.
Each exponent is OLS of the rung mean of log SD on log clusters; a Student-t interval on five degrees of freedom.
Consistency rule, fixed before the fits: an exponent is consistent with -0.5 when its 95\% interval contains -0.5.
Pooling rule, fixed before the fits: the four agree when every pair of intervals overlaps and Cochran's Q does not reject homogeneity at 0.05.
Every pair of exponent intervals overlaps, and Cochran's Q is 14.90 on 3 degrees of freedom (p = 0.0019, alpha 0.05).
No pooled exponent is reported: the four per-cell exponents do not agree, so the declaration's condition for reporting a single pooled exponent is not met.
Two subsamples at an upper rung overlap heavily, and at 165 clusters all twenty replicates are the same subsample, so a rung mean is more precise in the arithmetic than in the sampling. The per-cell errors are therefore likely too small and Cochran's Q correspondingly too large, which is a reason to read the disagreement as mild rather than a reason to overrule it. The pooling rule was fixed before the fits and is reported as it came out.
These are the rows Table~\ref{tab:tau-design} marks $^{\dagger}$.
Generated by \texttt{tools/emit\_tau\_design\_table.py}.}
\label{tab:tau-scaling}
\begin{tabular}{@{}lp{0.30\linewidth}rcc@{}}
\toprule
Cell & Declared because & Exponent & 95\% interval & Consistent \\
\midrule
100.full & does not separate at the present sample & $-0.497$ & $[-0.512, -0.483]$ & yes \\
100.auditable (size+deps) & does not separate at the present sample & $-0.474$ & $[-0.493, -0.454]$ & no \\
25.size (flat) & control: separates, mid-range effect & $-0.489$ & $[-0.513, -0.465]$ & yes \\
25.pyod (ECOD) & control: separates, largest effect on the ladder & $-0.459$ & $[-0.483, -0.436]$ & no \\
\bottomrule
\end{tabular}
\end{table}

\begin{figure}[t]
\centering
\includegraphics[width=\textwidth]{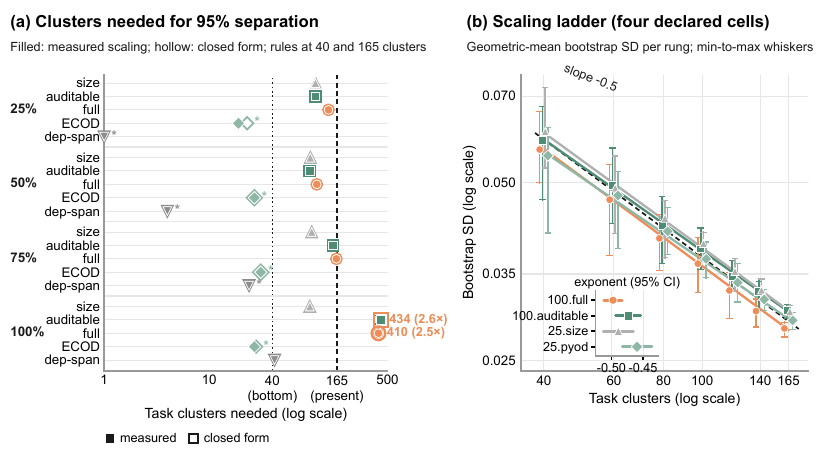}
\caption{Task clusters needed for a 95\% interval to exclude zero at the observed point estimate across twenty committed tau-bench cells, ordered by prefix and method.
(a)~Filled markers indicate measured scaling; hollow markers indicate the closed-form count ($N^{-0.5}$), joined by a tick when they differ.
Dashed vertical rule marks the present sample (165 clusters); dotted vertical rule marks the ladder floor (40 clusters).
Counts below 40 clusters are marked with an asterisk ($^\ast$) as extrapolations below the checked ladder range.
The two cells that do not separate at the present sample (100\% auditable and 100\% full) are ringed in coral with their counts (434 and 410) and sample multiples printed.
(b)~Scaling ladder of the clustered bootstrap standard deviation across the four declared cells (100.full, 100.auditable, 25.size, 25.pyod) on log-log axes. Each rung shows the geometric-mean standard deviation across twenty subsamples with min-to-max whiskers (cells are offset horizontally for legibility); the fitted lines are ordinary least squares, and the dashed reference line has slope $-0.5$.
Inset: fitted exponents with 95\% Student-t intervals on five degrees of freedom against the $-0.5$ reference line.
No pooled exponent is reported because the four exponents do not agree under Cochran's Q ($p = 0.0019$).
The four declared cells (the rows marked $^\dagger$ in Table~\ref{tab:tau-design}) use their own measured exponents; the remaining sixteen rows use the bracket exponent.
These counts are design targets holding the observed point estimates fixed, not a claim that a cell would clear the bar at that sample.
Exact cells: Table~\ref{tab:tau-design} and Table~\ref{tab:tau-scaling}.}
\label{fig:tau_design}
\end{figure}

Figure~\ref{fig:tau_design} visualizes the task-cluster counts at which each tau-bench cell's 95\% interval would exclude the 0.70 bar at its observed point, under measured and closed-form scaling, alongside the standard-deviation scaling ladder on the four declared cells.

\subsection{Additional POST Localization Results}
\paragraph{Model Identity Decides This Board.} The field-standard control asks an LLM to name the
decisive mistake step from a failed run. For our all-at-once panel, every step is represented, with
whitespace normalized and each step's content capped at 1500 characters. This truncates 268 of
1099 steps across 108 of the 126 runs. We run
this control as an eleven-model panel. The released \texttt{tools/llm\_judge\_codex.py} documents
the GPT-5.5 reference as generated through the Codex subscription CLI (\texttt{codex exec}).
Its historical cache does not record the channel. The panel runner supports the NAIRR gateway for
GPT-5.4, Claude-Opus-4.8, and Gemini, and AWS Bedrock for open-weights and small proprietary models
(Llama-3.3-70B, Qwen3-32B, DeepSeek-R1, and four smaller models). Each is prompted once over the
trace
in the Who\&When all-at-once
protocol \citep{zhang2025whoandwhen}. Predictions are cached and committed, so the board scores them
deterministically and re-running the benchmark needs no API call. Figure~\ref{fig:localization-panel} scores the panel
against Who\&When's human mistake-step labels, alongside the structural methods and the three
elicitation protocols. Exact Top-1, Top-3, and MRR cells are in
Tables~\ref{tab:loc} and~\ref{tab:protocol} of Appendix~\ref{app:board-values}.\looseness=-1

\begin{figure}[t]
\centering
\includegraphics[width=\textwidth]{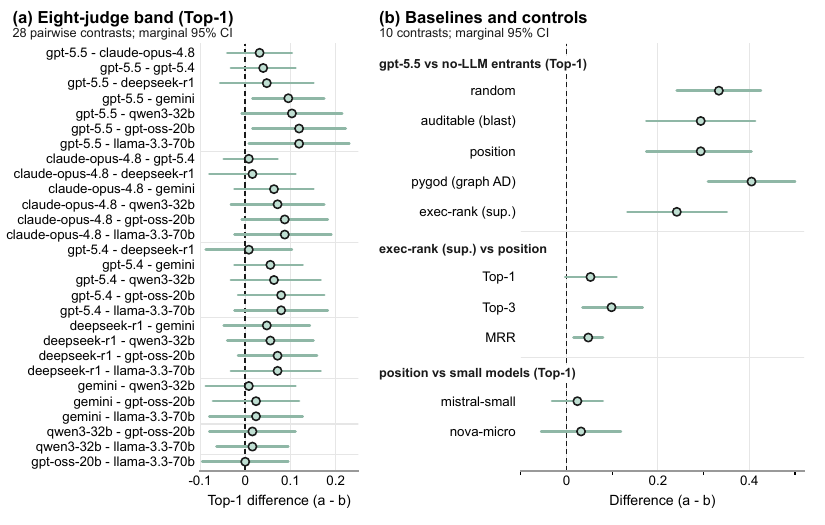}
\caption{Registered fault localization contrasts on Who\&When (126 failed runs, human mistake-step labels). (a)~All 28 pairwise Top-1 differences among the eight judges in the main band under the all-at-once protocol, ordered by the first judge. (b)~Ten baseline, ranking, and prior contrasts: GPT-5.5 against the five no-LLM entrants (Top-1), exec-rank (sup.) against the position prior across Top-1, Top-3, and MRR, and the position prior against two smaller models (Top-1). Each row shows a point estimate (circle) with its marginal 95\% percentile bootstrap interval (bar); the dashed vertical rule marks zero difference. Intervals are marginal and uncorrected; overlap between rows does not constitute a paired comparison. Elicitation protocol contrasts are omitted here; see Figure~\ref{fig:localization-panel}(b) and Table~\ref{tab:protocol} for them. Exact cells: Table~\ref{tab:all-contrasts} (localization groups) and Table~\ref{tab:loc}.}
\label{fig:localization_contrasts}
\end{figure}

\paragraph{Strong Judges Cluster in One Narrow Band.}
Three things stand out. The LLM judges are the strongest localizers here, and eight of them fall
inside a band from 0.333 to 0.452 Top-1 at $n = 126$; Table~\ref{tab:all-contrasts} carries all 28
pairwise differences in that band with their intervals. GPT-5.5 holds the top of that
band at 0.452. Its all-at-once input represents every step, with whitespace normalized and content
capped at 1500 characters per step. At this cap, its Top-1 score exceeds every structural method on
this board. The panel spans 0.127 to 0.452. It does not
divide into frontier and open-weight tiers, since the open-weights DeepSeek-R1 at 0.405 sits above
the lowest frontier judge. The smallest models sit below the trivial position prior on point
estimate, with Mistral at 0.135 and Nova at 0.127 against 0.159, and their displayed cells rise
above the prior under the two other
elicitation protocols of Table~\ref{tab:protocol}. Thus, ``just ask an LLM'' is only as good as
the LLM. Figure~\ref{fig:localization_contrasts} draws 38 of the 58 registered localization contrasts with their marginal 95\% intervals; the 20 elicitation-protocol rows are in Figure~\ref{fig:localization-panel}(b).

\paragraph{Where the Later Judges Land.}
Later releases were measured in two passes, both on the same 126 Who\&When runs under the declared
capped all-at-once protocol (Table~\ref{tab:judge-addendum}). The 2026-09-06 addendum scored
\texttt{claude-opus-5} and \texttt{gpt-6-astra} through subscription CLIs, and \texttt{llama-3-70b}
and \texttt{llama-3.1-70b} through Bedrock. Its five-model list, its Top-1 definition, its cache
keys and its retry budget were all fixed before that pass began. \texttt{gpt-5.6-sol} was declared
that day and omitted from it, and the recorded reason is a decision rather than an obstacle: before
either OpenAI score existed, the authors preferred \texttt{gpt-6-astra} as the more advanced model
in the family, and left unset the SSH forward that \texttt{gpt-5.6-sol}'s only route needs. The 2026-09-12
gateway declaration was written after those CLI scores were known, and it enumerates three further
gateway arms. The reference reruns and the two same-channel repeats came after those three had run,
so they are post-hoc additions rather than declared ones. Both models of the OpenAI pair now carry
a score, and none of these runs enters the nine boards, the 72 entrants or the 138 recorded
comparisons.

On the gateway, \texttt{gpt-5.4}, \texttt{gpt-5.5}, \texttt{gpt-5.6-sol} and \texttt{gpt-6-astra}
score 0.405, 0.429, 0.429 and 0.476 Top-1, and \texttt{claude-opus-4.8} and \texttt{claude-opus-5}
score 0.421 and 0.452. At fixed 70B, Llama 3, 3.1 and 3.3 score 0.325, 0.294 and 0.333, an
observed sequence that is not monotone in generation. These are descriptive measurements with
marginal per-arm intervals. The block declares no contrast and resolves no difference among these
arms, and it establishes neither equal judge performance nor a ceiling imposed by the corpus.

Repeating a run on one fixed channel already moves the answer. Two gateway runs of
\texttt{claude-opus-5} both score 0.452 Top-1 and name a different step on 9 of the 126 runs. The
two \texttt{gpt-6-astra} gateway runs score 0.476 and 0.500 and differ on 6. The CLI cache and the
first gateway cache differ on 11 runs for \texttt{claude-opus-5} and on 13 for \texttt{gpt-6-astra}.
Gateway caches differ from the corresponding published caches on 7 runs for
\texttt{claude-opus-4.8}, 14 for \texttt{gpt-5.4} and 19 for \texttt{gpt-5.5}. That 7 lies between
the two observed repeat counts of 6 and 9. These comparisons are between cached answers and do not
isolate a route effect, and for most of these pairs the published channel is unrecorded, so what
differs between the two caches is itself not established.

The gateway was expected to record an exact serving that a subscription CLI cannot, and for the
OpenAI arms it did not: the response named the deployment rather than a dated version. The
2026-09-12 declaration describes the Claude gateway name as pinned and the two Azure deployments as
mutable, which is a statement about the declared configuration. No cache or reconstructed sidecar
here retains an observed served checkpoint version, and a returned model identifier is not one. The
2026-09-06 declaration specified a family of paired tests with a multiplicity correction across it. That
apparatus was retired for the whole paper when the comparison registry left the main text, and it is
not reinstated for this block, which reports each arm's own score with per-arm uncertainty, in
descending order and without a rank.

\begin{table}[t]
\centering
\footnotesize
\setlength{\tabcolsep}{3pt}
\caption{Judge addendum on POST localization: the same 126 Who\&When runs and all-at-once protocol as the eleven published judges.
Ten addendum arms are shown with the published \texttt{gpt-5.5} and \texttt{llama-3.3-70b} references.
Top-1 and Top-3 carry marginal 95\% Procedure BIN intervals, and the references' Top-1 cells repeat Table~\ref{tab:protocol}.
This block declares no registered contrast and tests no difference between any two arms; interval overlap establishes neither equivalence nor the absence of an improvement.
Rows run in descending Top-1 order and carry no rank: no contrast is declared here and no difference between arms is tested.
A Top-1 hit means the gold step ranks first after a stable sort of the per-step scores, as on the published board; a null output takes the all-zero ranking and can still take credit on the 20 runs labelled at step 0.
The addendum is excluded from the nine boards, the 72 entrants and the 138 recorded comparisons.
Two arms were generated a second time on the same channel and are held out of this table; those repeats name a different step on 9 and 6 of the runs, which is the floor any reading across routes or generations has to clear.
\texttt{gpt-5.6-sol} was declared on 2026-09-06 and not run then. Before either OpenAI score existed, the authors preferred \texttt{gpt-6-astra} as the more advanced model in the family and left the required SSH forward unset: a configuration decision, not model unavailability. \texttt{gpt-5.6-sol} was run over that forward on 2026-09-12 and its row here carries that measurement.
Addendum channels are reconstructed from declared configurations; \texttt{not-recorded} marks an absent channel record.
Mean reciprocal rank is carried in this block's comment lines rather than printed, so every printed score carries its interval.
Generated by \texttt{tools/emit\_addendum\_table.py}.}
\label{tab:judge-addendum}
\begin{tabular}{@{}lllcc@{}}
\toprule
Model & Channel & Source & Top-1 & Top-3 \\
\midrule
\texttt{claude-opus-5} & \texttt{claude-code-cli} & addendum & 0.476 {\scriptsize $[0.389, 0.563]$} & 0.786 {\scriptsize $[0.714, 0.857]$} \\
\texttt{gpt-6-astra} & \texttt{nairr-gateway} & addendum & 0.476 {\scriptsize $[0.389, 0.563]$} & 0.762 {\scriptsize $[0.690, 0.833]$} \\
\texttt{gpt-5.5} & \texttt{codex-cli} & published & 0.452 {\scriptsize $[0.365, 0.540]$} & 0.667 {\scriptsize $[0.587, 0.746]$} \\
\texttt{claude-opus-5} & \texttt{nairr-gateway} & addendum & 0.452 {\scriptsize $[0.365, 0.540]$} & 0.778 {\scriptsize $[0.706, 0.849]$} \\
\texttt{gpt-6-astra} & \texttt{codex-cli} & addendum & 0.452 {\scriptsize $[0.365, 0.540]$} & 0.794 {\scriptsize $[0.722, 0.857]$} \\
\texttt{gpt-5.5} & \texttt{nairr-gateway} & addendum & 0.429 {\scriptsize $[0.341, 0.516]$} & 0.730 {\scriptsize $[0.651, 0.802]$} \\
\texttt{gpt-5.6-sol} & \texttt{nairr-gateway} & addendum & 0.429 {\scriptsize $[0.341, 0.516]$} & 0.722 {\scriptsize $[0.643, 0.802]$} \\
\texttt{claude-opus-4.8} & \texttt{nairr-gateway} & addendum & 0.421 {\scriptsize $[0.333, 0.508]$} & 0.722 {\scriptsize $[0.643, 0.802]$} \\
\texttt{gpt-5.4} & \texttt{nairr-gateway} & addendum & 0.405 {\scriptsize $[0.317, 0.492]$} & 0.706 {\scriptsize $[0.627, 0.786]$} \\
\texttt{llama-3.3-70b} & \texttt{not-recorded} & published & 0.333 {\scriptsize $[0.246, 0.413]$} & 0.579 {\scriptsize $[0.492, 0.667]$} \\
\texttt{llama-3-70b} & \texttt{aws-bedrock} & addendum & 0.325 {\scriptsize $[0.246, 0.413]$} & 0.611 {\scriptsize $[0.524, 0.698]$} \\
\texttt{llama-3.1-70b} & \texttt{aws-bedrock} & addendum & 0.294 {\scriptsize $[0.214, 0.373]$} & 0.603 {\scriptsize $[0.516, 0.683]$} \\
\bottomrule
\end{tabular}
\end{table}

\paragraph{Execution Features Lead the Structural Methods.}
Among methods that use no LLM, \texttt{exec-rank (sup.)}, the supervised execution-feature ranker
reproduced from GRADE \citep{zhao2026grade}, holds the highest point estimate on all three metrics. On
the same 126-run split it scores 0.211 Top-1 and 0.614 Top-3, against 0.159 and 0.516 for the position
prior: a Top-1 gap of 6.6 runs out of 126, and a wider gap on Top-3. Table~\ref{tab:all-contrasts}
carries both differences and the one on MRR with their intervals. GRADE evaluated that structural comparison
without an LLM judge, on the cross-validation split axis. \sysname places the same ranker in the
wider comparison of Table~\ref{tab:loc}, where GPT-5.5 holds the top of the judge band at 0.452, and
adds run-level sampling variation to the picture. The full-context dependency prior coincides with
position because Who\&When assumes full-context dependencies; the two rank identically on all 126
runs.

\paragraph{How the Judge Is Elicited Barely Matters.} We compare three protocols adapted from
Who\&When \citep{zhang2025whoandwhen}. All-at-once represents every step in one call, with
whitespace
normalized and each step's content capped at 1500 characters. Step-by-step reveals growing prefixes
with normalized content capped at 1200 characters per visible step, asking whether the decisive
mistake has already occurred. Binary-search
halves the suspect interval through repeated first-half or second-half judgments. Each call
retains every step index, agent, and kind, with normalized content capped at 900 characters per
step.
Step-by-step changes the visible prefix; binary-search changes the candidate interval while
retaining this capped rendering. Because the content caps also differ, the comparison changes
both the elicitation rule and the per-step evidence budget.
Table~\ref{tab:protocol} scores all three on the same 126 runs. All-at-once leads or ties for seven of
the ten models on point estimates, and the movement elsewhere is small at this corpus size;
Table~\ref{tab:all-contrasts} carries each model's two protocol differences with their intervals.
DeepSeek-R1 and Gemini score identically under all-at-once and binary-search, although the two
protocols name the same step on roughly half of runs, so those ties reflect symmetric disagreement
rather than convergent answers. The two largest step-by-step losses belong to mid-tier open models
(Llama 0.333 to 0.222, Qwen 0.349 to 0.254), beside DeepSeek-R1's 0.405 to 0.317; the record carries
each model's protocol change on its own and no comparison between models. The
weakest models are higher under at least one alternative (Mistral 0.135 to 0.214 under binary-search,
Nova 0.127 to 0.167 under either, Gemma 0.206 to 0.230 under step-by-step but 0.206 to 0.159 under
binary-search). The largest single protocol movement
on this board is Qwen3-32B under binary-search, 0.349 to 0.127. Its prompt
represents every step, with whitespace normalized and each step's content capped at 900 characters.
We report all-at-once, with its 1500-character content cap per step, as the headline LLM-judge
reference because it
is Who\&When's default and costs one call per run against $\log_2 n$ or $n$ for the alternatives. It
is also the only protocol that returns a ranked shortlist rather than a single step, so it is the only
one for which Top-3 and MRR are defined. The mixed, non-monotonic gaps caution against over-reading any single
elicitation as the LLM-judge number.\looseness=-1

On this localization board, the no-LLM entrants score below GPT-5.5. Separate structural entrants
score incomplete prefixes on the LIVE boards (Section~\ref{sec:res-live}). Those evaluations do
not compare their timing or accuracy with the POST judges. The all-at-once POST judge reads
normalized content capped at 1500 characters per step. This board sets the LLM-judge reference.

\subsection{Additional POST Detection Results}
\paragraph{The Structural Contrast Reproduces GRADE.}
The \texttt{auditable} (size+deps) versus \texttt{size (flat)} contrast on this board is not a new
result and we do not claim it as one. Its recorded paired comparison applies DeLong to seed-averaged
out-of-fold run scores, so the quantity that comparison estimates need not equal the difference of the mean
per-seed board values printed in this paper. It reproduces GRADE \citep{zhao2026grade}, on GRADE's corpora,
using GRADE's feature construction and evaluation code, which \sysname imports rather than
reimplements. We report it because a benchmark has to establish that its reference method behaves as
published before the rest of the board means anything, and because every other entrant here is scored
against the same fixed task. What the arena adds is the comparison GRADE did not run: the two
published agent-specific detectors in Table~\ref{tab:det} and the wider unsupervised field in
Section~\ref{sec:res-transfer}. It adds the prefix decomposition in Section~\ref{sec:res-live}, where an
off-the-shelf unsupervised detector displays above this block at the first
prefix.\looseness=-1

\begin{table}[t]
\centering
\caption{Failure detection. ROC-AUC on SWE-Gym and tau-bench. GUARDIAN and G-Safeguard are simplified
adaptations of published agent-specific detectors, described in Section~\ref{sec:benchmark}; the wider PyOD / PyGOD detector arena is
summarized in Section~\ref{sec:res-transfer}. $\dagger$ marks the unsupervised PyGOD row, whose displayed
cell is a single initialization seed; its twenty-seed spread is 0.631 $\pm$ 0.164 on SWE-Gym
and 0.513 $\pm$ 0.042 on tau-bench. The supervised rows are also refit per seed and are reported with
their own spreads in the text.
Point estimates only. Several cells here could carry a marginal interval from committed data and the
rest could not without rerunning the scorer, and the supervised cells cannot borrow the pooled
intervals printed for their prefix counterparts, which are centred on a different quantity. Rather
than mark part of the table, none of it is marked.}
\label{tab:det}
\begin{tabular}{lcc}
\toprule
Method & SWE-Gym & tau-bench \\
\midrule
random & 0.483 & 0.498 \\
size (flat) & 0.663 & 0.619 \\
PyOD-flatten (ECOD) & 0.765 & 0.555 \\
PyGOD-DOMINANT (graph AD)$^{\dagger}$ & 0.547 & 0.550 \\
GUARDIAN (recon-AE) & 0.767 & 0.542 \\
\texttt{auditable} (size+deps) & 0.804 & 0.665 \\
full & 0.819 & 0.665 \\
G-Safeguard (sup GNN) & 0.828 & 0.626 \\
\bottomrule
\end{tabular}
\end{table}

\paragraph{The Structural Gain Is Specification-Dependent on SWE-Gym.}
Table~\ref{tab:det} reports the headline methods, including simplified adaptations of two published
agent-specific detectors: GUARDIAN, whose unsupervised reconstruction autoencoder enters here without
its adjacency-reconstruction and information-bottleneck terms \citep{zhou2025guardian}, and
G-Safeguard, whose supervised graph-message-passing idea enters over structural features as a
run-level failure classifier, without its agent-localization or remediation stages
\citep{wang2025gsafeguard}. The \texttt{auditable} (size+deps) lift over \texttt{size (flat)} is
$+0.142$ on SWE-Gym (0.804 against 0.663) and $+0.046$ on tau-bench (0.665 against 0.619). These
increments use unrounded scores. The two
carry very different weight, and not in the direction those two numbers suggest. Each of those two
lifts is the matched comparison that adds dependency features within the linear specification.
Repeating it inside the other two specifications gives SWE-Gym increments of $-0.010$ with a fixed
cubic spline for size and linear dependency terms, and $+0.039$ with boosted trees. On SWE-Gym the
increment therefore changes sign across the tested specifications. The tau-bench increments from
those same two specifications are $+0.045$ and $+0.049$, which agree in sign and in magnitude with
its linear $+0.046$. Tau-bench is the stronger case here.
Section~\ref{sec:res-post-det} reports both. On SWE-Gym the supervised graph network holds the
highest supervised point estimate, 0.828, with the single-seed PyGOD GAAN entry displaying higher at
0.850 on the seed-unstable axis described below. Across five seeds that jointly set its cross-validation split and its network
initialization it reaches 0.824 $\pm$ 0.007, and across five cross-validation split seeds the full
feature model reaches 0.819 $\pm$ 0.005. The two share their cross-validation splits seed by seed, so
the difference can be taken within seed, and the interval on that within-seed difference includes zero.

\paragraph{Tau-Bench Displays the Two Feature Sets at the Same Score.}
On tau-bench
\texttt{auditable} (size+deps) and the full-feature reference both round to 0.665 at three decimals,
and Table~\ref{tab:all-contrasts} carries the difference between them with its interval. They are not
interchangeable either. The two feature sets disagree on a sixth of the run pairs that define the
score, so the equal displayed values summarize partly different signal. Lower scores across
methods retain the
domain split seen on the other boards.\looseness=-1

OpenHands / SWE-rebench carries 600 runs, 312 failed and 288 solved, over 574 distinct issues, 548 attempted once and 26 attempted twice, and 0 rows with no issue identifier.
ScienceWorld carries 128 runs, 64 failed and 64 solved.
POST only: at least 2 parsed steps. The LIVE four-step filter is not applied. Label 1 means failure.
The four arms are the declaration's: size (flat) is the linear size reference, \texttt{auditable} (size+deps) is the linear structural reference, size (spline) is the fixed flexible size reference and size-spline + linear-deps is the matched structural control.
Arm cells are the ROC-AUC of five-seed-averaged out-of-fold failure probabilities, which is the quantity the three estimands are defined on: section E defines the estimands on the seed-averaged out-of-fold quantity. The mean of the 25 fold AUCs is what a board prints and never receives the interval computed for the quantity above it.
Every column is rounded on its own, so the difference of two printed arm scores can differ from the printed effect in the last place.
OpenHands / SWE-rebench folds are \texttt{StratifiedGroupKFold(n\_splits=5, shuffle=True, random\_state=seed), grouped by instance\_id}.
ScienceWorld folds are \texttt{StratifiedKFold(n\_splits=5, shuffle=True, random\_state=seed)}.
Both partitions establish five seeds of five folds; every row held out exactly once per seed; both classes present in all 25 folds; no group split across a fold boundary; and all four arms scored on one identical partition on every feature layer.
Protocol difference: section D: the grouped OpenHands protocol differs from the row-split protocol the original POST boards use. The manuscript states that difference rather than presenting the numbers as like for like.
OpenHands / SWE-rebench intervals are a paired percentile bootstrap over resampled issue groups, carrying 4 score vector(s) in every draw.
ScienceWorld intervals are a paired percentile bootstrap over resampled rows, carrying 4 score vector(s) in every draw.
$S$ carries all four score vectors in every draw, because S is a difference of differences.
Conditionality: section F: these intervals are conditional on the saved fits. They do not include resample-and-refit training uncertainty, and folds and seeds are not treated as independent runs.
The primary cell is marked $^{\dagger}$: OpenHands / SWE-rebench $S$ = $-0.0066$, $[-0.0141, +0.0010]$, which is the quantity section E nominated before any of the six existed.
Branch rule: section G, with the overlap between its first two rows resolved toward the more specific row: both corpora attenuating is substantial-attenuation, one attenuating beside a stable-positive boundary is mixed-informative, both bounding is stable-positive-both, anything else is unresolved-or-modest.
The recorded branch is unresolved-or-modest. Neither corpus reaches substantial attenuation and neither gives an informative stable-positive boundary, so section G reads the pair as unresolved or modest.
No promotion: section G: no result at one corpus is promoted to stand for both.
Failing to reject is not stability: section G: an interval for S that includes zero while also including the declared magnitude is unresolved and is reported as unresolved.
The 0.03 magnitude is a declared editorial magnitude about what would be a worthwhile result for this paper, stated as such in section G; not a deployment threshold and not a validated minimum useful effect.
Measured overlap between the 574 distinct issues of OpenHands / SWE-rebench and the 376 distinct issues of the scored SWE-Gym population over 376 rows: one shared issue, \texttt{iterative\_\_dvc-1651}, read from the raw shards.
No row dropped: section C: the audit refits within each corpus, so a shared issue does not train across the old and new populations and no row is dropped for it. Any nonzero overlap is disclosed in the appendix.
No disjoint-issue-set claim: section I as its dated entry of 2026-09-11 amends it: the measured overlap is nonzero, so the issue collections may not be described as disjoint and the shared issue is disclosed in the appendix.
No board-evaluation claim for either corpus: section I: neither corpus is a board evaluation and CatchBench's POST board does not cover four corpora. That board is Table~\ref{tab:det}.
No unseen-task-family claim for ScienceWorld: section I: inference on ScienceWorld is conditional on the selected source file and its measured task-family mixture, which is uneven.
No independent-scaffold claim for OpenHands: section I: SWE-Gym's scored runs already use the OpenHands scaffold and GRADE's SWE-Gym adapter imports \texttt{agent\_graph\_openhands.to\_steps} directly. This is a change of producing model and issue collection.
Chronology: the declaration's chronology section: this audit was proposed after Part A's result and after Round 31, on data whose neighbouring results are known. It stays labelled exploratory whichever way it comes out.
Stopping rule: section J: one declared fit batch and one reporting pass. A disappointing effect is not an implementation defect and is not grounds for a rerun.
Rosters: \texttt{detection.\_LOADERS}, every board roster and \texttt{live\_streaming\_methods()} are untouched; the nine boards, the 72 entrants and the 138-record registry are what they were.
The batch recorded no failures and no warnings.
Generated by \texttt{tools/emit\_post\_generalization\_table.py}.

\begin{table}[t]
\centering
\scriptsize
\setlength{\tabcolsep}{4pt}
\caption{Contribution (2) rests on two corpora: SWE-Gym's structural increment changes sign under a flexible size control and tau-bench's does not. On two further corpora, how much of the linear dependency increment survives the same change of size control.}
\label{tab:post-generalization}
\begin{tabular}{@{}lrr@{}}
\toprule
Arm or quantity & OpenHands / SWE-rebench & ScienceWorld \\
\midrule
Runs (failed, solved) & 600 (312, 288) & 128 (64, 64) \\
Resampling unit & issue group & row \\
Units resampled & 574 & 128 \\
\midrule
size (flat) & 0.696 & 0.814 \\
\texttt{auditable} (size+deps) & 0.695 & 0.883 \\
size (spline) & 0.688 & 0.852 \\
size-spline + linear-deps & 0.694 & 0.906 \\
\midrule
$L$: \texttt{auditable} (size+deps) $-$ size (flat) & $-0.0011$ & $+0.0681$ \\
95\% interval for $L$ & $[-0.0099, +0.0079]$ & $[+0.0165, +0.1285]$ \\
$D$: size-spline + linear-deps $-$ size (spline) & $+0.0055$ & $+0.0547$ \\
95\% interval for $D$ & $[-0.0069, +0.0183]$ & $[+0.0005, +0.1152]$ \\
$S = L - D$ & $-0.0066^{\dagger}$ & $+0.0134$ \\
95\% interval for $S$ & $[-0.0141, +0.0010]$ & $[-0.0488, +0.0799]$ \\
\midrule
Reading (section G) & unresolved & unresolved \\
\bottomrule
\end{tabular}
\end{table}

\begin{figure}[t]
\centering
\includegraphics[width=\textwidth]{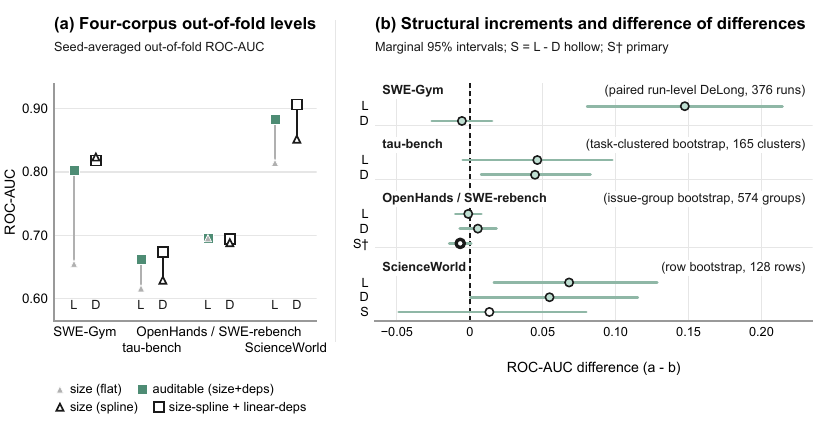}
\caption{(a)~Seed-averaged out-of-fold ROC-AUC for the linear pair ($L$: size (flat) to \texttt{auditable} (size+deps), filled) and the matched pair ($D$: size (spline) to size-spline + linear-deps, hollow). (b)~Increments $L$, $D$, and $S = L - D$ with marginal and uncorrected 95\% intervals against a dashed zero line. OpenHands / SWE-rebench supplies the primary cell, its $S$ row ($S^\dagger$). Its folds are issue-grouped, unlike the row splits on the detection boards; neither audit corpus is a board evaluation; one issue is shared with the scored SWE-Gym population; and the recorded reading (section G) is unresolved for both corpora. The SWE-Gym matched effect is $-0.005$ on seed-averaged out-of-fold predictions and $-0.010$ on the mean-fold estimand. $S$ is not computed for SWE-Gym or tau-bench. Exact cells: Table~\ref{tab:core-contrasts} and Table~\ref{tab:post-generalization}.}
\label{fig:structural_increments}
\end{figure}

Figure~\ref{fig:structural_increments} summarizes these linear and matched structural increments across the two board corpora and the two audit corpora.

\subsection{Full Unsupervised Detector Arena}
\paragraph{Off-the-Shelf Tabular Detectors Stay Far Below.}
The detection table reports the headline detectors. Behind it, \sysname runs a wider unsupervised
anomaly-detection arena, which tests whether any off-the-shelf detector recovers the failure signal
without task-aware features. Two families enter. PyOD tabular detectors run on the flat per-run
features (Isolation Forest, KNN, LOF, COPOD, HBOS) \citep{zhao2019pyod}, and PyGOD graph detectors run
on the typed graph (DOMINANT, CONAD, AnomalyDAE, GAAN) \citep{liu2022bond}. On SWE-Gym the tabular detectors span ROC-AUC 0.319 for HBOS to
0.625 for COPOD, and on tau-bench 0.504 for LOF to 0.593 for COPOD. The tabular detectors are
deterministic here. The four graph detectors train neural models from random initializations, so
we report them over twenty initialization seeds.

\paragraph{Graph Detectors Are Too Seed-Sensitive.}
Their instability is itself the first result. Across twenty detector-initialization seeds on SWE-Gym
the means are DOMINANT 0.631 $\pm$ 0.164, AnomalyDAE 0.488 $\pm$ 0.149, CONAD 0.596 $\pm$ 0.099,
and GAAN 0.774 $\pm$ 0.091. The supervised references vary on a different axis: 0.804 $\pm$ 0.004
for the structural features across cross-validation split seeds, and 0.824 $\pm$ 0.007 for the
supervised graph network across seeds that set both its split and its initialization. A detector whose
seed-to-seed standard deviation runs from twelve to twenty-two times the supervised graph network's
is not something an auditor can act on, whatever its mean. Quadrupling the seed count did not
settle the family down: it moved DOMINANT's mean by more than 0.10 and widened its spread, which is
what an unstable estimator does when it is sampled more.

\paragraph{GAAN Alone Approaches the Supervised Reference.}
GAAN is the one entrant whose mean approaches the supervised reference. Over twenty seeds its Welch
difference against the supervised graph network is $-0.050$
with a 95\% interval of $[-0.095, -0.006]$. That interval is over initialization seeds rather than
over runs, and it is a marginal reading of that one comparison.
Its run score correlates $-0.736$ with node count, in the
direction expected when the failed SWE-Gym runs are the shorter ones, which is consistent with a
length-related signal. Table~\ref{tab:transfer} in Appendix~\ref{app:board-values} carries every
value in this subsection, derived from the two committed seed records. Restricting comparisons to the 224 of 376 runs falling in the 42 node-count
strata that contain both outcomes gives a pair-weighted within-size ROC-AUC of 0.652 $\pm$ 0.169 over
262 positive-negative comparisons, with an interval across initializations of $[0.571, 0.734]$.
That control removes the node-count difference inside each comparison, and it also changes the
evaluated population. What it measures is a within-size score on the matchable subset under
initialization uncertainty. It is not a beyond-size score on the board, and not a reading about the
runs it drops. The interval is over initializations of a fixed 262-pair comparison set, so it does not carry
the sampling error of that set. On tau-bench the whole family sits at chance, from 0.513 $\pm$ 0.042
to 0.523 $\pm$ 0.031, and on Who\&When localization DOMINANT reaches
0.057 $\pm$ 0.019 Top-1, below the empirical random reference of 0.119. GUARDIAN, the agent-specific reconstruction autoencoder
\citep{zhou2025guardian}, scores 0.767 on SWE-Gym against ECOD's 0.765;
Table~\ref{tab:all-contrasts} carries that difference with its interval.

\begin{figure}[t]
\centering
\includegraphics[width=\textwidth]{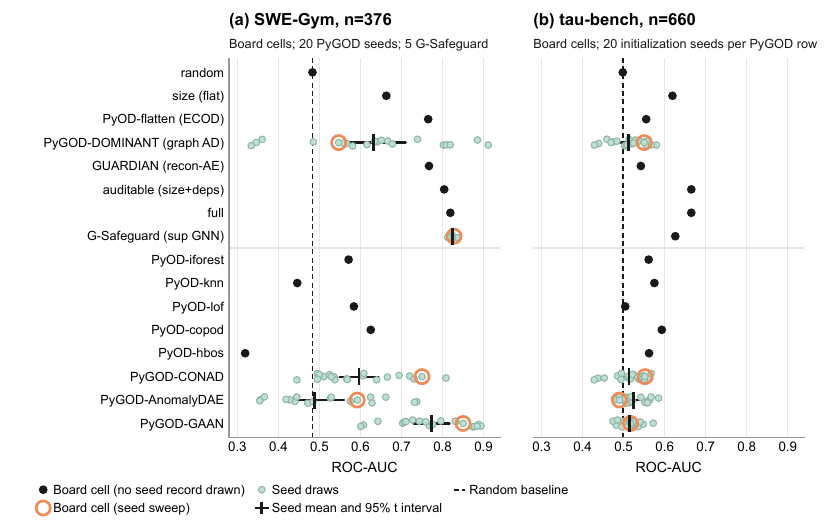}
\caption{Full unsupervised detector arena and seed stability on (a)~SWE-Gym and (b)~tau-bench. Each row shows one entrant of the sixteen-entrant POST detection board in committed board order, separated by a light horizontal line between the eight headline methods of Table~\ref{tab:det} and the eight arena methods. Entrants without a drawn seed record appear as single dark points at their board cells. Supervised rows are refit per split seed and their spreads are reported in the text beside Table~\ref{tab:det}; the PyOD rows are deterministic. For the four PyGOD graph detectors, strips show twenty initialization-seed ROC-AUC values as jittered mint dots, with the twenty-seed mean as a vertical tick and the 95\% t interval on the mean as a horizontal bar. The single-seed board cell is ringed in coral. On SWE-Gym, G-Safeguard shows five joint split-and-initialization seeds in the same manner beside its board cell. That board cell is the board's own five-split-seed value rather than one of the five joint-seed draws, so the ring there does not coincide with a dot. These strips show variation across random seeds on fixed runs, not sampling intervals over runs. The random baseline is a dashed rule. Exact cells: Table~\ref{tab:det} for the headline board cells, and Table~\ref{tab:transfer} for the means, standard deviations, intervals, the GAAN run-size control, and the Welch comparison.}
\label{fig:detector_arena_seeds}
\end{figure}

\paragraph{Seed Stability Qualifies the Single-Seed Board Maximum.}
What the corrected arena shows is narrower: no off-the-shelf detector establishes a lead on a
task-relevant board. GAAN's twenty-seed mean sits below the supervised reference, and its
exact-node-count control reads a within-size score on the matchable subset rather than a
beyond-size score on the board. PyGOD's dropped-grounding cells
display above the keyed span baselines, and their twenty-initialization-seed mean
stays below the matched random floor. The task-aware
structural features, and a network trained on them, are what carry the signal reliably. Five-seed
sweeps come from \texttt{tools/pygod\_seed\_stability.py}. Figure~\ref{fig:detector_arena_seeds} visualizes the full sixteen-entrant detection arena on both corpora alongside these seed sweeps.

\paragraph{Online Stale-State Detection.}
The LIVE stale-state board reuses the \sysname-Gold stale-state injection
(Section~\ref{sec:data-gold}), but scores it as online detection rather than post-hoc localization. Each
step receives a causal score, the z-score of its dependency span against the prefix strictly before
it. A run is flagged when its peak score crosses
a threshold calibrated on the paired clean runs to a target false-positive rate. Because the injection
adds one spike to an otherwise identical clean run, a true-positive rate above the false-positive rate
isolates the injected signal.\looseness=-1

\paragraph{Online Detection Recovers Little Stale-State Signal.}
Table~\ref{tab:live-stale} reports a representative seed. At realized false-positive rates of
6.1\% and 11.0\%, the causal span z-score displays true-positive rates of 0.061 and 0.110; raw span
displays 0.122 and 0.159. Across five injection seeds, the corresponding means are 0.054
$\pm$ 0.012 and 0.124 $\pm$ 0.012 for the z-score, and 0.098 $\pm$ 0.017 and 0.151
$\pm$ 0.012 for raw span. The same injection reaches 0.703 Top-1 under post-hoc within-run
localization on Gold (Appendix~\ref{sec:res-gold}), but that value ranks a step once a failure is known
while this board flags a run at a fixed false-alarm rate. The two neither estimate nor bound relative
task difficulty. A dependency-count control checks the unchanged-count construction: redirecting an
edge changes no step's edge count, so its true-positive rate equals the realized false-positive rate.
At the displayed 5\% target, the dependency-count control and z-score each display 0.061, while raw
span displays 0.122. No comparison on this board is recorded with an interval, so these cells are
point estimates only and
support no claim about the effect of per-run normalization. Target false-positive rates are not
exactly attainable on 82 paired runs, so the realized rate is reported beside each true-positive
rate.

\begin{table}[t]
\centering
\caption{LIVE online stale-state detection on 82 paired runs (a representative seed). True-positive
rate at each target false-positive rate, with the realized clean-flag rate in parentheses. The
dep-count control preserves edge count, so its true-positive rate equals its false-positive rate (no
signal); it is the negative control, not a competitor.
Point estimates only. Each of the eight cells carries a true-positive rate and a realized
false-positive rate, and both need paired peak scores and a threshold recalibration inside every
resample, neither of which the released record carries. The 5\% and 10\% targets are fixed settings
rather than measurements and take no interval.}
\label{tab:live-stale}
\begin{tabular}{lcc}
\toprule
Method & TPR @ 5\% FPR & TPR @ 10\% FPR \\
\midrule
random & 0.024 (6.1\%) & 0.024 (11.0\%) \\
dep-count (control) & 0.061 (6.1\%) & 0.061 (6.1\%) \\
\texttt{auditable} (span z-score) & 0.061 (6.1\%) & 0.110 (11.0\%) \\
raw span & 0.122 (6.1\%) & 0.159 (11.0\%) \\
\bottomrule
\end{tabular}
\end{table}


\subsection{Additional Discussion Details}

\paragraph{Localization and Detection Endpoints.} The headline fault-localization reading is
GPT-5.5 at Top-1 0.452 against 0.211 for \texttt{exec-rank (sup.)}, the strongest of the five no-LLM
entrants; Table~\ref{tab:all-contrasts} carries GPT-5.5's difference against each of those five with
its interval. Every member of the eight-model band outscores that entrant, and only GPT-5.5's
comparisons against them are recorded, so the rest of the band is a point-estimate ordering. Three judges outside the band score below \texttt{exec-rank
(sup.)}: \texttt{gemma-3-12b} at 0.206, \texttt{mistral-small} at 0.135, and \texttt{nova-micro} at
0.127. The full-context dependency prior scores 0.159, identical to the position prior because the
full-context assumption makes the dependency score monotone in position. Consequently, the two rank
every one of the 126 runs alike. A best judge reaching only 0.452 shows that the board retains
substantial headroom. Structural methods do not yet beat that band at POST localization, but they carry
complementary evidence. On failure detection, \texttt{auditable} (size+deps) scores 0.804 against
0.663 for \texttt{size (flat)} on SWE-Gym, and 0.665 against 0.619 on tau-bench;
Table~\ref{tab:all-contrasts} carries the corresponding out-of-fold contrasts with their
intervals. At the first SWE-Gym
LIVE prefix, unsupervised ECOD reaches 0.756 against that block's 0.742, and both stay below the
full supervised model at 0.813. The four comparisons recorded in that column appear with their
intervals in the same table.

\paragraph{Headroom Endpoints.} The causal span z-score catches 6.1\% of stale reads at a realized
6.1\% false-positive rate, which is what its signal-free control also reaches. The 0.703 the same
injection reaches under post-hoc localization scores a different decision and is not an endpoint of
this axis. No reported LIVE point estimate reaches 0.70 on tau-bench; the best reported
score is 0.665.\looseness=-1

\paragraph{Evidence Qualification.} The standalone Gold v2 diagnostic is outside the shipped board.
AppWorld is a candidate named-value substrate for the next release. Results are reported per source
and label process, so the four processes do not pool silently. The PRE labels require caution.\looseness=-1

\section{PRE Corpus, Licences, and Label Reliability}
\label{app:pre-details}

\begin{table*}[htbp]
\centering
\scriptsize
\begin{tabular}{p{1.3cm}r p{2.3cm}p{1.7cm}p{5.2cm}}
\toprule
Source & Configs & Capability balance & Config balance & Selection and label rule \\
\midrule
crewai & 298 & 144 excess / 454 needed (24.1\%) & 90 / 208 &
Agent YAML entries from 105 commit-pinned public repositories; require a nonempty role and tool
roster; deduplicate by lowercased role plus sorted roster. Labels require both judges to agree. \\
\addlinespace
injecagent & 340 & 510 / 340 (60.0\%) & 340 / 0 &
Four test-case files; skip user and attacker overlap; deduplicate rosters; at most 20 per user tool,
round-robin across attack types. The user tool is needed and attacker tools are excess. \\
\addlinespace
mcp & 144 & 1457 / 1545 (48.5\%) & 94 / 50 &
Registry source list; require resolvable GitHub metadata, a live unauthenticated
\texttt{tools/list}, and a nonempty roster. Labels require both judges to agree. \\
\addlinespace
n8n & 219 & 46 / 504 (8.4\%) & 36 / 183 &
Frozen ordered template-ID list; keep published workflows with an AI Agent node and a nonempty wired
roster. Labels require both judges to agree. \\
\addlinespace
sweagent & 130 & 625 / 929 (40.2\%) & 130 / 0 &
An explicit 130-instance list; require a nonempty history beginning with a system message. Declared
minus observed use defines excess. \\
\addlinespace
synthetic & 56 & 111 / 69 (61.7\%) & 56 / 0 &
Fourteen authored cases for each of four frameworks; add declared excess capabilities to the authored
minimum and shuffle roster order under a fixed seed. \\
\midrule
Total & 1187 & 2893 / 3841 (43.0\%) & 746 / 441 & Six sources, four label processes. \\
\bottomrule
\end{tabular}
\caption{PRE corpus composition, computed from the six committed \texttt{data/pre/*.json} files. The
four label processes differ in strength, which is why per-source results are reported alongside the
pooled board rather than instead of it.}
\label{tab:pre-corpus-stats}
\end{table*}

\begin{table}[t]
\centering
\scriptsize
\begin{tabular}{lrp{5.6cm}}
\toprule
Source & Records & Licence values on released records \\
\midrule
crewai & 298 & Apache-2.0 9; GPL-3.0 9; MIT 132; \texttt{NOASSERTION} 148 \\
injecagent & 340 & MIT 340 \\
mcp & 144 & AGPL-3.0-only 1; Apache-2.0 15; BUSL-1.1 1; CC-BY-4.0 2; MIT 98; \texttt{NOASSERTION} 27 \\
n8n & 219 & \texttt{NOASSERTION} 219 \\
sweagent & 130 & \texttt{NOASSERTION} 130 \\
synthetic & 56 & MIT 56 \\
\bottomrule
\end{tabular}
\caption{Licence values read from the \texttt{license} field recorded with each released record. Every
record also carries \texttt{repo}, \texttt{commit}, and \texttt{path}. Of the 1187 records, 663 carry
an established declaration and 524 carry \texttt{NOASSERTION}: every n8n and SWE-agent record, and
about half of crewai. \texttt{NOASSERTION} records the absence of a grant rather than a permissive
one, so they should not be read as a claim that the upstream prose may be redistributed. The
repository distributes derived features and labels rather than that prose. An earlier reading of the
same records recorded 106 crewai and MCP entries as undeclared that their upstream states in a README
or package manifest; the counts here are the corrected ones.}
\label{tab:pre-licences}
\end{table}

\begin{table}[t]
\centering
\scriptsize
\begin{tabular}{lrrrr}
\toprule
Source & Configs & Capabilities & Agreement & Cohen's $\kappa$ \\
\midrule
crewai & 298 & 598 & 0.753 & 0.492 \\
n8n & 219 & 550 & 0.880 & 0.515 \\
mcp & 144 & 3002 & 0.838 & 0.674 \\
\midrule
Pooled & 661 & 4150 & 0.832 & 0.666 \\
\bottomrule
\end{tabular}
\caption{Capability-level agreement between the two judge vendors, the exact output of
\texttt{tools/pre\_merge\_judges.py}. A capability is labeled excess only when both judges agree it is
not needed, so disagreement costs recall rather than precision. Agreement is moderate on crewai
($\kappa = 0.492$) and n8n ($\kappa = 0.515$) and substantial on mcp. The two moderate values are a
real limit on those two sources and are the reason the labels are conjunctive.}
\label{tab:pre-kappa}
\end{table}

\begin{figure}[t]
\centering
\includegraphics[width=\textwidth]{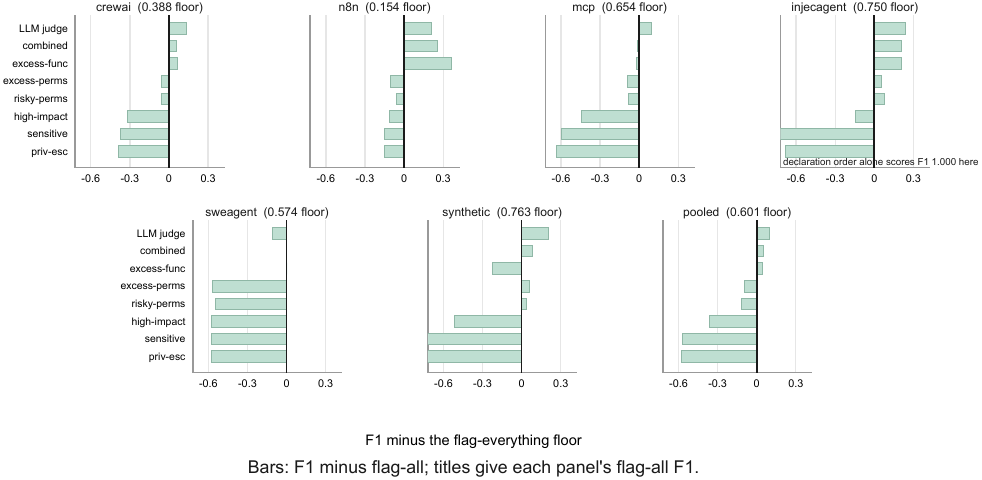}
\caption{The PRE board read against its own floor: F1 minus that source's \texttt{flag\_all} F1, so a
bar left of the line scores below flagging every declared capability. \texttt{flag\_none} (0.000) and
the oracle identity check (1.000) are omitted, and \texttt{flag\_all} is the line itself. The
comparison recorded for each source uses an estimand that need not equal the longest bar drawn
here.}
\label{fig:pre-source}
\end{figure}

Before a run, the available evidence is the agent's stated task or role and the capabilities granted
by its harness. PRE asks whether a method can flag granted capabilities that the task does not need.
The threat model is OWASP LLM06:2025 Excessive Agency, read statically from the declared artifact.
Tool poisoning is a separate threat. Over-privileged tool selection during execution belongs to LIVE,
where a method can inspect the growing trace.

The corpus contains 1187 derived feature records representing configurations from six sources: crewai
298, injecagent 340 \citep{zhan2024injecagent}, mcp 144, n8n 219, sweagent 130
\citep{yang2024sweagent}, and synthetic 56. FORTIS benchmarks the same threat from the model side,
scoring whether ten frontier models pick the minimally sufficient skill and stay inside the tools it
grants \citep{li2026fortis}. \sysname instead scores a detector over labeled pre-run configuration
records, so the subject under test here is the auditing method and the audited object is the declared
artifact. AuthBench and ToolPrivBench label the same threat on their own
substrates, file-level permissions for 120 terminal tasks and privilege-ranked tool
choices \citep{yan2026authbench,yang2026toolprivbench}. Each supplies task-relative
least-privilege labels, and each holds one artifact type from one construction. What \sysname adds
is a detector scored over heterogeneous declared harness records drawn from six sources. Their four
label processes stay explicit through a tag on every instance, so the board does not pool them
silently.
They are cross-vendor LLM judging for crewai, n8n, and mcp; roster relabeling for injecagent;
declared-minus-used labels from paired sweagent traces; and synthetic over-grant injection following
the controlled-anomaly precedent of BOND \citep{liu2022bond}. On the cross-vendor subset, inter-rater reliability
was Cohen's $\kappa=0.666$, per source in Table~\ref{tab:pre-kappa}. Offline harvesters produced the committed PRE feature records.
The board reads these derived records locally. The current record files omit upstream task and role
prose; earlier public revisions containing it remain accessible through repository history, as
described in the reproducibility statement.
Each instance records its origin and license; Table~\ref{tab:pre-licences} gives the
licence tallies per source and Table~\ref{tab:pre-corpus-stats} the corpus composition.

\subsection{PRE Scanner and Source Diagnostics}
\paragraph{Static Rules Cover Distinct Risks.}
Table~\ref{tab:pre-main} makes static coverage explicit. The permissions rule matches task verbs to
grant levels, while functionality compares the non-generic subjects of the task and capability.
The three narrower rules flag unrequested tokens from their respective risk classes.
\texttt{owasp\_asi\_combined} takes the union of all five rules.

\paragraph{Standards Fix the Scanner Scope.}
Every standard reference in Table~\ref{tab:pre-main} names its edition, so a later revision cannot
silently change what a rule is claimed to cover. The rules cite the OWASP Top 10 for LLM Applications
2025, which is the published edition, and MITRE CWE. One citation already deserves a note: CWE 4.20
marks CWE-269 as discouraged for direct vulnerability mapping and points instead to CWE-250, which
now names over-privileged agent components explicitly. The citation here documents conceptual
lineage rather than mapping a vulnerability, so it stands, and a later rule revision should prefer
CWE-250. Neither point changes what a rule computes, since each rule is a fixed token test over the
declared artifact.

\paragraph{Static Coverage Has Known Limits.}
Some standard concerns exceed a static single-configuration audit. A full excessive-autonomy check
requires an approval-gate field, which the schema does not carry. Full tool-misuse coverage requires
declared operation, scope, and allowlist controls that the schema does not express. Detecting a
deprecated or duplicate extension requires deployment history. Accordingly,
\texttt{unrequested\_high\_impact} approximates autonomy, and \texttt{sensitive\_access} measures an
LLM02 exposure surface rather than complete tool misuse.

\paragraph{Source-Specific Floors Reveal Method Value.}
Figure~\ref{fig:pre-source} reads the per-source board against each source's own
flag-everything floor, which is more informative than the pooled column because the four label
processes differ.
The label sources are LLM judge for crewai, n8n, and mcp; roster
relabel for injecagent; declared minus used for sweagent; and synthetic injection for synthetic.

\paragraph{Why the Judge Baseline Must Be Held Out.}
Section~\ref{sec:res-pre} reports the held-out LLM judge ahead of the combined scanner on the
configurations both judged. ``Held out'' is essential: two other judges produced the crewai, n8n,
and mcp labels, while Llama-3.3-70B produced none of them, so the baseline does not grade its own
outputs. The cost of skipping that design is
measurable. On the 656 judge-labeled configurations every method judged, the two label makers read
0.824 and 0.955 F1 against the held-out model's 0.703. Scored on all 661 they voted on, the label
makers read 0.854 and 0.960, but the held-out cache is five configurations short, because a strict
parser discarded five replies whose capability spelling did not match the declared roster. Only the
common population supports a head-to-head reading, and
\texttt{tools/pre\_label\_maker\_diagnostic.py} prints both. Part of that gap is mechanical rather
than skill. The merge marks a capability excess only when both judges call it unneeded, so the
released key is a subset of what either judge flagged. Each label maker therefore scores recall 1.000
by construction, and its F1 is decided by precision alone. That is exactly why a label maker cannot serve
as a baseline, and why the held-out row is the one the board reports.

\paragraph{Source Results Expose Label-Process Limits.}
The held-out judge is near the top on the roster-relabeled injecagent
and injected synthetic labels, at 0.990 and 0.972, but reaches only 0.362 to 0.744 on the
judge-labeled sources. The three narrow rules make few predictions, so low pooled recall indicates
that they cover small slices of excess rather than failing within those slices. Keyword rules are
language-brittle. A task written outside their keyword coverage falls to the read-only permission
floor and is over-flagged. When excess is rare, one over-flag can sharply reduce precision.

\paragraph{A Second Defect in Our Own Harness.} An earlier version of this board scored the held-out
judge at 0.659 F1 and read that as the judge scoring no better than the rule scanners. It was our
parser. The judge answered all 1187 configurations, but its reply is matched against the declared
capability roster by exact name. Five replies named a capability whose spelling did not match.
Three declared names carried a double or trailing space, one reply corrected a
misspelling in the source data, and one shortened name was ambiguous between two declared
capabilities. In each case the parser discarded the whole judgment, and the board recorded the method
as flagging nothing. Four of those five are small. The fifth is an MCP server declaring 622 capabilities and
carrying 337 excess labels, so that single discard removed 11.7\% of the corpus positive class and
cost the judge 337 false negatives it never made. We now treat an unparsed reply as an abstention:
the configuration leaves that method's denominator, and every row reports the coverage this costs, so
a method scored on fewer configurations can never be compared silently against one scored on all of
them. That correction moves the judge from 0.741 to 0.839 recall and from 0.659 to 0.695 F1, and moves
its mcp cell from 0.662 to 0.744. No other board in this paper changes.

\paragraph{Two of Six Sources Give the Floor No Stable Margin.} Those pooled gains hide
\texttt{sweagent} and \texttt{mcp}, the two panels of Figure~\ref{fig:pre-source} where the best
method's margin over its own floor is either near zero or decided by one file. On \texttt{sweagent}
the judge abstains on none of its 130
configurations, so the reading
is clean: \texttt{flag\_all} reaches 0.574, the excess-functionality rule ties it at 0.574 with a
paired difference of $-0.0001$, the combined scanner reads 0.570, and the judge falls to 0.467. On
\texttt{mcp} one file moves the margin by an order of magnitude.
Across all 144 configurations the best method sits 0.008 above the floor; across the 143 the judge
parsed, it sits 0.104 above, 0.744 against 0.640. Both readings turn on a single MCP server declaring
622 capabilities, 20.7\% of the source and 9.2\% of the whole corpus by scoring unit. The recorded
comparison for \texttt{mcp} is selection-aware over 143 configurations with the best of eight
candidates selected, and Table~\ref{tab:all-contrasts} carries its difference and interval.

\paragraph{Two of the Four Widest Source Margins Measure Construction Rather Than Reasoning.} The
four sources where the best candidate clears \texttt{flag\_all} by the widest margin are the
roster-relabeled \texttt{injecagent} source at 0.990 against 0.750, the injected
\texttt{synthetic} source at 0.972 against 0.763, and the two judge-labeled sources where excess is
rarest, \texttt{n8n} at 0.528 against 0.154 and \texttt{crewai} at 0.518 against 0.388. The first two
we report as construction. The \texttt{injecagent} harvester writes each roster as the user tool
followed by the attacker tools, and the released records preserve that order, so on all 340
configurations the first declared capability is exactly the minimum and the rest are exactly the
excess. A rule that reads nothing but position, keeping the first capability and flagging the tail,
scores 1.000 F1 there, above every method in Figure~\ref{fig:pre-source}. It makes 510 true
positives, no false positives, and no false negatives. That the leak is the corpus and not the rule
is visible in the same rule's other columns: 0.446 on \texttt{crewai} and 0.106 on \texttt{n8n}. The
rule is recorded as \texttt{reported\_quantities.pre.declaration\_order\_leak} in
\texttt{tools/statistical\_tests\_results.json}, scored on every source, and \texttt{injecagent} is
the only one where it is perfect. We report it as an exact
construction diagnostic on the released corpus rather than as a population-level method comparison,
so it carries no interval.
The \texttt{synthetic} source is authored, and its injected capabilities are separable from surface
features the scanners already read. Only \texttt{crewai} and \texttt{n8n} show a margin that no
construction accounts for, and those are the two sources whose labels are weakest, at
$\kappa = 0.492$ and $0.515$ (Table~\ref{tab:pre-kappa}).

\paragraph{The Floor Tracks Each Source's Base Rate, and the Pooled Column Tracks One File.} Since
\texttt{flag\_all} has perfect recall, its F1 is a monotone reading of each source's base rate
(Table~\ref{tab:pre-corpus-stats}), which runs from 8\% on \texttt{n8n} to 62\% on
\texttt{synthetic}, and the two sources it wins are the mid-range ones. A method earns its false
alarms only where it clears that floor by a margin an operator would notice; on \texttt{sweagent}
that margin is $-0.0001$, and on \texttt{mcp} it is either 0.008 or 0.104 depending on how one file
is handled. The pooled column carries a second distortion. Its
score is micro-averaged over capabilities while the sampling unit is the configuration, and
configuration size runs from a median of 3 capabilities to a maximum of 622. So \texttt{mcp}
contributes 44.6\% of the pooled denominator from 12.1\% of the configurations, and one file
contributes 9.2\% on its own. The per-source columns are where this board carries its result.

\clearpage
\section{Corpus Composition}
\label{app:corpus-composition}

\begin{table*}[htbp]
\centering
\scriptsize
\begin{tabular}{p{1.9cm}p{1.5cm}p{2.2cm}p{1.7cm}p{4.4cm}}
\toprule
Population & Board size & Label balance & Step filter & Selection rule \\
\midrule
Who\&When Algorithm-Generated \citep{zhang2025whoandwhen} &
126 runs, 1099 steps &
126 fault steps, 973 other &
$\geq 3$ converted steps; failed; valid \texttt{mistake\_step} &
The loader requires \texttt{is\_correct=false}, which selects all 126 Algorithm-Generated files and
excludes all 58 Hand-Crafted files, whose key is \texttt{is\_corrected}. \\
\addlinespace
SWE-Gym POST and LIVE \citep{pan2024swegym} &
376 runs &
188 failed, 188 resolved &
POST $\geq 2$, LIVE $\geq 4$; no current row is lost by either &
From 6055 rows in the dump, take the single \texttt{run\_id} with the largest balanced capacity, then
retain 188 runs per class. \\
\addlinespace
tau-bench POST and LIVE \citep{yao2024taubench} &
660 runs &
363 failed, 297 resolved &
$\geq 3$ messages with \texttt{db\_match}; POST $\geq 2$, LIVE $\geq 4$ &
All rows from four fixed model files at the pinned revision; all 660 pass both filters. The
population is 200 airline and 460 retail runs. \\
\addlinespace
File-level Gold &
188 injected runs with 188 paired clean controls &
82 stale-state, 106 dropped-grounding &
Resolved SWE-Gym runs with $\geq 4$ steps and at least one dependency &
Alternate the intended fault kind by run order, falling back to the other kind when the intended one
is ineligible. \\
\bottomrule
\end{tabular}
\caption{POST, LIVE, and file-level Gold populations. Who\&When counts come from
\texttt{tools/whoandwhen\_split\_report.py}, which exits non-zero if a corpus refresh changes the
scored population. Gold counts come from \texttt{src/catchbench/gold.py} and were reproduced by
\texttt{tools/gold\_artifact\_diagnostic.py}.}
\label{tab:post-live-corpus-stats}
\end{table*}

\begin{figure}[t]
\centering
\includegraphics[width=\textwidth]{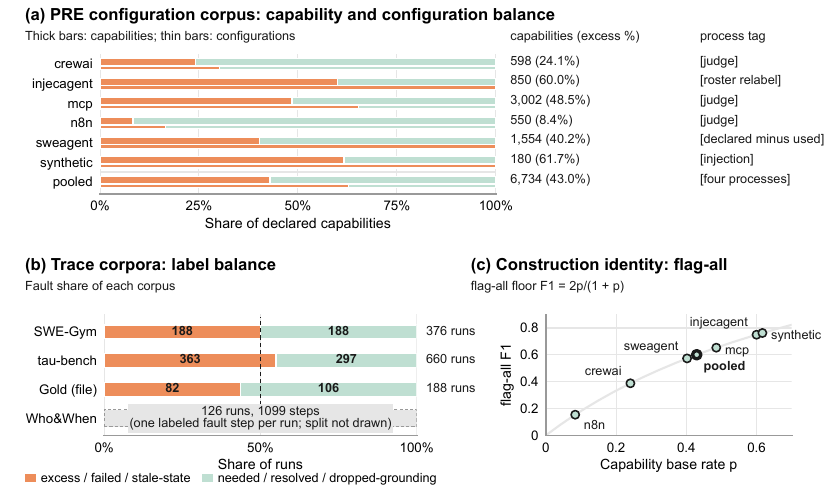}
\caption{Dataset composition across the benchmark.
(a)~PRE configuration corpus: share of declared capabilities split into excess (coral) and needed (mint) with total counts and excess rates at the bar end; thinner bars give configuration-level balance with and without excess; right tags mark the label process.
(b)~Trace corpora: stacked bars display label balance for SWE-Gym (failed vs resolved), tau-bench (failed vs resolved), and file-level Gold (stale-state vs dropped-grounding), with a dashed rule at an even split. Who\&When reports 126 failed runs and 1099 steps; its 126-of-1099 fault-step split is printed in Table~\ref{tab:post-live-corpus-stats} and not drawn.
(c)~Recorded construction identity for the flag-all floor, $F_1 = 2p/(1+p)$, against capability base rate $p$, with the seven recorded source cells.
Exact cells: Table~\ref{tab:pre-corpus-stats} (Appendix~\ref{app:pre-details}) for the PRE counts and label rules, and Table~\ref{tab:post-live-corpus-stats} for the trace counts and selection rules.}
\label{fig:corpus_composition}
\end{figure}

The SWE-Gym dump holds 491 resolved and 5564 unresolved rows, a pass rate of 8.1\%, so the balanced
188-and-188 board is far denser in successes than the source population and should not be read as a
sample of it. Within the selected \texttt{run\_id}, 188 resolved and 2241 unresolved runs are
available; balancing keeps all 188 resolved runs and the first 188 unresolved ones. The 58 excluded
Hand-Crafted Who\&When runs hold 2993 steps with median length 32 against 10 for the included split,
which is the length difference cited in Section~\ref{sec:data-corpora}.


A benchmark is only as strong as the data its methods run on, so \sysname makes the dataset the
asset rather than any one method. A new task or corpus uses the shared interface, but it is scored only
by methods that explicitly declare support for that task ID. Figure~\ref{fig:corpus_composition} summarizes the composition of the PRE configurations, trace corpora, and the flag-all construction identity. Table~\ref{tab:post-live-corpus-stats} gives each POST and LIVE population's selection
rule and label balance, and Table~\ref{tab:pre-corpus-stats} does the same for PRE.
The release grows the data along two axes,
more information states and more corpora, while preserving the result interface.\looseness=-1

The Who\&When localization board contains 126 naturally occurring CaptainAgent failures with human decisive-step labels.
They are the Algorithm-Generated split of Who\&When, which also ships a 58-run Hand-Crafted
split that this board excludes as a scope choice. Hand-Crafted trajectories omit the per-step agent
\texttt{name} field, in 0 of their 2993 steps against 1099 of 1099 in Algorithm-Generated. The
converter falls back to \texttt{role}, so the structural feature vector stays defined: over those
2993 steps seven of the eight execution features vary and only \texttt{is\_tool} is constant. What
separates the splits is schema and trace length. Their median run is 32 steps against 10, which by
itself moves the analytic random Top-1 floor, the mean of $1/n$, from 0.120 on this split to 0.095
pooled over both. We report Algorithm-Generated alone to keep those two
populations out of every comparison. The role fallback already makes a separate Hand-Crafted board
possible; pooling the two would first need a validated cross-split comparison design. The repository ships \texttt{tools/whoandwhen\_split\_report.py}, which prints this composition
and exits non-zero if a corpus refresh changes the scored population.
Who\&When Pro \citep{liu2026whowhenpro} is a broader controlled-injection corpus whose available cards
do not document dependency edges or event timestamps, so structural methods require a validated
conversion layer.\footnote{The project site links to the mutable, unpinned artifact at
\url{https://huggingface.co/datasets/Leoxx/whowhen_pro}. When accessed on 2026-08-15, it listed
10{,}784 rows and a 17-mode taxonomy. A separate card at
\url{https://huggingface.co/datasets/tmpxv7/who-when-pro} advertised more than 13{,}500 traces and 18
modes on the same date, but the project site did not link to it.}

\section{Defensibility Bar Verdicts}
\label{app:defensibility-verdicts}

\begin{table*}[htbp]
\centering
\scriptsize
\begin{tabular}{p{1.9cm}p{4.9cm}p{5.1cm}}
\toprule
Bar item & File-level Gold v1 & Named-value Gold v2 \\
\midrule
1. Fault realism &
\textbf{Pass as a taxonomy mapping.} The injector implements stale-state redirection and dropped
grounding rather than an abstract graph perturbation. The stale edge remains an inferred
dependency-misattribution proxy, so empirical realism is deferred to item 5. &
\textbf{Undetermined.} Dropped grounding changes one real argument leaf using a same-key donor from
the corpus. Stale-state has 16 sites in 6 runs and supplies no natural validation example, and
development was adaptive and in-sample. \\
\addlinespace
2. No artifact leakage &
\textbf{Fail.} The diagnostic uniquely ranks 82 of 82 stale and 106 of 106 dropped targets and flags 0
of 188 paired clean runs, at every one of five seeds (Section~\ref{sec:data-gold}). &
\textbf{Undetermined.} All six process controls meet the current margin, but positive-control power is
demonstrated for only three of them and only for Top-1. \\
\addlinespace
3. Distributional validity &
\textbf{Undetermined.} Paired summaries are reported, but no acceptance threshold or global separation
test is. &
\textbf{Undetermined.} The fixed-margin readings place each control near the matched floor, but three
controls lack demonstrated power and the injector was tuned on these same diagnostics. \\
\addlinespace
4. Non-circular labels &
\textbf{Pass.} The target is the injection site fixed at construction, independent of every detector. &
\textbf{Pass.} Each paired graph stores the mutated event and argument path as the label; no control or
oracle chooses that label. \\
\addlinespace
5. Validation &
\textbf{Not met.} No human-audited slice is reported, and the substrate uses inferred file-level
dependencies rather than tracked value flow. &
\textbf{Not met.} No human-audited slice is reported, and stale-state lacks enough sites for a realism
study. \\
\bottomrule
\end{tabular}
\caption{Per-item verdicts against the five-item bar of Section~\ref{sec:data-gold}. A substrate clears the
bar only when every row passes. Neither does: v1 fails item 2 and lacks item 5, and v2 is undetermined
on items 1 to 3 and lacks item 5. This is why both Gold boards are reported as mechanism diagnostics
and neither is promoted to evidential weight.}
\label{tab:defensibility-verdicts}
\end{table*}

The Who\&When localization, failure-detection, and LIVE streaming boards reuse source-corpus labels.
Gold localization, cause attribution, and LIVE online stale-state detection use constructed injection
labels. Gold is the benchmark's run-time injection methodology, together with the admissibility checks
that govern its use.
It plants a known fault in a real run and asks whether a method points to it. Labeled real
dependency-state failures do not exist at scale, so we synthesize them, following ADBench and BOND
\citep{han2022adbench, liu2022bond}. As in BOND, the construction and its checks are reported
plainly. A check that fires is evidence that the protocol is working.

\paragraph{The Defensibility Bar.}
An injection is defensible only if it clears five checks, each tested rather than assumed.
Table~\ref{tab:defensibility-verdicts} records the per-item verdict for both Gold substrates.

\begin{enumerate}
  \item \textbf{Fault realism.} Each injected fault maps to a documented agent failure mode, not an
    abstract graph perturbation.
  \item \textbf{No artifact leakage.} No baseline keyed only to the substrate construction or the
    injection mechanism wins trivially. If such a detect-the-artifact baseline separates injected
    from clean, the injection leaks.
  \item \textbf{Distributional validity.} Injected runs resemble clean runs at the run level, so a
    global giveaway cannot stand in for the fault.
  \item \textbf{Non-circular labels.} The label is the injection site, correct by construction and
    independent of any detector.
  \item \textbf{Validation.} A human-audited slice confirms annotators judge the injected step a real
    fault, and an airtight substrate removes inference noise.
\end{enumerate}

\paragraph{The Fault Taxonomy.}
We realize two grounded faults on the inferred dependency layer of a real run, alternated across runs.

\begin{itemize}
  \item \textbf{Stale-state read.} Redirect one dependency from the latest event on a file to an
    earlier, superseded event on the \emph{same} file, so the step relies on an out-of-date version of
    that resource. Signature: an unusually long dependency span.
  \item \textbf{Dropped grounding.} Remove one required dependency, so the step acts ungrounded.
    Signature: an unusually low dependency count.
\end{itemize}

\paragraph{The Substrate.}
The substrate is 188 resolved SWE-Gym runs with inferred dependency edges, one fault per run at a
known step, 82 stale-state and 106 dropped-grounding (a run affords a stale-state fault only when it has an earlier same-file read to redirect to).

\paragraph{The Construction Artifact This Substrate Does Not Control.}
The clean file-level substrate constructs every dependency edge by the rule
\texttt{deps = [last\_on\_file[f]]}. Thus every file event with an earlier same-file event points to
its immediate same-file predecessor, without exception. Both injections mutate that stored edge:
stale-state redirects it to an older event, while dropped-grounding removes it. Both therefore break
an invariant that clean construction never breaks.

A broken-predecessor baseline uses only this construction rule. In each of five injection seeds, it
uniquely ranks all 82 stale-state targets and all 106 dropped-grounding targets Top-1, while flagging
0 of the 188 paired clean runs. Both fault kinds therefore fail Defensibility Bar item 2 on this
substrate. The eligibility-matched control is a target-selection control, and its span lift measures
mechanism signal; it does not control the missing-predecessor marker. Item 4 passes because the label
is the injection site, fixed during construction and independent of
every detector.\looseness=-1

The public repository ships this check as
\texttt{tools/gold\_artifact\_diagnostic.py}; it exits 0 exactly when the documented separation
reproduces. The Gold boards above are therefore mechanism diagnostics rather than
artifact-controlled benchmark evidence. Detector quality cannot repair this construction invariant;
the substrate must change.

\paragraph{Distributional Check and Caveats.}
Stale-state preserves the run-level dependency-edge count (mean 9.2, unchanged); dropped-grounding
removes exactly one edge (mean 7.9 to 6.9); stale-state lengthens the run-level maximum dependency span
by construction. The board's span line is not split by fault kind: across all 188 paired runs the
mean rises from 8.6 to 9.4 and 53 increase, and only 82 of those runs carry a stale-state
injection. These are distributional properties of
the artifact-limited Gold results. The dependency-aware detector is essentially the raw
\texttt{max-span} control, so it is keyed to the stale-state mechanism and is reported beside that
control rather than as a general detector. SWE-Gym dependencies are inferred, not gold value-flow, so
a redirected edge is a
dependency-misattribution proxy for a true stale read. Section~\ref{sec:data-goldv2} reports the
named-value rebuild with explicit writes and reads, where the admissibility checks must be applied
again; a human-audited validation slice remains part of the bar.

Gold v2 replaces the artifact-limited file-level Gold substrate with a named-value design. It mutates
an argument value and derives dependency edges from the values present when the graph is built. The
rebuilt graph can therefore be the output of a real run without storing a separately edited edge.

The substrate uses tau-bench trajectories \citep{yao2024taubench} pinned at revision
\texttt{382e57d}: 660 runs, of which 363 failed and 297 resolved. The extractor reads scalar leaves
from raw messages because the existing loader discards values. At graph-build time, each consumed
value links to the latest earlier result containing that value. Every consumption receives one
grounding class: \emph{derived} from a prior result, \emph{given} by the user or system, or
\emph{ungrounded}. The shipped board does not report named-value consumption counts or edge-share
diagnostics; those quantities await board integration.\looseness=-1

\paragraph{Stale-State Read.}
An entity is indexed by tool and primary identifier. Let two successive observations of one entity
hold value sets $V_1$ and $V_2$ at the same field $p$. A site is eligible when
$V_1\setminus V_2$ is nonempty and a later call consumes a scalar equal to a value in $V_2$. The
current implementation does not require the consumption argument path to match field $p$. The
injector replaces the exact argument leaf with a sampled value from $V_1\setminus V_2$; the argument
key and count remain unchanged.

\paragraph{Dropped Grounding.}
Any identifier-shaped argument leaf derived from a prior result is eligible. The injector samples
real corpus donors from the same collapsed argument key in other runs, rejecting any donor consumed,
produced, or supplied in the target run. It then selects a donor whose string-distance profile most
closely matches the original value, with seeded random tie breaking. Only the selected leaf changes;
the key, argument count, and list siblings remain unchanged.

\paragraph{Standalone Admissibility Diagnostic.}
Six process-artifact controls test editing traces: \texttt{format-outlier}, \texttt{schema-shape},
\texttt{position-prior}, \texttt{field-prior}, \texttt{tool-prior}, and \texttt{edit-distance}. Two
fault-definitional oracles are reported separately: \texttt{superseded-value} for stale reads and a
grounding oracle for ungrounded consumptions. The former controls should remain at a matched random
floor; the latter oracles test whether a scorer recognizes the programmed fault predicate.

The intended matched floor assumes a uniformly random ranking over the injector's eligible pool. An
earlier evaluator credited only the first tied candidate in eligible-pool order, which rewards a
control that scores many candidates equally. Both Gold substrates now compute the expected score
under uniform tie breaking. The two evaluators keep separate implementations, because v1 returns a
Top-1, Top-3, and MRR triple while v2 returns Top-1 alone, so a regression test on each side pins a
constant-score baseline to the analytic random floor and holds the two rules together.
The same change moved the Gold v1 full-pool table in Appendix~\ref{sec:res-gold}: it raised the controls
there and lowered the headline stale-state figure from 0.707 to 0.703. The rule was applied to both
substrates at once, and it moved numbers in both directions. Gold v2 now runs through
\texttt{run.py} under \texttt{-{}-task gold-v2}, and \texttt{tools/namedvalue\_admissibility.py}
keeps the standalone fixed-margin diagnostic alongside the artifact check of
Section~\ref{sec:data-gold}.

\paragraph{What the Shipped Panel Reports.}
The diagnostic now prints as the board block \texttt{gold\_v2\_namedvalue} on the tau-bench
named-value substrate, over 614 injected pairs at each of five seeds on dropped grounding. All six
process controls sit inside the fixed margin: Top-1 gaps against the matched floor run from
$-0.019$ to $+0.034$, and run AUC from 0.500 to 0.512, so the panel is PASS on both axes. The
grounding oracle recovers the injected site at 1.000 Top-1, so the target is reachable and a failing
control would be a property of that control rather than of an unfindable site. Its stale-state
counterpart sits at the floor, as a predicate for the other fault kind should. Even so, the
no-artifact-leakage bar stays \textsc{undetermined}, because a control that passes without
demonstrated power carries no information. Stale-state is not scored here: the full corpus affords 16
eligible sites in 6 runs.\looseness=-1

Every value in that block is a displayed diagnostic cell. Its rows are a matched floor, six
construction controls, and two oracles rather than competing entrants, and no comparison over it is
recorded, so it enters no count in this paper. The nine boards, the 72 entrants, and the 138
recorded comparisons all exclude it. It joins the arena when it carries recorded comparisons and
entrants.

\paragraph{Development Record.}
Development was adaptive and in-sample: string-distance donor matching followed earlier standalone
runs, so a future PASS on a new substrate requires a frozen injector.\looseness=-1

The positive-control inventory is incomplete: its standalone script covers three of six controls on
Top-1 alone, and no run-AUC path carries a power test.

The first evaluator built its eligible pool from the injected graph, which removed the true site from
its original eligibility class and gave every scorer Top-1 zero. The harness then reported
\textsc{not admissible} with all six controls failing. The grounding oracle exposed the bug, showing
why the control panel must retain an oracle with a known correct value.

\subsection{Gold Board Mechanism Diagnostics}
\label{sec:res-gold}
\begin{figure}[t]
\centering
\includegraphics[width=\textwidth]{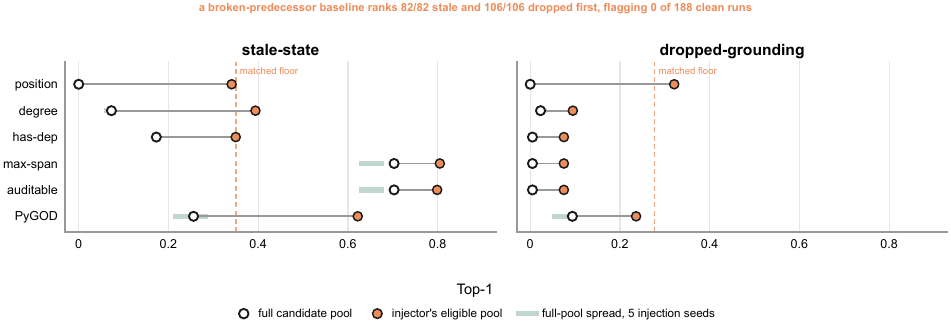}
\caption{What the eligibility control removes. Each row is one method; the open marker ranks
it over the whole candidate pool and the filled marker only inside the pool the injector could
have targeted, both at the same representative injection seed. The band is that method's
full-pool spread over five injection seeds. The full-pool random floor is 0.029 on stale-state
and 0.035 on dropped-grounding, which at this scale is the axis; the matched floor is drawn.
Ties are resolved in expectation, so a constant-score baseline lands on the floor rather than
wherever the pool order puts it. Exact Top-1 cells are in
Tables~\ref{tab:gold-full} and~\ref{tab:gold-matched}. Produced by
\texttt{figure/make\_gold.py} from the shipped board.}
\label{fig:gold}
\end{figure}

\paragraph{Gold Boards Remain Mechanism Diagnostics.}
The three Gold-derived boards are reported here as mechanism diagnostics rather than as
artifact-controlled benchmark evidence, for the reason established in
Section~\ref{sec:data-gold}. Methods rank the steps of a run; the injected step is the target.
Figure~\ref{fig:gold} reports both pools at a representative injection seed,
split by fault kind because the aggregate hides that the two faults behave differently.

\paragraph{Stale State Localizes; Dropped Grounding Does Not Follow.}
Table~\ref{tab:defensibility-verdicts} limits the evidential status of these results. The injected
stale-state mechanism is localizable on this substrate. A dependency-span detector reaches 0.703
Top-1 for this seed and 0.653 $\pm$ 0.028 across five injection seeds. Both exceed the 0.029 random
floor for that fault kind.
Dropped grounding is another matter. Across the same five
injection seeds the span detectors reach 0.005 $\pm$ 0.000 and degree 0.029 $\pm$ 0.008, while the
generic graph detector is the only displayed reading above that kind's 0.035 floor. Its two seed axes
answer different questions and we report both: holding its
initialization fixed, it reaches 0.072 $\pm$ 0.024 across those five injection seeds, and holding
injection seed 0 fixed while varying five initialization seeds gives 0.085 $\pm$ 0.010. Only
\texttt{has-dep} carries a recorded comparison against the dropped-grounding floor; span, degree,
and the graph detector carry none, so those three are displayed cells read against a displayed
floor. Removing one dependency among many leaves little signal the current detectors catch, an
open problem rather than a result to average away.

\paragraph{The Eligibility-Matched Control.} The injector selects sites from the clean run using
fault-specific eligibility, and the full-pool controls show that the selection leaks. \texttt{has-dep}
scores 0.173 on stale-state against a 0.029 random floor for that kind, and 0.078 overall against
0.032. On dropped grounding the same marker inverts, scoring 0.005 against a 0.035 floor, because that
injection removes the dependency the marker selects on. Both directions are the same eligibility
artifact, and the second is the sharper evidence for it: a control that is anti-informative is reading
the construction, not the fault. Degree's 0.045 overall is a displayed cell above the 0.032 floor,
with no recorded comparison between the two, so it is reported for completeness. This is what eligibility selection should be expected to produce, and it is
why the matched control is necessary rather than a precaution. We rank each method only within the steps the injector could have targeted for that run's
fault kind (mean 7.4 candidates per run). The eligible pool mirrors the injector's own precondition,
so the true injected step is always inside it. Both Gold tables score ties in expectation, which
matters here because two of these baselines assign the same score to many candidates.

\paragraph{Eligibility Matching Does Not Match Degree.}
What the pool holds constant is eligibility, and only eligibility. For stale-state, \texttt{has-dep}
is constant in the eligible pool and scores 0.350 against a 0.350 floor. For dropped grounding, the
target can lose its only dependency, so \texttt{has-dep} scores 0.075 against a 0.277 floor. Degree is
not equalized: eligible steps still differ in how many dependencies they carry, so the control is a
selection control rather than a full degree match. The board makes that visible.
\texttt{degree} scores 0.516 Top-3 inside the pool against a 0.622 random floor, and a statistic held
genuinely constant could not fall below its floor at all. Reading that gap as mild anti-information is
more accurate than reading the pool as degree-matched, and it is why the control is a partial check on
construction leakage rather than a complete one. Figure~\ref{fig:gold} draws the move as the
filled marker in each pair, and Tables~\ref{tab:gold-full} and~\ref{tab:gold-matched} in
Appendix~\ref{app:board-values} carry the exact cells.

\paragraph{Eligibility-Matched Localization Readings Are Untested.}
Within the pool, max-span displays 0.805 on stale-state against the 0.350 random floor, and 0.795
$\pm$ 0.020 across five injection seeds. The corresponding \texttt{auditable} (dep-anomaly) score is
0.799, and degree displays 0.394. For dropped grounding, position displays 0.321, degree 0.095, and
\texttt{has-dep}, max-span, and \texttt{auditable} (dep-anomaly) each display 0.075 against the
0.277 floor. No matched-pool method-versus-floor comparison is recorded on this board, so every
reading here is a displayed cell set beside a displayed floor.

\paragraph{Aggregate Gold Rankings Hide Fault Differences.}
The overall column should not be read as a ranking, and PyGOD shows why. It is nominally highest there
at 0.404, and its two uncertainty axes are not interchangeable. Across twenty
initialization seeds it reaches 0.425 $\pm$ 0.038, and against max-span's deterministic 0.394 that
gives $+0.031$ with a 95\% interval of $[0.012, 0.049]$; that interval is over initializations of a
fixed run set. Only 39 of the 188 runs rank the two
differently at all, so a run-resampling interval answers a different question about the same pair of
cells. This is the distinction the overall column is too coarse to carry. It leads neither fault
kind on either axis: max-span is far ahead on stale-state (0.805 against 0.662 $\pm$ 0.036) and PyGOD
does not clear the floor on dropped grounding (0.242 $\pm$ 0.066, interval $[0.210, 0.273]$, against
0.277). What
the aggregate rewards is evenness across two kinds on which every entrant is weak somewhere, so the
per-kind columns carry the result and the overall column carries an average of unlike things. The
construction-artifact verdict in Table~\ref{tab:defensibility-verdicts} determines the evidential status of all of it.

\paragraph{Cause Attribution.}
The third forensic question is what kind of fault occurred, given that one did. We answer it on
\sysname-Gold (Section~\ref{sec:data-gold}), where the cause is known by construction. To keep the
label the fault rather than the run, the substrate is paired. Every run that affords both faults is
injected twice, once with a stale-state read and once with dropped grounding. The two classes are
therefore the same runs, and a classifier cannot separate them on run identity. This is the run-level analogue of
the localization eligibility-matched selection control. On 166 paired runs the two faults leave opposite structural
traces, and a single run-level feature tells them apart above the 0.498
chance floor. A stale
read lengthens the maximum dependency span (ROC-AUC 0.675, and 0.671 $\pm$ 0.005 across five injection
seeds), while dropped grounding removes one edge (edge-count ROC-AUC 0.566). Each feature is keyed to
one mechanism. As an artifact-limited Gold result, this board reports mechanism-discriminative
information only.

\paragraph{Gold v2 Admissibility Status.}
Stale-state fails on corpus adequacy, and that finding stands. Enumerating injectable sites over the
full 660-run corpus yields 16 stale-state sites in 6 runs, against 2077 dropped-grounding sites in
614 runs. This census is independent of the scorer. Sixteen sites in six runs cannot support a board
and supply no natural example for a realism check. Field-level supersession is rare on tau-bench
because an agent mostly supersedes its own values rather than reading a value that a third party has
since overwritten.

\paragraph{Dropped Grounding Passes the Fixed Margins.}
Dropped grounding is plentiful. Under the corrected tie rule, all six process-artifact controls sit
at the matched floor within the declared margin across five injection seeds. The tightest is \texttt{edit-distance} at $+0.034 \pm 0.008$ Top-1 above floor,
whose 0.042 upper bound falls inside the $\pm 0.05$ band, with run-level AUC $0.512 \pm 0.002$ inside
$[0.45, 0.55]$. The \texttt{provenance} oracle reaches $+0.503 \pm 0.000$, so the panel does register
a fault predicate when one is present.

\paragraph{Missing Control Power Blocks Admissibility.}
This is still not a PASS, and the obstacle is control power rather than the readings themselves.
\texttt{tools/namedvalue\_control\_power.py} plants a known artifact and checks that the matching
control fires. It demonstrates power for three of the six controls on Top-1 alone: format-outlier at
$+0.529$, schema-shape at $+0.393$, and position-prior at $+0.379$ above floor. Power for field-prior,
tool-prior, and \texttt{edit-distance} is undemonstrated, and \texttt{edit-distance} is the control
sitting closest to the margin. A control at the floor is evidence of a clean injector only if that
control would have left the floor had the injector been dirty. Dropped grounding therefore passes the
current fixed-margin readings, and its admissibility stays undetermined until all six controls have
demonstrated power on Top-1 and on run-level AUC. What the tie fix settles is the earlier
\textsc{not admissible} reading, which came from the tie rule rather than from the injector. For the
next release, we plan to test whether AppWorld cross-app writes can supply the third-party
supersession that tau-bench lacks.

\section{Artifact and Reproducibility Detail}
\label{app:artifact}

This section carries what the Reproducibility Statement points at. It describes the artifact's
guarantees because paths and commands in a paper go stale when they move.\looseness=-1

\paragraph{The Scored Board Needs No Model Call.} The scoring entry point reads the committed POST
prediction caches and the committed PRE held-out-judge cache, and calls no model service. Each
full-board run needs network access for its corpus preflight; missing local copies also require
corpus downloads. All three corpora have recorded full commit hashes in
Table~\ref{tab:revision-record}. The preflight resolves each dataset's current head, compares it
against the recorded commit, and stops before scoring if any differs or cannot be resolved.
The \sysname GRADE bridge also forces the loaders for all three corpora to use their recorded
hashes, covering Who\&When snapshot downloads, SWE-Gym shard discovery and download, and tau-bench
file downloads. It rejects a conflicting explicit revision. Before scoring, the full-board runner
checks for an observed pinned download or, for a Who\&When cache-only load, matching snapshot
metadata. The tau-bench loader also passes its hash directly inside GRADE; direct GRADE execution
outside \sysname leaves Who\&When and SWE-Gym unpinned. A released tool prints the three hashes,
and the board header records them. Prediction caches are addressed by a digest of the source record
rather than by load order, so adding or removing a corpus file cannot silently reassign a cached
prediction to a different run.\looseness=-1

\paragraph{Figures and Tables Read One Artifact.} The figure scripts ship with this manuscript's
source rather than with the benchmark repository, and each parses a released artifact: the board
file for the four board figures, and the statistics registry for the localization, detection, and
PRE-source panels. A figure
and the table beside it therefore read the same numbers. One exception is marked in the source: the
four outside-the-arena values drawn below the rule in Figure~\ref{fig:localization-panel}(a) are
literals in that panel's script, because the addendum caches they come from sit in the benchmark
repository rather than beside the manuscript. A test in that repository fails if either stops
matching the other. The benchmark repository carries its own
generators for the two board figures it publishes, with a check that fails when a committed image no
longer matches the board it was drawn from.

\paragraph{Continuous Integration.} The test suite runs on Python 3.10 and 3.12 under two hash
seeds. A committed manifest enumerates every test node id the suite claims to run and marks the
nineteen that may skip, four needing the graph-AD stack and fifteen needing a manuscript checkout; the
job fails on a skip anywhere else, so a contract cannot rot behind a growing exemption. An offline smoke job asserts the six PRE source counts, the 1187 total, and the two PRE
floors. A separate job checks the JSON shape of the 31 POST caches and the one PRE method cache.
Prediction generation sits deliberately outside the scoring path, because remote-model outputs vary
between calls; the committed predictions are the scored artifacts.

\paragraph{What a Third Party Can and Cannot Rerun.} Without an API key a third party can run the
complete scored board, the tests and smoke checks, the Who\&When split report, the PRE label merge
from committed votes, the PII scan, and the Gold diagnostics. Generating new POST judge predictions
requires access to a supported model backend. The released generators support the NAIRR gateway,
AWS Bedrock, OpenAI-compatible endpoints, and subscription CLIs for the configured Codex and
Claude models. The GPT-5.5 reference is documented as generated with \texttt{codex exec}; its
historical cache does not record the channel. These backends produce new predictions; exact
rescoring uses the committed caches. Regenerating the PRE held-out-judge cache additionally
requires a prose-retaining PRE harvest held outside the repository, because the released records
have had that prose removed, so the released records cannot rebuild those prompts. What is not
pinned is stated in Table~\ref{tab:revision-record}.
\begin{table*}[t]
\centering
\scriptsize
\begin{tabular}{p{1.6cm}p{6.0cm}p{4.8cm}}
\toprule
Component & Revision & Enforcement \\
\midrule
tau-bench & \texttt{382e57d1784b55c5155f4ef394ef48f1c747a287} &
Checked by the full-board preflight and enforced by the \sysname Hub wrapper. The GRADE loader also
passes this hash directly. \\
Who\&When & \texttt{59b9fcba1aaed7bbf206b5f4d3c68b8face2f49c} &
Checked by the full-board preflight and enforced on snapshot downloads by the \sysname Hub wrapper.
Cache-only full-board loads require matching snapshot metadata. Direct GRADE execution outside
\sysname remains unpinned. \\
SWE-Gym & \texttt{baf3a4e4bff514d48ddc08a93a2ade5c126212c7} &
Checked by the full-board preflight and enforced on shard discovery and download by the \sysname
Hub wrappers. Direct GRADE execution outside \sysname remains unpinned. \\
GRADE & \texttt{3839a57ac165d58a807fce0a3ff38346732ee936} &
Pinned by the CI workflow. \\
PRE records & Per-record \texttt{repo}, \texttt{commit}, \texttt{path}, \texttt{license} &
Committed with each record. n8n stores template identifiers and SWE-agent stores a submission
identifier rather than a repository commit. \\
\bottomrule
\end{tabular}
\caption{Corpus and code revisions. A commit pin is not an archival guarantee: if an upstream dataset
is deleted or access is withdrawn, a fresh download fails rather than silently returning different
data. The exact \texttt{auditable} revision and a resolved dependency environment are not yet locked.}
\label{tab:revision-record}
\end{table*}

Every generated table in this paper carries a \texttt{--check} mode that exits non-zero and prints
the delta when the paper falls behind. Three examples are Table~\ref{tab:boards} from the shipped
board, Table~\ref{tab:all-contrasts} from the recorded comparisons under
\texttt{tools/emit\_stats\_table.py -{}-contrasts},
and Table~\ref{tab:transfer} from the two seed records. Those checks run locally before a submission rather than in CI, because the
manuscript repository is not public and no workflow supplies its path, so the seven tests that hold
the paper equal to the boards always skip there. Table~\ref{tab:diagnostic-source} lists the body
quantities that a committed script prints rather than a committed record holds, while the
Who\&When Pro trace count and the AuthBench task count come from their cited papers rather than from
any command here. The distinction
matters when reading this paper: a value in a generated table has an automated
comparison against its source, run by hand, while a value in Table~\ref{tab:diagnostic-source} is
compared by reading or is marked there as not reproducible from the released artifact.

\begin{table*}[t]
\centering
\scriptsize
\begin{tabular}{@{}>{\raggedright\arraybackslash}p{4.2cm}>{\raggedright\arraybackslash}p{5.25cm}>{\raggedright\arraybackslash}p{3.5cm}@{}}
\toprule
Quantity, and where the body prints it & Command, run from the repository root & What that command needs \\
\midrule
Who\&When split report: the scored population, the \texttt{is\_correct} against
\texttt{is\_corrected} key split, and the per-split random Top-1 floors
(Section~\ref{sec:data-corpora}) &
\texttt{tools/whoandwhen\_split\_report.py} &
The cached Who\&When corpus, which the localization board writes on its first run. \\
\addlinespace
PRE label makers scored as methods, and the held-out model on the 656 configurations all three
judged (Section~\ref{sec:res-pre}) &
\texttt{tools/pre\_label\_maker\_diagnostic.py} &
The committed judge votes and the held-out prediction cache. It rescores stored votes and calls no
model. Generating fresh judgments is the operation that would need model access and the prose the
de-identified records exclude. \\
\addlinespace
Gold v2 named-value admissibility: site and run counts, the fixed-margin readings, and the oracle
control (Appendix~\ref{sec:res-gold}) &
\texttt{tools/namedvalue\_admissibility.py} and
\texttt{tools/namedvalue\_control\_power.py} &
The tau-bench corpus and the GRADE checkout. Gold v2 is built over tau-bench trajectories, unlike
the file-level Gold board below, which uses SWE-Gym. \\
\addlinespace
Gold file-level construction leakage (Section~\ref{sec:data-gold}) &
\texttt{tools/gold\_artifact\_diagnostic.py} &
The SWE-Gym corpus and the GRADE checkout. \\
\bottomrule
\end{tabular}
\caption{Body quantities that a committed script prints rather than a committed record holds. This
is weaker than the generated tables: nothing recomputes these when the paper changes, so a reader
who wants to check one runs the command. Two values comparing this benchmark with published work,
the Who\&When Pro trace count and the AuthBench task count, come from those papers rather than from
any command here and carry their citations in place.}
\label{tab:diagnostic-source}
\end{table*}

\end{document}